\documentclass{article}

\PassOptionsToPackage{numbers, compress}{natbib}

\usepackage[main, final]{neurips_2026}

\usepackage{nicefrac}

\usepackage{graphicx}

\usepackage[utf8]{inputenc} %
\usepackage[T1]{fontenc}    %

\usepackage{float}
\usepackage{wrapfig}

\usepackage[dvipsnames,table]{xcolor}

\usepackage[colorlinks,linktoc=all]{hyperref}
\usepackage[all]{hypcap}
\hypersetup{citecolor=Brown}
\hypersetup{linkcolor=RoyalPurple}
\hypersetup{urlcolor =RoyalPurple}

\usepackage[colorinlistoftodos,backgroundcolor=orange!20,textwidth=2.6cm]{todonotes}
\makeatletter
\renewcommand{\todo}[2][]{\tikzexternaldisable\@todo[#1]{\small{#2}}\tikzexternalenable}
\makeatother

\usepackage{url}            %
\usepackage{booktabs}       %
\usepackage[detect-all]{siunitx}
\usepackage{amsfonts}       %
\usepackage{nicefrac}       %
\usepackage{microtype}      %
\usepackage[ruled,linesnumbered,noend]{algorithm2e}

\SetCommentSty{mycommfont}

\usepackage{amssymb,amstext,amsmath,amsthm,amsfonts,pifont,array}
\usepackage{mathtools}

\usepackage{nicematrix}

\usepackage[capitalize,nameinlink]{cleveref}
\crefname{section}{\S}{\S\S}
\Crefname{section}{\S}{\S\S}
\creflabelformat{equation}{#2\textup{#1}#3}

\definecolor{mylightgray}{gray}{0.95}
\definecolor{mylightgray2}{gray}{0.96}

\usepackage{pifont}

\usepackage{enumitem}

\usepackage[acronym,nowarn]{glossaries}
\setacronymstyle{long-sc-short}
\glsdisablehyper
\makeglossaries
\usepackage{multirow}
\usepackage{tabularx}
\usepackage{mleftright}
\usepackage{xr-hyper}
\usepackage{adjustbox}
\usepackage{footmisc}
\usepackage{arydshln}
\usepackage{amssymb}

\usepackage{listings}

\definecolor{codegreen}{rgb}{0,0.6,0}
\definecolor{codegray}{rgb}{0.5,0. ,0.5}
\definecolor{codepurple}{rgb}{0.58,0,0.82}
\definecolor{backcolour}{rgb}{0.95,0.95,0.92}
\lstdefinestyle{mystyle}{
    backgroundcolor=\color{white},
    numberstyle=\tiny\color{codegray},
    stringstyle=\color{codepurple},
    basicstyle=\fontsize{7.5pt}{7.5pt}\ttfamily\selectfont,
    commentstyle=\fontsize{7.5pt}{7.5pt}\color{codegreen},
    keywordstyle=\fontsize{7.5pt}{7.5pt}\color{magenta},
    breakatwhitespace=false,
    breaklines=true,
    captionpos=b,
    keepspaces=true,
    numbers=left,
    numbersep=5pt,
    showspaces=false,
    showstringspaces=false,
    showtabs=false,
    tabsize=2
}
\usepackage{epstopdf,caption,tablefootnote}
\usepackage{tikz,tikzscale,subcaption}
 \usetikzlibrary{arrows.meta,calc,decorations.markings,math,arrows.meta,shapes,arrows}
 \tikzset{%
             base/.style = {rectangle, draw=black,
                            minimum width=4cm, minimum height=1cm,
                            text centered, font=\rmfamily},
   binary/.style = {base, minimum width=1cm},
        startstop/.style = {base, fill=red!30, minimum width=2cm},
     activityRuns/.style = {base, fill=green!30},
          process/.style = {base, minimum width=2cm, fill=gray!15,
                            font=\ttfamily},
          sum/.style      = {draw, circle, node distance = 1.5cm},
 }
 \usetikzlibrary{positioning}
 \usetikzlibrary{shapes.misc}
 \usetikzlibrary{automata,arrows,positioning,calc}

\newcommand{\contributionnum}[1]{
  \begingroup
  \tikzexternaldisable
  \tikz[baseline=(num.base)]{
    \node[
      circle,
      fill=Brown,
      text=white,
      font=\bfseries\small,
      inner sep=0pt,
      minimum size=1.0em
    ] (num) {#1};
  }%
  \endgroup
}

\usepackage{pgfplots}
\pgfplotsset{compat=1.6}
\usepgfplotslibrary{groupplots}
\tikzstyle{every picture}+=[font=\rmfamily]
\tikzstyle{optimized} = [circle,fill=white,draw=black, dashed,inner sep=1pt, minimum size=20pt, font=\fontsize{10}{10}\selectfont, node distance=1]
\pgfkeys{/pgf/number format/.cd,1000 sep={}}
\pgfplotsset{
	tick label style = {font=\rmfamily},
	every axis label/.append style={font=\rmfamily},
	typeset ticklabels with strut,
}
\pgfplotsset{every axis/.append style={
			every x tick label/.append style={font=\fontsize{6pt}{6pt}\rmfamily, yshift=.5ex,},
			every y tick label/.append style={font=\fontsize{6pt}{6pt}\rmfamily, xshift=.5ex},
			every y label/.append style={xshift=10ex, font=\rmfamily},
			every x label/.append style={yshift=3ex, font=\rmfamily},
			every title/.append style={font=\rmfamily}
		},
}
\pgfplotsset{
  xticklabel={$\mathsf{\pgfmathprintnumber{\tick}}$},
  yticklabel={$\mathsf{\pgfmathprintnumber{\tick}}$},
}
\pgfplotsset{every axis title/.append style={yshift=-1.5ex}}
\newlength\figureheight
\newlength\figurewidth
\usepgfplotslibrary{external}
\usepackage{todonotes}

\definecolor{color_binarynet}{HTML}{bcbd22}
\definecolor{color_binaryconnect}{HTML}{17becf}
\definecolor{color_xnornet}{HTML}{2ca02c}
\definecolor{color_birealnet}{HTML}{e377c2}
\definecolor{color_real2binary}{HTML}{8c564b}
\definecolor{color_reactnet}{HTML}{9467bd}
\definecolor{color_meliusnet}{HTML}{7f7f7f}
\definecolor{color_bnext}{HTML}{ff7f0e}
\definecolor{color_pokebnn}{HTML}{1f77b4}
\definecolor{color_booldl}{HTML}{d62728}

\definecolor{my_box_color}{HTML}{f5f1ea}

\usepackage{tcolorbox}
\definecolor{lightblue}{HTML}{ff7f2a}
\definecolor{lighterblue}{HTML}{ffe6d5}%
\newtcolorbox{mybox}{colback=my_box_color,colframe=black!90,boxrule=1.1pt,top=0.4pt,bottom=0.4pt,right=1.8pt,left=1.8pt}
\newtcolorbox{myboxcontrib}{colback=mylightgray2,colframe=mylightgray2,top=1.2pt,bottom=1.2pt,right=1.8pt,left=1.8pt}
\newtcolorbox{myboxmain}{
    colback=my_box_color, 
    colframe=black!90,       %
    boxrule=1.1pt,        %
    top=0.4pt, 
    bottom=0.4pt, 
    right=1.8pt, 
    left=1.8pt
}

\definecolor{asparagus}{rgb}{0.53, 0.66, 0.42}
\definecolor{b_df_c}{rgb}{0, 0.69, 0.31}
\definecolor{b_block_c}{rgb}{0.57, 0.81, 0.31}
\definecolor{fp_df_c}{rgb}{0.78, 0, 0.04}
\definecolor{fp_block_c}{rgb}{0.78, 0, 0.04}
\definecolor{int_df_c}{rgb}{0, 0.2, 0.8}

\usepackage{setspace}

\usepackage{algorithmic}

\usepackage{amsmath}
\usepackage{mathtools, nccmath}
\usepackage{color}

\definecolor{color_blue}{HTML}{1f77b4}
\definecolor{color_orange}{HTML}{ff7f0e}
\definecolor{color_green}{HTML}{2ca02c}
\definecolor{color_red}{HTML}{d62728}
\definecolor{color_purle}{HTML}{9467bd}
\definecolor{color_olive}{RGB}{188,189,34}

\definecolor{cFlat}{HTML}{4DAF4A}    %
\definecolor{cSharp}{HTML}{E41A1C}   %
\definecolor{cMid}{HTML}{FF7F00}     %
\definecolor{cBlue}{HTML}{377EB8}    %
\definecolor{cGray}{HTML}{999999}    %
\definecolor{cPurple}{HTML}{984EA3}  %
\definecolor{cYellow}{HTML}{FFD92F}  %

\newcommand{\cubeedges}[1]{%
  \draw[hedge] (#1-000)--(#1-100); \draw[hedge] (#1-000)--(#1-010);
  \draw[hedge] (#1-110)--(#1-100); \draw[hedge] (#1-110)--(#1-010);
  \draw[hedge] (#1-001)--(#1-101); \draw[hedge] (#1-001)--(#1-011);
  \draw[hedge] (#1-111)--(#1-101); \draw[hedge] (#1-111)--(#1-011);
  \draw[hedge] (#1-000)--(#1-001); \draw[hedge] (#1-100)--(#1-101);
  \draw[hedge] (#1-010)--(#1-011); \draw[hedge] (#1-110)--(#1-111);}
\newcommand{\cubecoords}[2]{%
  \coordinate (#1-000) at (#2,0);
  \coordinate (#1-100) at ({#2+\dx},0);
  \coordinate (#1-010) at (#2,\dy);
  \coordinate (#1-110) at ({#2+\dx},\dy);
  \coordinate (#1-001) at ({#2+\sx},\sy);
  \coordinate (#1-101) at ({#2+\dx+\sx},\sy);
  \coordinate (#1-011) at ({#2+\sx},{\dy+\sy});
  \coordinate (#1-111) at ({#2+\dx+\sx},{\dy+\sy});}

\NewDocumentEnvironment{repeatenv}{m m o}{%
\expandafter\renewcommand\csname the#1\endcsname{#2}%
\IfValueTF{#3}{%
  \begin{#1}[#3]%
}{%
  \begin{#1}%
}%
}{%
\end{#1}%
}

\newacronym{KD}{kd}{knowledge distillation}
\newacronym{BNN}{bnn}{binarized neural network}
\newacronym{STE}{ste}{straight-through-estimator}
\newacronym{SE}{se}{squeeze-and-excitation}
\newacronym{CNN}{cnn}{convolutional neural network}
\newacronym{FP}{fp}{full-precision}
\newacronym{SOTA}{sota}{state-of-the-art}
\newacronym{LLM}{llm}{large language model}
\newacronym{OP}{op}{compute operation}
\newacronym{NN}{nn}{neural network}
\newacronym{BN}{bn}{batch normalization}
\newacronym{FC}{fc}{fully-connected}
\newacronym{ASPP}{aspp}{atrous pyramid pooling}
\newacronym{MSE}{mse}{mean-squared error}
\newacronym{SVID}{svid}{sign-value-independent decomposition}
\newacronym{SVD}{svd}{singular value decomposition}
\newacronym{PWCCA}{pwcca}{projection weighted canonical correlation analysis}
\newacronym{KL}{kl}{Kullback–Leibler}
\newacronym{PTQ}{ptq}{Post-Training Quantization}
\newacronym{QAT}{qat}{Quantization-Aware Training}

\newacronym{MBOK}{mbok}{Multiple Boolean Kernels}

\newacronym{CCA}{cca}{Canonical Correlation Analysis}

\theoremstyle{plain}
\newtheorem{theorem}{Theorem}[section]
\newtheorem{proposition}[theorem]{Proposition}
\newtheorem{lemma}[theorem]{Lemma}
\newtheorem{corollary}[theorem]{Corollary}

\newtheorem{definition}[theorem]{\textit{Definition}}

\theoremstyle{remark}
\newtheorem{remark}[theorem]{\textit{Remark}}%

\usepackage{xargs}

\makeatletter

\def\env@sqcases{%
	\let\@ifnextchar\new@ifnextchar
	\left\lbrack
	\def\arraystretch{1.2}%
	\array{@{}l@{\quad}l@{}}%
}
\makeatother

\renewcommand{\d}[1]{\ensuremath{\operatorname{d}\!{#1}}}

\DeclarePairedDelimiterX{\infdivx}[2]{[}{]}{%
  #1\;\delimsize\|\;#2%
}

\title{Navigating the Boolean Hypercube: Sharpness-Aware Minimization Directly in Discrete Loss Landscapes}

\vspace{-5.3ex}

\author{%
Ba-Hien Tran \\
Huawei Technologies France SASU \\
Paris Research Center \\
\texttt{ba.hien.tran@huawei.com} \\
}

\vspace{-3.2ex}

\begin{document}

\addtocontents{toc}{\protect\setcounter{tocdepth}{0}}

\maketitle

\vspace{-4ex}

\begin{abstract}
  \vspace{-0.5ex}

Sharpness-Aware Minimization (SAM) improves generalization in deep learning with continuous weights.
However, applying it to binary neural networks through latent-weight
heuristics suffers from a geometric mismatch:
continuous perturbations do not faithfully probe the discrete loss
landscape on $\{-1,+1\}^n$.  We introduce Hamming-SAM, the first
sharpness-aware optimizer that operates natively on the Boolean
hypercube.  It replaces the Euclidean $\ell_p$-ball with a $k$-bit
Hamming ball and solves the resulting discrete min-max problem by greedy
ascent followed by sharpness-aware bit-flip descent.  The objective is
theoretically justified through the lens of PAC-Bayes, compression, and
distributionally robust optimization.  We prove an
approximation guarantee for the ascent step and a finite-time convergence
bound for the descent.  Both are governed by the discrete interaction
Hessian, the classical second-order derivative of pseudo-Boolean analysis,
which serves as the unifying quantity connecting ascent quality to
convergence rate. We further connect discrete flatness to Forman-Ricci curvature and algorithmic complexity.
Experiments on a wide range of models and datasets demonstrate consistent improvements in
in-distribution accuracy and out-of-distribution robustness over standard binary
training and latent-space SAM baselines.

\end{abstract}

\vspace{-2.0ex}

\section{Introduction}
\label{sec:intro}

\vspace{-1.1ex}

Deep learning faces a growing tension between the computational demands of high-dimensional
continuous optimization and the physical limitations of hardware at the
edge.  Traditional neural architectures rely on weight optimization within
Euclidean space.  This approach has been very successful, but at the
cost of high energy consumption and large memory
footprints.  Interest has therefore returned to discrete models, in
particular Binarized Neural Networks
(BNNs)~\citep{hubara2016binarized, rastegari2016xnor}, e.g., BitNet~\citep{wang2025bitnet}.
A parallel line of work has progressively removed the real-valued
weights traditionally used to train such models.
\cite{helwegen2019latent} showed that latent weights act as inertia
rather than as parameters in their own right, and the Boolean Logic Deep
Learning (BOLD) framework~\citep{nguyen2024bold,tran2026highly} removes
them altogether, operating directly on the Boolean hypercube
$\{-1,+1\}^n$.  However, optimizing over a discrete domain raises
challenges absent from continuous training.  The loss landscape
is defined on an exponentially large combinatorial space ($2^n$
configurations), gradient-based methods cannot be applied directly, and
 finding a global minimum is NP-hard in
general~\citep{blum1988training}.  The loss surface also exhibits complex
interactions between bit-flips, where flipping two bits together can
produce a loss change far exceeding the sum of their individual
contributions.  This makes the landscape very different
from the smooth manifolds of continuous optimization.

\vspace{-0.3ex}

Despite their efficiency, these models share a key
weakness.  Optimizers tend to converge toward \emph{sharp}
minima, that is, weight configurations where a few bit-flips
cause large loss increases, which degrades
generalization.  While the connection between flat minima and
generalization is well-established in continuous space~\citep{hochreiter1997flat, keskar2017largebatch,
jiang2020fantastic}, the relationship is subtle:
\cite{dinh2017sharp} showed that common flatness measures are sensitive
to reparameterization in continuous networks, complicating their use as
generalization predictors.
Sharpness-Aware Minimization
(SAM)~\citep{foret2021sam} has become an effective and well-founded remedy
with strong theoretical
foundations~\citep{kwon2021asam, khanh2024sam}, yet transferring these
insights to the binary or discrete domain remains unexplored.
Existing approaches apply SAM to BNNs and quantized networks through
latent-weight proxies~\citep{liu2021saq, liu2023bnnsam, pu2024bnnsam}.  Some
reshape the perturbation to counter the sign barrier, but all of them perturb a
continuous surrogate.  This approach faces two geometric pathologies, which we
formalize in \cref{sec:pathology}.  Unless the perturbation is reshaped, its
signal \emph{vanishes} as latent weights grow, and in every case Euclidean
proximity \emph{fails to predict} Hamming proximity.
More broadly, a theoretical understanding of flatness in discrete
parameter spaces remains lacking.  What flatness means there, how to
measure it, and how to optimize for it are all open questions.
In contrast, the reparameterization ambiguities
that affect continuous flatness measures~\citep{dinh2017sharp} are
\emph{absent} on the Boolean hypercube.  Since $\{-1,+1\}^n$ admits no
smooth rescaling symmetries, Hamming-ball sharpness is an unambiguous
measure of the loss landscape geometry.

\vspace{-2.2ex}

\paragraph{Contributions.}
This paper addresses these issues with Hamming-SAM, a native
Boolean optimizer that replaces the continuous $\ell_p$-ball with a
$k$-bit Hamming ball, the natural neighborhood on $\{-1,+1\}^n$, and
solves the resulting discrete min-max problem via a two-phase greedy
procedure.
The framework is not tied to any binary training method.  Its
two ingredients are the first- and second-order discrete derivatives of
the loss, classical objects of pseudo-Boolean function
theory~\citep{crama2011boolean, odonnell2014analysis} that are defined on
$\{-1,+1\}^n$ independently of how they are computed.  Any per-bit
discrete-derivative oracle can therefore serve as the backend, such as
BOLD's chain rule~\citep{nguyen2024bold} or a straight-through estimator
(STE). 
Our theoretical analysis is general and does not depend on the specific backend used.
Our contributions span both formulation and analysis:
\contributionnum{1}~the objective is theoretically
justified from three perspectives, namely PAC-Bayes, compression,
and distributionally robust optimization (\cref{sec:method});
\contributionnum{2}~an approximation guarantee for
the greedy ascent based on the submodularity ratio, with graceful
degradation for non-submodular landscapes (\cref{thm:ascent});
\contributionnum{3}~a finite-time convergence
proof linking the SAM-style descent to the discrete interaction
Hessian (\cref{thm:convergence,thm:convergence_sam}), which serves as
the unifying quantity governing both ascent quality and convergence rate;
\contributionnum{4}~connections between discrete
flatness, Forman-Ricci curvature, and algorithmic complexity
(\cref{sec:perspectives});
and \contributionnum{5}~experiments on CIFAR-10/100
and ImageNet with VGG, ResNet, and ViT demonstrating consistent
improvements in clean accuracy and out-of-distribution robustness
over binary training and latent-space SAM baselines
(\cref{sec:experiments}).

\vspace{-1.4ex}

\section{Related Work}
\label{sec:related}

\vspace{-1.5ex}

\paragraph{SAM and its variants.}
\cite{foret2021sam} introduced SAM as a min-max optimizer seeking
parameters with uniformly low loss in an $\ell_p$-ball neighborhood.
Subsequent work refined the perturbation geometry~\citep{kwon2021asam},
reduced cost~\citep{liu2022looksam}, and improved the ascent
direction~\citep{kim2022fishersam}.
\cite{khanh2024sam} recently established convergence of continuous SAM
by viewing it as inexact gradient descent.
All existing variants operate exclusively in continuous Euclidean
parameter spaces.

\vspace{-1.1ex}

\paragraph{Binary neural networks.}
Modern BNNs maintain latent real-valued weights and project them via the
sign function, using the Straight-Through Estimator (STE) for gradient
approximation~\citep{hubara2016binarized, rastegari2016xnor}, with
architectural advances \citep{liu2020reactnet, liu2018bi, zhang2022pokebnn, xing2022towards}.
A separate line asks whether the latent weights are needed at all.
\cite{helwegen2019latent} reinterpreted them as inertia and optimized the
signs directly, and BOLD~\citep{nguyen2024bold} removed them and
introduced a chain rule that yields a learning signal for every bit in a
single backward pass.  We adopt that chain rule as our backend, but the
sharpness framework developed here is independent of it and applies to
any oracle for the same classical quantity.
\cite{liu2023bnnsam, pu2024bnnsam} apply SAM to BNNs, but operate on
latent continuous weights.
\cite{liu2021saq} documented ``perturbation diminishment'' in quantized
networks, partially anticipating our Vanishing Perturbation Effect.

\vspace{-1.1ex}

\paragraph{Flat minima in discrete spaces.}
\cite{baldassi2015subdominant} identified dense clusters of solutions in
binary networks, and \cite{baldassi2020shaping} formalized local
entropy as a discrete flatness measure.  Their optimization relies on
replicated SGD and belief propagation.  Our
objective~\eqref{eq:hamming_sam_objective} instead minimizes the worst-case
loss over a Hamming ball, which yields a deterministic guarantee rather
than a statistical one.

\vspace{-1.1ex}

\paragraph{Pseudo-Boolean optimization and discrete curvature.}
The discrete Hessian is rooted in pseudo-Boolean function
theory~\citep{crama2011boolean, boros2002pseudo}, where second-order
discrete derivatives have been studied
extensively~\citep{grabisch1999axiomatic, odonnell2014analysis}.  We
apply these tools to the loss landscape of Boolean networks,
connecting it to the submodularity ratio that governs the ascent and to
the convergence rate of the descent.
Forman-Ricci curvature~\citep{forman2003bochner} has been applied to
data graphs~\citep{topping2022understanding} and network
architectures~\citep{rieck2019discrete}.  We are the first to apply it
to a neural network's \emph{parameter space}.

\vspace{-1.8ex}

\section{The Geometric Pathology of Latent-Weight SAM}
\label{sec:pathology}

\vspace{-1.3ex}

The primary motivation for developing a native Boolean SAM arises from two
geometric pathologies of applying continuous SAM to binarized networks (see
also~\cite{liu2021saq} for related observations in quantized settings). In the
standard approach, high-precision latent weights $\tilde{w} \in \mathbb{R}^n$
are maintained and projected onto the Boolean space via $w =
  \mathrm{sign}(\tilde{w}) \in \{-1,+1\}^n$. When continuous SAM is applied to
these latent weights, it searches for an adversarial perturbation $\epsilon$
within an $\ell_p$-ball of radius $\rho$ to maximize the loss. This approach
suffers from two failures that we formalize below.

\vspace{-2.1ex}

\subsection{The Vanishing Perturbation Effect}

\vspace{-0.7ex}

\begin{myboxmain}
  \begin{proposition}[Vanishing Perturbation]
    \label{prop:vanishing}
    Let $\mathcal{C}(\rho) = \{i : |\tilde{w}_i| > \rho\}$ be the ``confident'' coordinates.
    For any $\epsilon$ with $\|\epsilon\|_\infty \leq \rho$, $\mathrm{sign}(\tilde{w}_i + \epsilon_i) = \mathrm{sign}(\tilde{w}_i)$ for all $i \in \mathcal{C}(\rho)$.
    If $|\mathcal{C}(\rho)| = n$, SAM reduces to standard gradient descent on the latent weights.
  \end{proposition}
\end{myboxmain}

\vspace{-0.7ex}

The proof is available in \cref{app:pathology_proofs}. In practice,
as training progresses and latent weights grow in
magnitude, the fraction of confident coordinates $|\mathcal{C}(\rho)|/n$
approaches $1$, and the continuous SAM objective reduces to standard
gradient descent (see \cref{fig:vanishing} for an illustration).

\vspace{-1.8ex}

\subsection{Euclidean--Hamming Geometric Mismatch}

\vspace{-1.1ex}

The second failure comes from a geometric mismatch: flatness in the continuous
latent space does not imply robustness in the actual discrete operating domain.
Writing $d_H$ for the Hamming distance:

\vspace{-0.7ex}

\begin{myboxmain}
  \begin{proposition}[Metric Distortion]
    \label{prop:distortion}
    Map $\pi : \mathbb{R}^n \to \{-1,+1\}^n$ ($\pi(\tilde{w}) = \mathrm{sign}(\tilde{w})$) can exhibit arbitrarily
  large metric distortion: for any $D > 0$, there exist $\tilde{u}, \tilde{v} \in \mathbb{R}^n$
  with $\|\tilde{u} - \tilde{v}\|_2 < D^{-1}$ but $d_H(\pi(\tilde{u}), \pi(\tilde{v})) = n$,
  and conversely, $\tilde{u}', \tilde{v}'$ with $\|\tilde{u}' - \tilde{v}'\|_2 > D$
  but $d_H(\pi(\tilde{u}'), \pi(\tilde{v}')) = 0$.
  \end{proposition}
\end{myboxmain}

\vspace{-0.7ex}

The proof is available in \cref{app:pathology_proofs}. The distortion arises
because the sign function collapses continuous neighborhoods: two points
straddling the origin in every coordinate are $\varepsilon$-close in $\ell_2$
but maximally far in Hamming distance. Euclidean proximity is therefore neither
necessary nor sufficient for Hamming proximity. \cref{fig:distortion}
illustrates this phenomenon in $\mathbb{R}^2$.

\vspace{-1.1ex}

\begin{figure}[H]
  \centering

  \begin{minipage}[b]{0.38\textwidth}
    \centering
    \tikzexternaldisable
    \resizebox{\linewidth}{!}{
      \begin{tikzpicture}[>=Stealth, font=\small, transform shape, scale=0.87]
    \draw[->, thick] (0.1,0) -- (5.8,0) node[right]{$\tilde{w}_i$};

    \draw[thick, cSharp, dashed] (1.5,-0.55) -- (1.5,1.6);
    \node[cSharp, font=\footnotesize] at (1.5,-0.7) {$\mathrm{sign}$ threshold};

    \node[font=\footnotesize, RawSienna] at (0.6,-0.3) {$w_i = -1$};
    \node[font=\footnotesize, RawSienna] at (2.3,-0.3) {$w_i = +1$};
        
    \draw[very thick, cMid, fill=cMid!15, fill opacity=0.5] (3.5,-0.15) -- (4.9,-0.15) -- (4.9,0.15) -- (3.5,0.15) -- cycle;
    \draw[<->, cMid, thick] (3.5,-0.35) -- (4.9,-0.35) node[midway, below, font=\footnotesize]{$\rho{=}0.7$};
    
    \filldraw[cBlue] (4.2,0) circle (2.5pt);
    \node[cBlue, above=3pt] at (4.2,0) {$\tilde{w}_i{=}2.5$};

    \draw[->, thick, ForestGreen] (4.2,0.45) -- (4.2,0.9);

    \node[ForestGreen, font=\scriptsize, align=center] at (4.2,1.25) {$\mathrm{sign}(\tilde{w}_i{+}\epsilon_i) = {+}1$\\[1pt] \textbf{always} --- signal vanishes};

\end{tikzpicture}%
    }
    \tikzexternalenable

    \vspace{-1.2ex}

    \caption{\footnotesize{\textbf{The vanishing perturbation effect} (\cref{prop:vanishing}).  Continuous SAM perturbs $\tilde{w}_i = 2.5$ within a ball of radius $\rho = 0.7$. Since the ball never crosses the sign threshold, the sharpness signal vanishes.
      }}
    \label{fig:vanishing}
  \end{minipage}\hfill
  \begin{minipage}[b]{0.60\textwidth}
    \centering
    \tikzexternaldisable
    \resizebox{\linewidth}{!}{%
      \begin{tikzpicture}[>=Stealth, font=\small]

    \draw[->, thick, darkgray] (-0.2,0) -- (4.1,0) node[right, font=\footnotesize]{$\tilde{w}_1$};
    \draw[->, thick, darkgray] (0,-0.2) -- (0,2.3) node[above, font=\footnotesize, xshift=-12pt, yshift=-12pt]{$\tilde{w}_2$};
    \draw[dashed, darkgray!90] (2.1,-0.2) -- (2.1,2.3);
    \draw[dashed, darkgray!90] (-0.2,1.1) -- (3.9,1.1);

    \node[font=\footnotesize, darkgray] at (1.0,1.85) {$(-1,+1)$};
    \node[font=\footnotesize, darkgray] at (3.0,1.85) {$(+1,+1)$};
    \node[font=\footnotesize, darkgray] at (1.0,0.4) {$(-1,-1)$};
    \node[font=\footnotesize, darkgray] at (3.0,0.4) {$(+1,-1)$};

    \filldraw[cBlue] (2.25,1.25) circle (2.5pt) node[above right, font=\footnotesize]{$\tilde{u}$};
    \filldraw[cPurple] (1.95,0.95) circle (2.5pt) node[below left, font=\footnotesize]{$\tilde{v}$};

    \draw[>={Stealth[scale=0.8]}, <->, thin, cMid, shorten <=2.5pt, shorten >=2.5pt] (2.25,1.25) -- (1.95,0.95) node[midway, right, font=\footnotesize, xshift=1.0pt, yshift=-5.9pt]{$\|\cdot\|_2 \approx \varepsilon$};

    \begin{scope}[xshift=5.2cm]

        \draw[->, thick, darkgray] (-0.2,0) -- (4.3,0) node[right, font=\footnotesize]{$\tilde{w}_1$};
        \draw[->, thick, darkgray] (0,-0.2) -- (0,2.3) node[above, font=\footnotesize, xshift=-12pt, yshift=-12pt]{$\tilde{w}_2$};
        \draw[dashed, darkgray!90] (0.4,-0.2) -- (0.4,2.3);
        \draw[dashed, darkgray!90] (-0.2,0.4) -- (4.1,0.4);

        \node[font=\footnotesize, darkgray] at (1.3,1.5) {$(+1,+1)$};

        \filldraw[cBlue] (0.7,0.7) circle (2.5pt) node[below right, font=\footnotesize, yshift=2.5pt]{$\tilde{u}'$};
        \filldraw[cPurple] (3.65,2.0) circle (2.5pt) node[below right, font=\footnotesize]{$\tilde{v}'$};

        \draw[>={Stealth[scale=0.8]}, <->, thin, cMid, shorten <=2.5pt, shorten >=2.5pt] (0.7,0.7) -- (3.65,2.0) node[midway, left, font=\footnotesize, xshift=38.1pt, yshift=-7.8pt]{$\|\cdot\|_2 \gg 1$};

    \end{scope}
\end{tikzpicture}
      \tikzexternalenable
    }

    \vspace{-0.2ex}

    \caption{\footnotesize{\textbf{Euclidean--Hamming metric distortion} (\cref{prop:distortion}, $n{=}2$).
        \emph{Left:}~Two latent-weight vectors $\tilde{u}, \tilde{v}$ that are $\varepsilon$-close in $\mathbb{R}^2$ but straddle the $\mathrm{sign}$ boundary, producing $d_H = 2$.
        \emph{Right:}~Two vectors far apart but in the same quadrant, so $d_H = 0$. Euclidean proximity is neither necessary nor sufficient for Hamming proximity.}}
    \label{fig:distortion}
  \end{minipage}

  \vspace{-2.5ex}
\end{figure}

\section{Hamming-SAM: Formulation and Algorithm}
\label{sec:method}

\vspace{-1.2ex}

\subsection{The Hamming-Ball Min-Max Objective}

\vspace{-1.2ex}

Let $w \in \{-1,+1\}^n$ denote the $n$ Boolean weights in our model,
$\mathcal{L}(w)$ the population loss, and $\hat{\mathcal{L}}(w) \equiv
  \hat{\mathcal{L}}(w;\mathcal{D})$ the empirical loss on training data
$\mathcal{D}$ (we omit $\mathcal{D}$ when clear from context).
\cref{tab:notation} summarizes our main notations.
The Hamming-SAM
framework replaces the continuous $\ell_p$-ball with the $k$-bit Hamming ball
$\mathcal{B}_k(w) = \{w' \in \{-1,+1\}^n : d_H(w,w') \leq k\}$ and solves:

\vspace{-1.1ex}

\begin{equation}
  \min_{w \in \{-1,+1\}^n}\;
  \textcolor{BrickRed}{\mathcal{L}_{\mathrm{adv}}(w)}
  \;:=\;
  \textcolor{NavyBlue}{\max_{w' \in \mathcal{B}_k(w)}\;
    \hat{\mathcal{L}}(w';\mathcal{D})}.
  \label{eq:hamming_sam_objective}
\end{equation}

\vspace{-1.1ex}

The \textcolor{BrickRed}{adversarial loss} is defined as the
\textcolor{NavyBlue}{worst-case empirical loss over all configurations
  within $k$ bit-flips} of the current weights.
This objective is the native Boolean analogue of the
PAC-Bayes generalization bound that motivates the original continuous SAM
objective~\citep{foret2021sam}. We give two justifications and one equivalent reading below.
The detailed proofs are in \cref{app:objective_justifications}.
A structural comparison between continuous SAM and Hamming-SAM is given in
\cref{tab:foret_parallel}.

\begin{table}[t]
  \caption{\small{Continuous vs.\ Boolean landscape analysis.}}
  \label{tab:foret_parallel}
  \centering\small
  \scalebox{.801}{
    \setlength{\tabcolsep}{3.5pt}
    \begin{tabular}{llll}
      \toprule
      \textbf{Component}             & \textbf{Continuous SAM}                                & \textbf{Hamming-SAM (Boolean)}                & \textbf{Reference}           \\
      \midrule
      Parameter space                & $\mathbb{R}^n$                                         & $\{-1,+1\}^n$                                 & \cref{eq:hamming_sam_objective} \\
      Perturbation set               & $\ell_2$-ball $\{\epsilon: \|\epsilon\|_2 \leq \rho\}$ & Hamming ball $\mathcal{B}_k(w)$               & \cref{eq:hamming_sam_objective} \\
      PAC-Bayes stochastic posterior           & $\mathcal{N}(w, \rho^2 I)$                             & $\mathrm{BernoulliFlip}(w, q)$                & \cref{thm:discrete_foret}    \\
      PAC-Bayes complexity (data-indep.\ prior)     & $h(\|w\|_2^2/\rho^2)$ via Gaussian KL                  & $\mathrm{KL}(P\|P_0)$ via Bernoulli KL        & \cref{thm:discrete_foret}    \\
      PAC-Bayes complexity (data-dep.\ prior)  & $\|w-w_0\|_2^2/\sigma^2$ via Gaussian KL               & $d_H(\hat{w}, w_0)\log(1/p)$ via Bernoulli KL & \cref{lem:kl}                \\
      First-order                    & Gradient $\nabla\mathcal{L}$                           & Boolean Variation $\mathrm{BV}_i$             & \cref{sec:bv_main}           \\
      Second-order                   & Hessian $\nabla^2\mathcal{L}$                          & Interaction Hessian $H_{ij}$                  & \cref{eq:hessian}            \\
      Curvature                      & Hessian eigenvalues                                    & Forman-Ricci $\mathrm{Ric}_F$                 & \cref{prop:curvature}        \\
      Regularizer                    & Weight decay                                           & BDM complexity                                & \cref{app:bdm}               \\
      \bottomrule
    \end{tabular}
  }

  \vspace{-3.7ex}

\end{table}

\vspace{-1.1ex}

\paragraph{Justification I: PAC-Bayes upper bound (Boolean analogue of \cite{foret2021sam}).}
In continuous SAM, \cite{foret2021sam} use a Gaussian posterior $Q = \mathcal{N}(w, \rho^2 I)$ so that the expected loss is controlled by the worst case over the $\ell_2$-ball.
We repeat this argument on the Boolean hypercube using a Bernoulli bit-flip distribution.
Consider the distribution $P$ centered at $w$ that independently flips each bit with probability $q = k/(2n)$.
Splitting the expectation over $\mathcal{B}_k(w)$ and its complement,
and bounding the tail mass of the Binomial distribution, $\beta_k = \Pr[\mathrm{Bin}(n,q) > k] \leq e^{-k/6}$, via Chernoff \citep{chernoff1952measure, Boucheron2013concentration}:
\begin{equation}
  \mathbb{E}_{w'\sim P}[\hat{\mathcal{L}}(w')] \;\leq\; (1-\beta_k)\,\mathcal{L}_{\mathrm{adv}}(w) + \beta_k B,
  \label{eq:stochastic_split}
\end{equation}
where $B = \max_{w'}\hat{\mathcal{L}}(w')$.
Combining with the PAC-Bayes--McAllester inequality~\citep{mcallester2003pac}:

\vspace{-0.7ex}

\begin{myboxmain}
  \begin{theorem}[Discrete PAC-Bayes Sharpness Bound]
    \label{thm:discrete_foret}
    For any prior $P_0$ independent of the training set $\mathcal{S}$
  of size $m$, and any $\delta \in (0,1)$, with probability
  $\geq 1 - \delta$:
  \begin{equation}
    \mathbb{E}_{w'\sim P}\bigl[\mathcal{L}(w')\bigr]
    \;\leq\;
    \underbrace{\textcolor{BrickRed}{(1 - \beta_k)\,\mathcal{L}_{\mathrm{adv}}(w) + \beta_k B}}_{\textcolor{BrickRed}{\text{sharpness-aware empirical term}}}
    \;+\;
    \underbrace{\textcolor{NavyBlue}{\sqrt{\frac{\mathrm{KL}(P\|P_0)
      + \log(2\sqrt{m}/\delta)}{2m}}}}_{\textcolor{NavyBlue}{\text{complexity term}}},
    \label{eq:stochastic_bound}
  \end{equation}
  where $P = \mathrm{BernoulliFlip}(w, q)$ with $q = k/(2n)$,
  $\beta_k \leq e^{-k/6}$, and
  $B = \max_{w'}\hat{\mathcal{L}}(w')$.
  \end{theorem}
\end{myboxmain}

\vspace{-0.7ex}

The structure mirrors \cite{foret2021sam}: the \textcolor{BrickRed}{sharpness-aware
term} replaces $\max_{\|\epsilon\|\leq\rho}\hat{\mathcal{L}}(w{+}\epsilon)$,
with $\beta_k \leq e^{-k/6}$ replacing the Gaussian tail, while the
\textcolor{NavyBlue}{complexity term} uses a Bernoulli KL in place of a Gaussian KL.  The key difference is
that our bound operates directly on $\{-1,+1\}^n$.  Since $\beta_k$
is exponentially small for practical $k$, the bound is effectively
$\mathbb{E}_{w'\sim P}[\mathcal{L}(w')] \leq
\mathcal{L}_{\mathrm{adv}}(w) + \text{KL term}$, directly
motivating the minimization of $\mathcal{L}_{\mathrm{adv}}(w)$.  With a
uniform prior the complexity term is vacuous whenever $n > m$, and
\cref{sec:pac_bayes} replaces it with a data-dependent prior.
Full proof in \cref{app:objective_justifications}.

\vspace{-1.4ex}

\paragraph{Justification II: Compression bound.}
If the Hamming-SAM solution $\hat{w}$ satisfies
$\mathcal{L}_{\mathrm{adv}}(\hat{w}) \leq \epsilon$, \emph{every}
configuration in $\mathcal{B}_k(\hat{w})$ also has empirical loss $\leq
  \epsilon$.  A flat solution therefore covers an entire ball of
low-loss configurations at once, so we can use a PAC-Bayes posterior $Q_{\hat{w}}$
that is uniform over $\mathcal{B}_k(\hat{w})$, with a uniform prior:
\begin{equation}
  \mathbb{E}_{w' \sim Q_{\hat{w}}}\bigl[\mathcal{L}(w')\bigr]
  \leq \textcolor{BrickRed}{\epsilon}
  + \textcolor{NavyBlue}{\sqrt{\frac{n\log 2 - \log V_k + \log(2\sqrt{m}/\delta)}{2m}}},
  \label{eq:compression}
\end{equation}
where $V_k = \sum_{j=0}^{k}\binom{n}{j}$ and
$\mathrm{KL}(Q_{\hat{w}}\|P_0) = n\log 2 - \log V_k$ exactly.  The bound controls
the expected loss of a network drawn at random from the ball, that is, $\hat{w}$
after a random set of up to $k$ bit flips (\cref{app:deri_compression}).  The
\textcolor{BrickRed}{empirical term} is the worst-case loss over the ball, and the
\textcolor{NavyBlue}{complexity term} is reduced by
$\log V_k \approx k\log(n/k)$, because the bound pays for the whole ball
rather than for a single configuration.  With a uniform prior this reduction is
small relative to $n\log 2$, and the bound is vacuous when $n \gg m$.  The
data-dependent prior of \cref{sec:pac_bayes} yields a non-vacuous bound, in
which flatness again keeps the complexity term small.

\vspace{-1.4ex}

\paragraph{An equivalent perspective: distributionally robust optimization.}
The objective~\eqref{eq:hamming_sam_objective} can be written as a
DRO problem~\citep{rahimian2022dro}, $\mathcal{L}_{\mathrm{adv}}(w) = \max_{\mu \in
    \Delta(\mathcal{B}_k(w))} \mathbb{E}_{w'\sim\mu}[\hat{\mathcal{L}}(w')]$, where
$\Delta(\mathcal{B}_k(w))$ is the set of distributions supported on the
Hamming ball, and the maximum is attained at the Dirac measure on the
worst-case point.  This places Hamming-SAM within robust optimization, with the Hamming ball in
the role that a Wasserstein ambiguity set plays in the continuous
setting~\citep{kuhn2019wasserstein} (\cref{app:deri_dro}).

\begin{remark}[The role of $k$]
  In all three perspectives, $k$ controls a bias-sharpness tradeoff: \emph{(i)}~in the PAC-Bayes view,
  larger $k$ makes $\beta_k$ smaller (better concentration) but forces $\mathcal{L}_{\mathrm{adv}}$ higher;
  \emph{(ii)}~in the compression view, larger $k$ increases $\log V_k$ (more complexity reduction) but requires uniformly low loss over a larger set;
  \emph{(iii)}~in the DRO view, larger $k$ gives more conservative population coverage but a harder optimization problem.
\end{remark}

\subsection{Boolean Variation as a Discrete Gradient}
\label{sec:bv_main}

\vspace{-1ex}

The first-order discrete derivative of the loss at bit $i$ is the finite
difference
\begin{equation}
  \mathrm{BV}_i(w) \;=\; \hat{\mathcal{L}}(w^{(i)}) - \hat{\mathcal{L}}(w),
  \label{eq:bv}
\end{equation}
where $w^{(i)}$ is $w$ with bit $i$ negated, so it is the loss change caused by
flipping the current value of bit $i$.  It is defined on $\{-1,+1\}^n$
regardless of how it is obtained.  We use it throughout the theoretical
sections, where it keeps the set function, the interaction
Hessian~\eqref{eq:hessian} and the convergence analysis free of sign factors.

Any backend that estimates this signal can be used, such as BOLD or an STE. 
We mainly use BOLD \cite{nguyen2024bold} for convenience, since it yields all $n$ signals in a single
backward pass, and confirm the gains with STE.
The BOLD's chain rule (Theorem~3.12
of~\cite{nguyen2024bold}) enables backpropagation through deep Boolean
networks.  The BOLD variation of the loss with respect to
weight $w_i$ is
\begin{equation}
  f'_i(w) = \mathbf{xnor}\bigl(\delta(w_i \to \neg w_i),\,
  \delta\hat{\mathcal{L}}(w \to w^{(i)})\bigr),
  \label{eq:bold_variation}
\end{equation}
where $\delta(\cdot \to \cdot)$ denotes the direction of change.
In the $\{-1,+1\}$ conversion, where $\mathbf{xnor}$ corresponds to
multiplication, this evaluates to
$f'_i(w) = -w_i\,\mathrm{BV}_i(w) = \hat{\mathcal{L}}(w|_{w_i=+1}) - \hat{\mathcal{L}}(w|_{w_i=-1})$,
the partial derivative of pseudo-Boolean function
theory~\citep{crama2011boolean, odonnell2014analysis}.  It compares the two
values of bit $i$ rather than flipping it. Thus, it plays the role of a
gradient: a bit is flipped when $f'_i \cdot w_i > 0$, equivalently
$\mathrm{BV}_i < 0$, which moves $w_i$ against
the sign of $f'_i$ and decreases the loss.

\textbf{Exact values and first-order estimates.}  Evaluating \cref{eq:bv} exactly for
every bit would require $n+1$ forward passes, one per flipped configuration.
A backward pass instead returns all $n$ signals at once, as a
first-order estimate: flipping bit $i$ moves $w_i$ by $-2w_i$, so a Taylor
expansion gives $\mathrm{BV}_i(w) \approx -2w_i\,\partial\hat{\mathcal{L}}/\partial w_i$.
The positive factor $2$ is immaterial to \cref{alg:hamming_sam}, since \textcolor{NavyBlue}{\textbf{Phase~1}} selects
bits by rank and \textcolor{BrickRed}{\textbf{Phase~2}} by sign, and both are invariant to positive rescaling.
The analysis below is stated for the exact quantity \cref{eq:bv}.  \cref{app:ascent}
quantifies how an inexact oracle affects the ascent guarantee.

\vspace{-1ex}

\subsection{The Hamming-SAM Algorithm}

\vspace{-0.9ex}

\begin{figure}[t]

  \centering
  \tikzexternaldisable

  \begin{tikzpicture}[>=Stealth, font=\small, scale=0.82,
    vertex/.style={circle, draw, thick, minimum size=10.4pt, inner sep=0pt, font=\fontsize{4}{5}\selectfont\bfseries},
    hedge/.style={thick, cGray!35},
    lbl/.style={font=\fontsize{5.5}{6.5}\selectfont, text opacity=1, inner sep=1pt, rounded corners=1pt}]
  \def\dx{1.65} \def\dy{1.65} \def\sx{0.638} \def\sy{0.563}

  \cubecoords{a}{0}
  \node[anchor=south, font=\bfseries\footnotesize] at ({0+\dx/2+\sx/2},{\dy+\sy+0.35}) {(a) Loss landscape \& $\mathcal{B}_1(w^t)$};
  \cubeedges{a}
  \draw[very thick, cBlue!60] (a-111)--(a-011); \draw[very thick, cBlue!60] (a-111)--(a-101); \draw[very thick, cBlue!60] (a-111)--(a-110);
  \node[vertex, fill=cMid!65] at (a-000) {.45}; \node[vertex, fill=cSharp!80] at (a-100) {.62};
  \node[vertex, fill=cMid!45] at (a-010) {.35}; \node[vertex, fill=cSharp!60] at (a-110) {.55};
  \node[vertex, fill=cFlat!30] at (a-001) {.28}; \node[vertex, fill=cFlat!65] at (a-101) {.20};
  \node[vertex, fill=cFlat!50] at (a-011) {.22};
  \node[vertex, fill=cMid!25, line width=1.2pt, draw=cYellow!90!black] at (a-111) {.30};
  \node[lbl, cBlue, xshift=1.5pt] at ($(a-111)+(0.39,0.19)$) {$w^t$};
  \node[lbl, cBlue, anchor=west, font=\fontsize{4.5}{5.5}\selectfont, xshift=19.5pt, yshift=-5.5pt] at ($(a-111)!0.45!(a-110)+(-1.0,0)$) {$\mathrm{BV}_2{=}.25$};
  \node[lbl, cBlue, anchor=east, font=\fontsize{4.5}{5.5}\selectfont, xshift=9.5pt, yshift=2.5pt] at ($(a-111)!0.45!(a-011)+(0.24,0.15)$) {$\mathrm{BV}_1{=}{-}.08$};
  \node[lbl, cBlue, anchor=west, font=\fontsize{4.5}{5.5}\selectfont, yshift=-2.5pt] at ($(a-111)!0.5!(a-101)+(0.02,-0.39)$) {$\mathrm{BV}_3{=}{-}.10$};

  \def\boff{5.0} \cubecoords{b}{\boff}
  \node[anchor=south, font=\bfseries\footnotesize] at ({\boff+\dx/2+\sx/2},{\dy+\sy+0.35}) {(b) \textcolor{NavyBlue}{\textbf{Phase~1}}: Ascent ($k{=}1$)};
  \cubeedges{b}
  \node[vertex, fill=cMid!65] at (b-000) {.45}; \node[vertex, fill=cSharp!80] at (b-100) {.62};
  \node[vertex, fill=cMid!45] at (b-010) {.35};
  \node[vertex, fill=cPurple!35, line width=1.2pt, draw=cPurple] (n-b-110) at (b-110) {.55};
  \node[vertex, fill=cFlat!30] at (b-001) {.28}; \node[vertex, fill=cFlat!65] at (b-101) {.20};
  \node[vertex, fill=cFlat!50] at (b-011) {.22};
  \node[vertex, fill=cMid!25, line width=1.2pt, draw=cYellow!90!black] (n-b-111) at (b-111) {.30};
  \draw[->, cPurple, line width=0.9pt] (n-b-111) -- (n-b-110);
  \node[lbl, cBlue, xshift=1.5pt] at ($(b-111)+(0.39,0.19)$) {$w^t$};
  \node[lbl, cPurple] at ($(b-110)+(-0.36,-0.11)$) {$\hat{w}$};
  \node[cPurple, font=\fontsize{5}{6}\selectfont, align=center] at ({\boff+\dx/2+\sx/2},-0.5) {$\arg\max_i \mathrm{BV}_i = $ bit 2};

  \def\coff{10.0} \cubecoords{c}{\coff}
  \node[anchor=south, font=\bfseries\footnotesize] at ({\coff+\dx/2+\sx/2},{\dy+\sy+0.35}) {(c) \textcolor{BrickRed}{\textbf{Phase~2}}: Descent};
  \cubeedges{c}
  \draw[thick, dashed, cPurple!60] (c-110)--(c-010); \draw[thick, dashed, cPurple!60] (c-110)--(c-100);
  \node[vertex, fill=cMid!65] at (c-000) {.45}; \node[vertex, fill=cSharp!80] at (c-100) {.62};
  \node[vertex, fill=cMid!45] at (c-010) {.35};
  \node[vertex, fill=cPurple!20, line width=0.8pt, draw=cPurple!50] at (c-110) {.55};
  \node[vertex, fill=cFlat!30] at (c-001) {.28}; \node[vertex, fill=cFlat!65] at (c-101) {.20};
  \node[vertex, fill=cFlat!50, line width=1.2pt, draw=cFlat!70!black] (n-c-011) at (c-011) {.22};
  \node[vertex, fill=cMid!25, line width=1.2pt, draw=cYellow!90!black] (n-c-111) at (c-111) {.30};
  \draw[->, cFlat!70!black, line width=0.9pt] (n-c-111) -- (n-c-011);
  \node[lbl, cBlue, xshift=1.5pt] at ($(c-111)+(0.39,0.19)$) {$w^t$};
  \node[lbl, cPurple!80] at ($(c-110)+(-0.36,-0.11)$) {$\hat{w}$};
  \node[lbl, cFlat!70!black, xshift=2.5pt] at ($(c-011)+(0.49,0.26)$) {$w^{t+1}$};
  \node[cFlat!60!black, font=\fontsize{5}{6}\selectfont, align=center] at ({\coff+\dx/2+\sx/2},-0.62)
    {$\mathrm{BV}_1(\hat{w}){=}{-}.20,\ \mathrm{BV}_3(\hat{w}){=}{+}.07$\\[-0pt]
     $\arg\min_{i \notin \mathcal{I}} \mathrm{BV}_i(\hat{w}) = $ bit 1};
\end{tikzpicture}

  \tikzexternalenable

  \vspace{-2.0ex}

  \caption{\footnotesize{\textbf{Hamming-SAM walkthrough on $\{-1,+1\}^3$ with $k{=}1$.}  Node values show loss $\hat{\mathcal{L}}$ (\textcolor{cFlat}{green}~=~low, \textcolor{cSharp}{red}~=~high).
  \textbf{(a)}~Current weight $w^t{=}{\scriptsize (+1,+1,+1)}$ (loss $0.30$); blue edges show $\mathcal{B}_1(w^t)$ and \textcolor{cBlue!60}{Boolean Variations}.
  \textbf{(b)}~\textcolor{NavyBlue}{\textbf{Phase~1}} selects bit~2 ($\mathrm{BV}{=}0.25$, the only increase) as the adversarial point $\hat{w}{=}{\scriptsize (+1,-1,+1)}$.
  \textbf{(c)}~\textcolor{BrickRed}{\textbf{Phase~2}} evaluates BV at $\hat{w}$ and flips bit~1 ($\mathrm{BV}_1(\hat{w}){=}{-}0.20$), yielding $w^{t+1}{=}{\scriptsize (-1,+1,+1)}$ with loss $0.22$.  Base-point descent would instead flip bit~3, reaching a point whose one-bit worst case is $0.62$ rather than $0.35$.}}

  \label{fig:hamming_sam_walkthrough}

  \vspace{-3.3ex}

\end{figure}
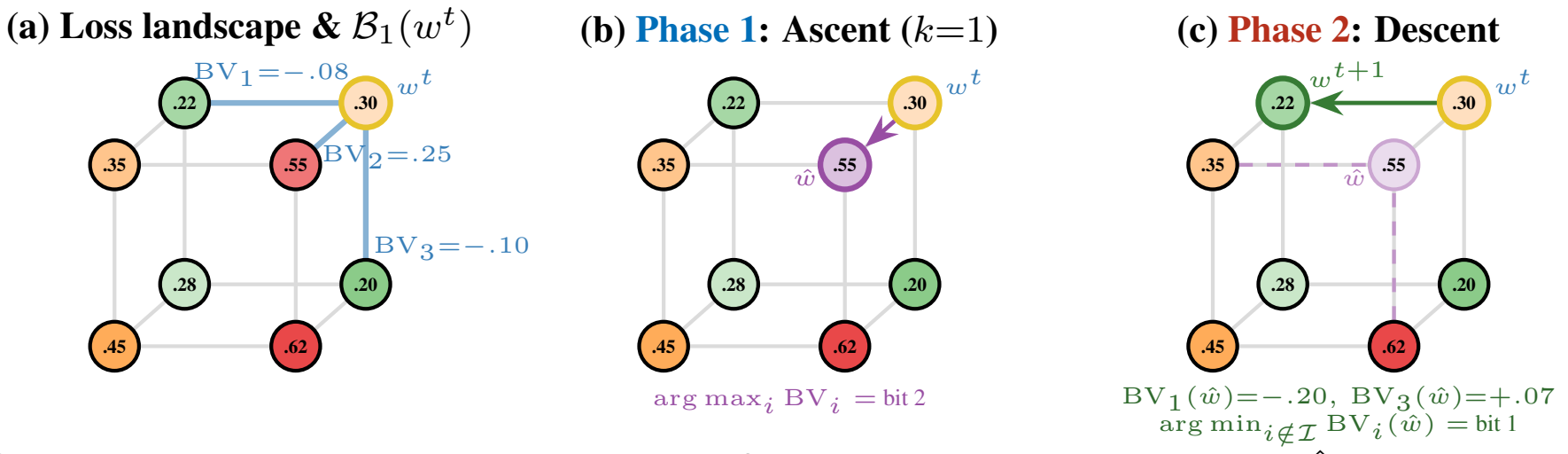

\begin{minipage}{0.45\textwidth} %
  \cref{alg:hamming_sam} illustrates our optimization procedure (example shown in \cref{fig:hamming_sam_walkthrough}).
  \textcolor{NavyBlue}{\textbf{Phase~1}} greedily identifies the $k$ bits that most increase the loss, constructing an adversarial point in $\mathcal{B}_k(w^t)$.
  \textcolor{BrickRed}{\textbf{Phase~2}} evaluates the Boolean Variation at this \emph{worst-case} point and flips the single bit that most reduces the loss, updating the original weights rather than the adversarial ones.
  Computing descent at this worst-case point ensures robustness against sharp loss landscapes.
  Using a backward pass (e.g., BOLD, \cref{sec:bold_background}), each
  \textcolor{Mulberry}{``Compute $\mathrm{BV}_i$'' loop} requires just one
  backward pass. The total per-step cost is therefore two backward passes (at
  $w^t$ and $\hat{w}$) (see details in \cref{app:compute_cost}).

\end{minipage}%
\hfill %
\begin{minipage}{0.51\textwidth}
  \begin{algorithm}[H] %
    \caption{\small Hamming-SAM Optimization Step}
    \label{alg:hamming_sam}

    \scalebox{0.8}{%
      \begin{minipage}{1.35\linewidth}

        \KwIn{Current weights $w^t \in \{-1,+1\}^n$, Hamming radius $k$}
        \BlankLine
        \textcolor{NavyBlue}{\textbf{Phase 1: Greedy Ascent (Adversarial Probing)}}\\
        \textcolor{Mulberry}{
          \For(\tcp*[h]{Done by single backward pass}){each bit $i = 1,\ldots,n$}{
            Compute $\mathrm{BV}_i(w^t)$\;
          }
        }
        $\mathcal{I} \leftarrow \arg\!\operatorname{top\text{-}k}\{\mathrm{BV}_i(w^t) : \mathrm{BV}_i(w^t) > 0\}$ \tcp*[h]{$\le k$ indices with largest positive BV}\\
        $\hat{w} \leftarrow w^t$ with bits in $\mathcal{I}$ flipped \tcp*[h]{Adversarial point $\hat{w} \in \mathcal{B}_k(w^t)$}

        \BlankLine
        \textcolor{BrickRed}{\textbf{Phase 2: Sharpness-Aware Descent (Weight Update)}}\\
        \textcolor{Mulberry}{
          \For(\tcp*[h]{Done by single backward pass}){each bit $i = 1,\ldots,n$}{
            Compute $\mathrm{BV}_i(\hat{w})$\;
          }
        }
        $i^* \leftarrow \arg\min_{i \notin \mathcal{I}}\; \mathrm{BV}_i(\hat{w})$ \tcp*[h]{Exclude ascent bits}

        \eIf{$\mathrm{BV}_{i^*}(\hat{w}) < 0$}{
          $w^{t+1} \leftarrow w^t$ with bit $i^*$ flipped\;
        }{
          $w^{t+1} \leftarrow w^t$\;
        }

      \end{minipage}%
    }
  \end{algorithm}
\end{minipage}

\begin{remark}[Practical Phase~2]
  \label{rem:practical_descent}
  With the BOLD backend, \textcolor{BrickRed}{\textbf{Phase~2}} feeds the backward signal at the adversarial point,
an estimate of $f'_i(\hat{w}^t)$, into the
accumulator of~\cite{nguyen2024bold} for every bit, which flips bit $i$ when
the accumulated evidence for $f'_i \cdot w^t_i > 0$ exceeds an adaptive
threshold (\cref{app:accumulator_threshold}).
For $i \notin \mathcal{I}$, $f'_i(\hat{w}^t)\,w^t_i = -\mathrm{BV}_i(\hat{w}^t)$,
which is positive exactly when the descent test $\mathrm{BV}_i(\hat{w}^t) < 0$
of \cref{alg:hamming_sam} holds.
For an
ascent bit, the signal predicts that restoring it lowers the loss, so the step
moves its accumulator away from the flip threshold (\cref{app:accumulator_threshold}).
\end{remark}

\vspace{-0.2ex}

\begin{remark}[Connection to local entropy]
  The Hamming-SAM objective~\cref{eq:hamming_sam_objective} is related to but distinct from the local entropy of \cite{baldassi2020shaping},
  which maximizes $\log|\{w' \in \mathcal{B}_k(w) : \mathcal{L}(w') \leq \theta\}|$, with a threshold $\theta$.
  Hamming-SAM minimizes the worst-case loss (a min-max objective), while local entropy maximizes the volume of low-loss configurations (a counting objective).
  Both encourage convergence to dense clusters of good solutions, but Hamming-SAM provides a deterministic, worst-case guarantee.
\end{remark}

\vspace{-2.2ex}

\subsection{Choosing the Hamming Radius $k$}
\label{sec:choosing_k}

\vspace{-1.1ex}

$k$ is the central hyperparameter, controlling the adversarial probe strength
and the generalization bound tightness.  The compression bound \cref{eq:compression}
decomposes into a sharpness term $\mathcal{L}_{\mathrm{adv}}(\hat{w};k)$,
increasing in $k$, plus a complexity term, decreasing in $k$ with reduction
$\log V_k \approx k\log(n/k)$ for $k \ll n$.

\begin{definition}[$(L,\sigma)$-Smooth Boolean Landscape]
  \label{def:smooth_landscape}
  The loss is $(L,\sigma)$-smooth if
  $\mathcal{L}_{\mathrm{adv}}(w;k) - \hat{\mathcal{L}}(w) \leq L\cdot k^\sigma$
  for all $w$ and $k \geq 1$, where $\sigma > 0$ controls how fast the worst-case
  loss grows with the radius: $\sigma < 1$ indicates diminishing returns, and
  $\sigma > 1$ that each extra flip adds more than the previous one.
\end{definition}

\vspace{-1.5ex}

\begin{myboxmain}
\begin{proposition}[Optimal $k$]
  \label{prop:optimal_k}
  Under $(L,\sigma)$-smoothness with $k \ll n$ and $\sigma > 1$, the compression
  bound is approximately minimized at
  $k^* \approx \bigl(\log n \,/\,(2L\sigma\sqrt{2mn\log 2})\bigr)^{1/(\sigma-1)}$.
\end{proposition}
\end{myboxmain}

\vspace{-1.0ex}

The case $\sigma > 1$ arises when flipping bits together costs more than the sum
of their individual effects.  For $\sigma \leq 1$ the bound has no interior
minimum and is minimized at the boundary.  The full statement for all $\sigma$
and its proof are in \cref{app:choosing_k}.
Since $\log V_k \approx n\,\alpha\log(1/\alpha)$ for $k = \alpha n$, a fixed
fraction $\alpha$ removes the same share of the description length at every
model size, which motivates parametrizing the radius as
$k = \lfloor\alpha \cdot n\rfloor$ across model sizes.
We use $k = 500$ in most cases ($\alpha \approx 10^{-5}$ to $10^{-3}$), and the
ablation of \cref{sec:exp_landscape} shows a broad plateau in $k$.

\vspace{-1.5ex}

\section{Theoretical Analysis}
\label{sec:theory}

\vspace{-1.2ex}

\subsection{Approximation Guarantee for the Ascent Step}
\label{sec:ascent}

\vspace{-1.1ex}

A key requirement is that the greedy ascent identifies a
near-worst-case configuration within $\mathcal{B}_k(w)$. In continuous SAM, the
ascent is justified as a first-order Taylor approximation. In the discrete
domain, finding the adversarial point $\hat{w}$ is a cardinality-constrained
maximization of the non-negative risk-increase set function $f(S) =
  \hat{\mathcal{L}}(w^t \oplus \mathbf{1}_S) - \hat{\mathcal{L}}(w^t)$ for $|S|
  \leq k$, where $w^t \oplus \mathbf{1}_S$ denotes flipping the bits indexed by
the set $S$. The quality of the greedy search depends on the approximate
submodularity of this function. We use the framework of
\cite{bian2017guarantees}:

\begin{definition}[Submodularity Ratio and Curvature \citep{das2011submodular, bian2017guarantees}]
  \label{def:submod_ratio}
  Let $\rho_j(A) = f(A \cup \{j\}) - f(A)$.  The submodularity ratio
  $\gamma_k \in (0,1]$ is the largest $\gamma$ such that
  $\sum_{j \in L}\rho_j(S) \geq \gamma\,[f(S \cup L) - f(S)]$ for all disjoint
  $L, S$ with $|L| \leq k$.  The generalized curvature $\alpha \in [0,1]$ is the
  smallest scalar such that
  $\rho_i\bigl((S\setminus\{i\}) \cup \Omega\bigr) \geq (1-\alpha)\,\rho_i(S\setminus\{i\})$
  for all $S, \Omega$ and $i \in S \setminus \Omega$.
\end{definition}

\vspace{-1.1ex}

The submodularity ratio quantifies how close $f$ is to submodular ($\gamma = 1$
for submodular functions).  The curvature $\alpha$ measures how much the
marginal gain of an element can shrink when other elements are added: $\alpha = 0$
for modular functions, and $\alpha = 1$ allows gains to vanish entirely (the
submodular worst case, recovering the classical $(1-1/e)$
bound)~\citep{nemhauser1978submodular}.

\vspace{-0.9ex}

\begin{myboxmain}
  \begin{theorem}[Sequential Greedy Ascent Guarantee]
  \label{thm:ascent}
  Let $\mathcal{V}_+ = \{i \in [n] : \mathrm{BV}_i(w^t) > 0\}$ and suppose $f$,
  restricted to subsets of $\mathcal{V}_+$, is non-negative, non-decreasing
  (monotone), and has submodularity ratio $\gamma_k$ and curvature $\alpha$
  (\cref{def:submod_ratio}).  Let $\hat{S}$ be the set of $k$ indices selected by
  the sequential greedy (which re-evaluates marginal gains after each selection)
  and $S^* = \arg\max_{S \subseteq \mathcal{V}_+,\,|S| \leq k} f(S)$.  Then:
  $f(\hat{S}) \;\geq\; \frac{1}{\alpha}\bigl[1 - e^{-\gamma_k\alpha}\bigr]\,f(S^*)$.
  When $\gamma_k{=}1$, $\alpha{=}1$ (submodular), this recovers the classical $(1{-}1/e)$ bound.
\end{theorem}
\end{myboxmain}
\vspace{-0.8ex}

The full proof is available in \cref{app:ascent}.

\begin{remark}[Monotonicity]
  \label{rem:monotonicity}
  Every bit in $\mathcal{V}_+$ increases the loss when flipped alone.  Flipping
  other bits first can shrink this gain through pairwise interactions, so $f$ is
  monotone on $\mathcal{V}_+$ as long as each gain exceeds what interactions can
  remove.  \cref{app:ascent} makes this precise in terms of the
  interaction Hessian introduced below.
\end{remark}

\vspace{-2.2ex}

\paragraph{Connection to the discrete interaction Hessian.}
The quality of the greedy approximation depends on the landscape curvature,
characterized by the discrete interaction Hessian, the second-order flip
difference of pseudo-Boolean function
theory~\citep{crama2011boolean, odonnell2014analysis}:

\vspace{-1.1ex}

\begin{equation}
  H_{ij}(w) = \hat{\mathcal{L}}(w^{(ij)}) - \hat{\mathcal{L}}(w^{(i)}) - \hat{\mathcal{L}}(w^{(j)}) + \hat{\mathcal{L}}(w).
  \label{eq:hessian}
\end{equation}

\vspace{-1.1ex}

This quantity measures the pairwise interaction between bit-flips $i$ and $j$.  If $H_{ij} = 0$, the two bits contribute independently to the loss change.  If $H_{ij} < 0$, flipping both gives \emph{less} increase than the sum of individual flips (diminishing returns), and if $H_{ij} > 0$, it gives \emph{more} (the flips reinforce each other).
See \cref{sec:hessian_app} for more details.

The connection to the ascent guarantee is direct. The second-order discrete
derivative of $f$ at context $S$ satisfies $\Delta_{ij}f(S) = H_{ij}(w^t \oplus
  \mathbf{1}_S)$ (\cref{app:ascent}).
   Therefore, $f$ is submodular, in which case $\gamma_k = 1$, if and only if
$H_{ij}(w^t \oplus \mathbf{1}_S) \leq 0$ for all $S$ and $i, j \notin S$.
   When the Hessian has positive entries, the submodularity ratio
$\gamma_k$ decreases below $1$ and the ascent guarantee degrades gracefully:
the bound $(1/\alpha)[1-e^{-\gamma_k \alpha}]$ is continuous and positive for
any $\gamma_k > 0$.

\vspace{-1.1ex}

\paragraph{The one-shot rule used in practice.}
\cref{alg:hamming_sam} selects the $k$ bits in a single backward pass rather than
re-evaluating marginal gains $k$ times.  Expanding $f$ in the M\"obius basis of
pseudo-Boolean analysis isolates the cost of this choice: the top-$k$ rule
already maximizes the first-order term exactly, so the deficit is carried
entirely by pairwise and higher-order interactions.

\vspace{-0.7ex}

\begin{myboxmain}
\begin{theorem}[Top-$k$ Ascent Guarantee]
  \label{thm:ascent_topk}
  Let $\hat{S}_{\mathrm{tk}}$ be the top-$k$ selection of \cref{alg:hamming_sam},
  restricted to bits with $\mathrm{BV}_i > 0$.  Suppose $|H_{ij}(w)| \leq h$ for
  all $w$ and $i \neq j$, and the third- and higher-order interaction residual
  satisfies $|R(S)| \leq R_{\max}$ for all $|S| \leq k$.  Then
  $f(\hat{S}_{\mathrm{tk}}) \;\geq\; f(S^*) - \textcolor{BrickRed}{k(k{-}1)h}
  - \textcolor{Mulberry}{2R_{\max}}$.
\end{theorem}
\end{myboxmain}

\vspace{-0.7ex}

Proof and empirical instantiation are in \cref{app:ascent}.

\vspace{-1.1ex}

\paragraph{Roles of the two guarantees.}
\cref{thm:ascent,thm:ascent_topk} are complementary rather than nested, and
both are needed.  They constrain the same object through different features of
it.  By the equivalence above, submodularity is a condition on the \emph{sign}
structure of the interaction Hessian, whereas \cref{thm:ascent_topk} depends
only on its \emph{magnitude}.  Neither assumption set implies the other.
\cref{thm:ascent} is multiplicative and scale-free, certifying a constant
fraction of the optimum however large $f(S^*)$ is, but it requires monotonicity
and degrades as $\gamma_k \to 0$.  \cref{thm:ascent_topk} does not require
monotonicity or submodularity, but is informative only while $k(k{-}1)h$ stays
small relative to $f(S^*)$.  The ascent is therefore near-optimal whenever
\emph{either} the landscape is approximately submodular \emph{or} its pairwise
interactions are weak.  \cref{app:gamma_k} reports sampled submodularity
ratios close to one throughout training, consistent with the first
condition.  The two also differ in cost.  The sequential greedy is the
strongest ascent available with unlimited compute, at $k$ backward passes per
step, while \cref{alg:hamming_sam} runs the one-shot rule in a single pass.  The
gap between them, measured in \cref{app:sequential_greedy}, is the price of
that tractability.

\vspace{-1.5ex}

\subsection{Generalization via Data-Dependent Priors}
\label{sec:pac_bayes}

\vspace{-0.9ex}

The PAC-Bayes justification in \cref{sec:method} (\cref{thm:discrete_foret})
uses a uniform prior, yielding a KL divergence of $n\log 2$ that is vacuous for
any model with $n > m$. Tighter bounds are possible with data-dependent
Bernoulli priors~\citep{dziugaite2021role}: split $\mathcal{D} = \mathcal{S}_0
  \sqcup \mathcal{S}_1$, train a reference $w_0$ on $\mathcal{S}_0$, and define
the prior as a product distribution that flips each bit of $w_0$ independently
with probability $p$.

\begin{lemma}[KL for Bernoulli Bit-Flip Prior]\label{lem:kl}
  For the deterministic posterior $Q = \delta_{\hat{w}}$ and prior $P$ with
  flip probability $p$: $\mathrm{KL}(Q\|P) = d\log(1/p) +
    (n{-}d)\log(1/(1{-}p))$, where $d = d_H(\hat{w}, w_0)$.
\end{lemma}

\noindent The resulting McAllester-style bound scales as $\mathcal{O}(d\log(1/p))$
rather than $\mathcal{O}(n\log 2)$, and is non-vacuous when $d \ll n$.  The
advantage of Boolean weights is that this KL is \emph{exact and
  closed-form}.  Hamming-SAM fits this framework well.
Flat basins are \emph{stable} under the distribution shift from
$\mathcal{S}_0$ to $\mathcal{S}_1$, requiring fewer bit-flips to adapt
and keeping $d$ small, whereas sharp minima must escape the basin
and incur a large Hamming distance.  Lower sharpness $B_k$ thus yields
smaller adaptation distance $d$ and a tighter bound.  Full derivation
in \cref{app:pac_bayes}.

\vspace{-1.0ex}

\subsection{Convergence Analysis}
\label{sec:convergence}

\vspace{-0.5ex}

We present two convergence results for the empirical loss $\hat{\mathcal{L}}(w^t)$.
\cite{khanh2024sam} recently established convergence of continuous SAM
under diminishing stepsizes and Lipschitz gradients.  Our discrete
setting differs in that \emph{(i)}~$\hat{\mathcal{L}}(w^t)$ decreases
\emph{monotonically} at each accepted step, and
\emph{(ii)}~the convergence condition involves the discrete Hessian
rather than Lipschitz constants.  \cref{rem:ladv} additionally bounds the
adversarial objective $\mathcal{L}_{\mathrm{adv}}(w^t)$ along the trajectory.

\begin{definition}[$\eta$-Discrete Stationarity]
  \label{def:stationary}
  A configuration $w$ is $\eta$-stationary if $\mathrm{BV}_i(w) \geq -\eta$ for all $i$, so that no single bit-flip can reduce $\hat{\mathcal{L}}(w)$ by more than $\eta$.
\end{definition}

\vspace{-1.0ex}

\begin{myboxmain}
\begin{theorem}[Finite-Time Convergence (Base-Point Descent)]
  \label{thm:convergence}
  Suppose $\hat{\mathcal{L}}:\{-1,+1\}^n \to [0,B]$.
  Consider a variant of \cref{alg:hamming_sam} where Phase~2 evaluates $\mathrm{BV}_i$ at $w^t$ (rather than at $\hat{w}$),
  selecting $i^* = \arg\min_i\,\mathrm{BV}_i(w^t)$ and flipping only when $\mathrm{BV}_{i^*}(w^t) < -\eta$.
  Then this variant converges to an $\eta$-stationary point in at most $T \leq \lfloor B/\eta \rfloor$ iterations.
\end{theorem}
\end{myboxmain}

\vspace{-1.0ex}

The proof is in \cref{app:convergence}.
We now show that the full SAM-style algorithm also converges, provided the
discrete Hessian is bounded.

\vspace{-0.7ex}

\begin{myboxmain}
\begin{theorem}[Finite-Time Convergence (SAM-Style Descent)]
  \label{thm:convergence_sam}
  Suppose $\hat{\mathcal{L}}:\{-1,+1\}^n \to [0,B]$ and $|H_{ij}(w)| \leq h$ for all $w$ and $i \neq j$.
  In ~\cref{alg:hamming_sam}, suppose Phase~2 selects $i^* \notin \mathcal{I}$ and accepts the flip only when $\mathrm{BV}_{i^*}(\hat{w}) < -\eta$ with $\eta > k\cdot h$.
  Then each accepted step satisfies
  $\hat{\mathcal{L}}(w^{t+1}) \;\leq\; \hat{\mathcal{L}}(w^t) - (\eta - kh)$,
  and convergence occurs in at most $T \leq \lfloor B/(\eta - kh) \rfloor$ iterations.
\end{theorem}
\end{myboxmain}

\vspace{-0.7ex}

The key idea is a telescoping identity: since $\hat{w}$ and $w^t$ differ in the
$k$ ascent bits $\mathcal{I}$, restoring these bits one at a time yields
$\mathrm{BV}_{i^*}(w^t) = \mathrm{BV}_{i^*}(\hat{w}) + \sum_{l=1}^{k}
  H_{i^*,j_l}(u^{l-1})$, where each $|H_{i^*,j_l}| \leq h$. Full proof in
\cref{app:convergence}.

\begin{remark}[Controlling the adversarial objective]
  \label{rem:ladv}
  The expansion behind \cref{thm:ascent_topk} also bounds the sharpness itself:
  $\mathcal{L}_{\mathrm{adv}}(w) - \hat{\mathcal{L}}(w) \leq
    \textcolor{NavyBlue}{k\,B_1(w)} + \textcolor{BrickRed}{\binom{k}{2}h}
    + \textcolor{Mulberry}{R_{\max}}$, where $B_1(w) = \max_i \max(\mathrm{BV}_i(w),0)$
  is the one-bit sharpness, available from a single backward pass.  Since
  \cref{thm:convergence_sam} drives $\hat{\mathcal{L}}(w^t)$ down
  monotonically, $\mathcal{L}_{\mathrm{adv}}(w^t)$ stays within an explicit
  and observable band above it, and full min-max convergence reduces to
  controlling $B_1$, which Phase~2 targets directly.  At $k = 1$ the bound is
  an identity.
\end{remark}

\begin{remark}[The condition $\eta > kh$ as a design principle]
  \label{rem:eta_kh}
  This condition admits a natural algorithmic interpretation.
  The descent signal must exceed the total interaction
between $i^*$ and the $k$ ascent bits, and the practitioner enforces this
by choosing $k$ small (default $k/n \approx 10^{-4}$) or $\eta$ large.
Structurally, $h$ is moderate because bits in different neurons of the same
layer perturb disjoint pre-activations and can interact only through
downstream layers.  With threshold activations a single flip shifts a
pre-activation by exactly $\pm 2$, so whenever either margin absorbs the
shift, $H_{ij} = 0$ exactly.  With the real-valued activations we train, the
interaction is nonzero but collapses during training
(\cref{sec:exp_landscape}).
The condition also reinforces itself.  The algorithm tends to
converge to low-$h$ regions, which have high
$\gamma_k$ (\cref{thm:ascent}), so the discrete Hessian is the
\emph{unifying quantity} governing both ascent quality and convergence
rate.
\end{remark}

\vspace{-1.5ex}

\paragraph{Multi-bit descent and the accumulator.}
In practice, an accumulator such as BOLD's may flip $m$ bits simultaneously. The
following proposition (proof in \cref{app:accumulator_threshold}) quantifies
when this is safe:

\vspace{-0.7ex}

\begin{myboxmain}
\begin{proposition}[Multi-Bit Descent under Bounded Hessian]
  \label{prop:multi_bit}
  Under $|H_{ij}(w)| \leq h$, if $m$ bits are flipped simultaneously,
  each with $\mathrm{BV}_{i_l}(w^t) < -\delta$ and $\delta > (m{-}1)h/2$,
  then $\hat{\mathcal{L}}(w^t \oplus \mathbf{1}_S) - \hat{\mathcal{L}}(w^t) \leq -m\delta + \binom{m}{2}h < 0$.
\end{proposition}
\end{myboxmain}

\vspace{-0.7ex}

\noindent Combined with \cref{thm:convergence_sam},
a multi-bit step decreases the loss when $\eta > (k + \tfrac{m-1}{2})h$, a three-way tradeoff between probe strength ($k$),
optimizer aggressiveness ($m$), and landscape curvature ($h$).
The condition applies when every flipped bit satisfies the premise of \cref{prop:multi_bit}.
The accumulator used in practice selects bits by accumulated signal, and \cref{app:accumulator_threshold} analyzes its effective threshold.

\vspace{-1.2ex}

\section{Diagnostic Extensions}
\label{sec:perspectives}

\vspace{-0.8ex}

Two further constructions are natural on the hypercube.  We develop them in the
appendix as a diagnostic and as an optional extension, not as part of the
method or of its guarantees.

\vspace{-1.3ex}

\paragraph{Forman-Ricci curvature.}
Weighting the Hamming graph by the loss and by the Boolean Variations yields a
Forman-Ricci curvature~\citep{forman2003bochner} on the parameter space.
Curvature is high when the loss and the Boolean Variations vary little across
the edges incident to a configuration, which is the combinatorial signature of
a flat basin.  Reducing the sharpness $B_k = \mathcal{L}_{\mathrm{adv}}(w) - \hat{\mathcal{L}}(w)$ raises a lower bound on its
average (\cref{app:ricci}).  It thus provides a computable descriptive
statistic of local flatness (\cref{sec:exp_landscape}).

\vspace{-1.7ex}

\paragraph{Algorithmic-complexity regularization.}
Because Hamming-SAM proceeds by bit-flips rather than gradients, a
non-differentiable complexity measure such as the Block Decomposition Method
(BDM)~\citep{zenil2018decomposition} can enter the objective directly, at
$\mathcal{O}(1)$ cost per candidate flip and with a convergence bound of the
same form.  \cite{bnn_compression} observed that BDM decreases during BNN
training but could not use it as a training-time regularizer, since it is
non-differentiable.  Hamming-SAM removes this obstacle, making
algorithmic-complexity regularization available as an optional extension of
the method (\cref{app:bdm}).

\vspace{-3.4ex}

\section{Experiments}
\label{sec:experiments}

\vspace{-1.2ex}

We evaluate Hamming-SAM on ResNet-18~\citep{he2016resnet},
VGG-16~\citep{simonyan2015vgg}, and ViT~\citep{dosovitskiy2021vit}
on CIFAR-10/100~\citep{krizhevsky2009cifar} and their corrupted
variants~\citep{hendrycks2019robustness}, as well as ResNet-50 on
ImageNet~\citep{deng2009imagenet}.  All binary methods replace
full-precision layers with Boolean (BOLD) or binary (STE) counterparts
without modifying the architecture or adding learnable parameters.
We compare: full-precision (FP), STE~\citep{hubara2016binarized},
STE with latent-space SAM~\citep{foret2021sam} (STE-SAM), standard
BOLD, and Hamming-SAM on the BOLD backend ($k{=}500$ by default, with a
$k$-ablation in \cref{sec:exp_landscape}).  Details
are in the Appendix, with additional theoretical
verifications in \cref{app:theory_verification}.

\vspace{-1.6ex}

\subsection{Generalization and Robustness}
\label{sec:exp_benchmark}

\vspace{-2.1ex}

\begin{figure}[H]
    \centering
    \tikzexternaldisable
    \centering
    \scriptsize
    \setlength{\figurewidth}{3.72cm}
    \setlength{\figureheight}{4.2cm}
    \input{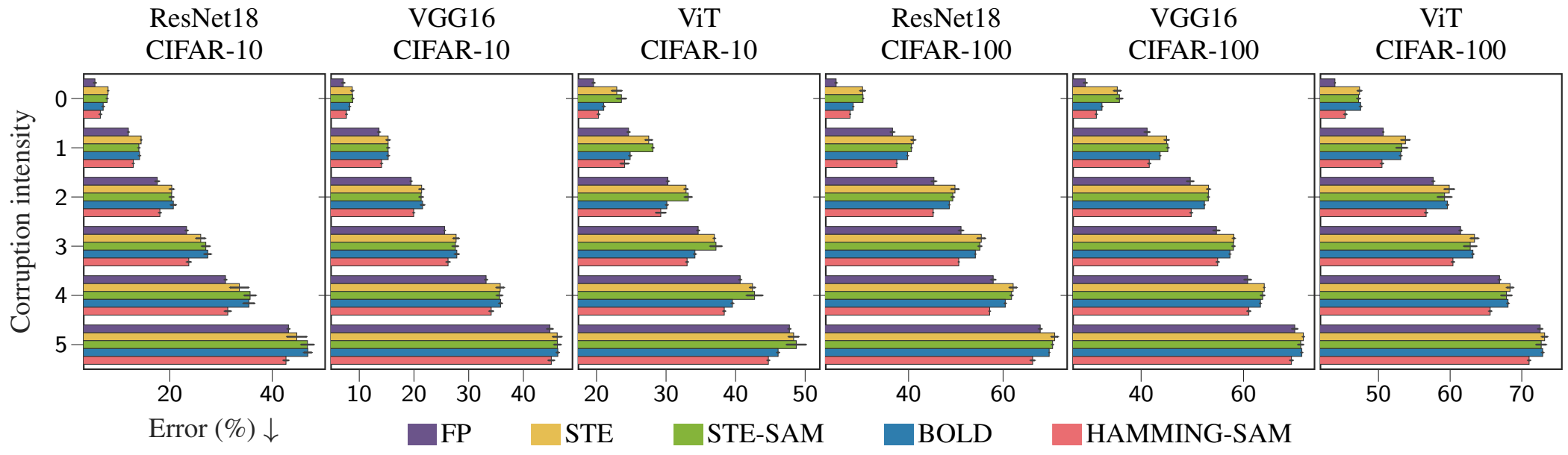}
    \tikzexternalenable

    \vspace{-3.3ex}
    \definecolor{darkslategray38}{RGB}{38,38,38}
\definecolor{darkslategray51}{RGB}{51,51,51}
\definecolor{darkslategray66}{RGB}{66,66,66}
\definecolor{salmon234109113}{RGB}{107,84,138}
\definecolor{lightgray204}{RGB}{204,204,204}
\definecolor{dimgray10784138}{RGB}{234,109,113}
\definecolor{sandybrown23019082}{RGB}{230,190,82}
\definecolor{steelblue46125174}{RGB}{46,125,174}
\definecolor{yellowgreen13318058}{RGB}{133,180,58}

\tikzexternaldisable

\scalebox{.99}{ 
    \setlength{\tabcolsep}{8pt} %
    \renewcommand{\arraystretch}{1.1}
    
    \begin{tabular}{cccccc}

        \qquad \qquad &
        
        {\protect\tikz[baseline=-.65ex] \protect\draw[fill=salmon234109113, draw=salmon234109113, line width=1.4pt] plot[mark size=2.0, only marks, mark=square*, mark options={xscale=1.5, yscale=1}] (0, 0);} FP & 
        
        {\protect\tikz[baseline=-.65ex] \protect\draw[fill=sandybrown23019082, draw=sandybrown23019082, line width=1.4pt] plot[mark size=2.0, only marks, mark=square*, mark options={xscale=1.5, yscale=1}] (0, 0);} STE & 
        
        {\protect\tikz[baseline=-.65ex] \protect\draw[fill=yellowgreen13318058, draw=yellowgreen13318058, line width=1.4pt] plot[mark size=2.0, only marks, mark=square*, mark options={xscale=1.5, yscale=1}] (0, 0);} STE-SAM & 
        
        {\protect\tikz[baseline=-.65ex] \protect\draw[fill=steelblue46125174, draw=steelblue46125174, line width=1.4pt] plot[mark size=2.0, only marks, mark=square*, mark options={xscale=1.5, yscale=1}] (0, 0);} BOLD & 
        
        {\protect\tikz[baseline=-.65ex] \protect\draw[fill=dimgray10784138, draw=dimgray10784138, line width=1.4pt] plot[mark size=2.0, only marks, mark=square*, mark options={xscale=1.5, yscale=1}] (0, 0);} HAMMING-SAM \\
        
    \end{tabular}
}

\tikzexternalenable

    \vspace{-1ex}

    \caption{\small{Error rate (\%, $\downarrow$) at corruption intensities for different architectures on CIFAR-10 and CIFAR-100.}}
    \label{fig:benchmark}

    \vspace{-3.9ex}

\end{figure}

\cref{fig:benchmark} reports error rates at corruption intensities 0 (clean) through 5 across all architecture--dataset combinations.
Three patterns appear consistently.

\vspace{-1.8ex}

\paragraph{Hamming-SAM improves over all Boolean baselines.}
On clean CIFAR-10, Hamming-SAM reduces the error of standard BOLD across architectures.
On CIFAR-100 the gains are comparable.  STE-SAM, which applies continuous SAM to latent weights, provides smaller and less consistent improvements
over STE, confirming that the vanishing perturbation effect (\cref{sec:pathology}) limits the effect of latent-space sharpness-aware training.

\vspace{-1.8ex}

\begin{figure}[H]
    \centering
    \tikzexternaldisable
    \centering
    \scriptsize
    \setlength{\figurewidth}{14.3cm}
    \setlength{\figureheight}{2.25cm}
    \input{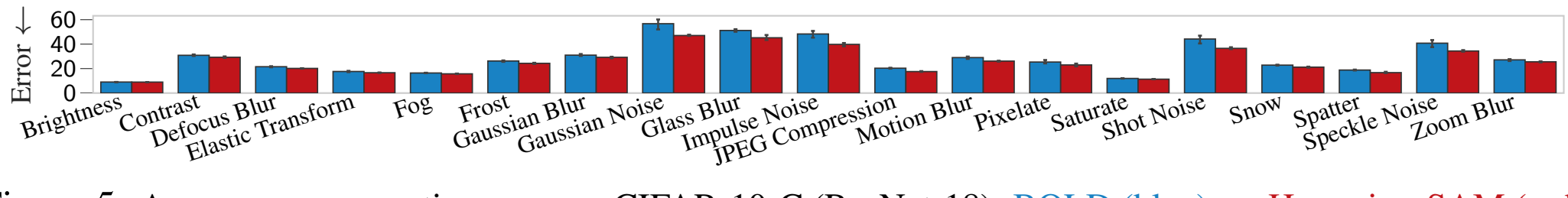}
    \tikzexternalenable

    \vspace{-2ex}

    \definecolor{brown1926161}{RGB}{193,21,27}
    \definecolor{darkcyan22112159}{HTML}{1982C4}

    \caption{\small{Average per-corruption error on CIFAR-10-C (ResNet-18).  {\textcolor{darkcyan22112159}{BOLD (blue)}} vs.\ \textcolor{brown1926161}{Hamming-SAM (red)}}.}
  \label{fig:per_corruption}

    \vspace{-4.3ex}

\end{figure}

\paragraph{Robustness gains exceed clean accuracy gains.}
As corruption intensity increases, the gap between Hamming-SAM and BOLD
widens.  The improvement in mean corruption accuracy exceeds the clean accuracy
improvement in every setting, by a factor of $3\times$ on average,
consistent with the theoretical prediction
that flat minima (low $B_k$) transfer better under distribution shift.
On several corruption types (e.g., ResNet-18 on CIFAR-10, see
\cref{fig:per_corruption}), Hamming-SAM narrows the gap to the
full-precision baseline substantially, despite using only $1$-bit
weights.  The gains also hold without BOLD.  On a standard STE backend,
Hamming-SAM improves both clean accuracy and robustness over STE-SAM
at the same cost (\cref{app:ste_backend}).

\vspace{-1.8ex}

\paragraph{Per-corruption analysis.}
\cref{fig:per_corruption} breaks down the CIFAR-10-C results for
ResNet-18.  Hamming-SAM reduces error on all 19 corruption types, with
the largest gains on high-frequency perturbations (Glass Blur, Shot
Noise, Impulse Noise) and smaller but consistent improvements on
low-frequency corruptions (Brightness, Fog).  This consistency
suggests that sharpness reduction is a general robustness
mechanism rather than a corruption-specific effect.

\vspace{-1.8ex}

\vspace{-1.6ex}

\subsection{Boolean Landscape Analysis and Theoretical Validation}
\label{sec:exp_landscape}

\vspace{-1.3ex}

\begin{figure}[H]

    \begin{subfigure}[c]{0.27\textwidth}
        \vspace{-0.5ex}
        \centering
        \tikzexternaldisable
        \centering
        \scriptsize
        \setlength{\figurewidth}{3.0cm}
        \setlength{\figureheight}{3.3cm}
        \input{figures/vgg_convergences.tikz}
        \tikzexternalenable
    \end{subfigure}
    \hfill
    \begin{subfigure}[c]{0.19\textwidth}
        \vspace{-1.5ex}
        \centering
        \tikzexternaldisable
        \centering
        \scriptsize
        \setlength{\figurewidth}{2.6cm}
        \setlength{\figureheight}{3.3cm}
        \begin{tikzpicture}

\definecolor{black17}{HTML}{1982C4}
\definecolor{crimson2385843}{RGB}{238,58,43}
\definecolor{darkslategray38}{RGB}{38,38,38}
\definecolor{darkslategray51}{RGB}{51,51,51}
\definecolor{firebrick1631420}{RGB}{163,14,20}
\definecolor{firebrick2022328}{RGB}{202,23,28}
\definecolor{lightgray204}{RGB}{204,204,204}
\definecolor{lightpink252186160}{RGB}{252,186,160}
\definecolor{salmon251145113}{RGB}{251,145,113}
\definecolor{tomato25010573}{RGB}{250,105,73}

\begin{groupplot}[group style={group size=2 by 1, horizontal sep=17.5pt}]
\nextgroupplot[
axis line style={darkslategray38},
height=\figureheight,
major tick length=1ex,
tick align=outside,
tick pos=left,
title style={align=center},
title={$|H|$ Std Dev \\ {\tiny($\times 10^{-7}$)}},
width=\figurewidth,
x grid style={lightgray204},
xmin=-0.74, xmax=6.74,
xtick=\empty,
xlabel={Method},
ymin=0, ymax=2.5305e-07,
scaled y ticks=false, %
ytick style={color=darkslategray38},
ytick={0,5e-08,1e-07,1.5e-07,2e-07,2.5e-07,3e-07},
yticklabels={0.0,0.5,1.0,1.5,2.0,2.5,3.0}
]
\draw[draw=darkslategray51,fill=black17,line width=0.48pt] (axis cs:-0.4,0) rectangle (axis cs:0.4,2.41e-07);
\draw[draw=darkslategray51,fill=lightpink252186160,line width=0.48pt] (axis cs:0.6,0) rectangle (axis cs:1.4,3.94e-09);
\draw[draw=darkslategray51,fill=salmon251145113,line width=0.48pt] (axis cs:1.6,0) rectangle (axis cs:2.4,6.17e-09);
\draw[draw=darkslategray51,fill=tomato25010573,line width=0.48pt] (axis cs:2.6,0) rectangle (axis cs:3.4,7.1e-09);
\draw[draw=darkslategray51,fill=crimson2385843,line width=0.48pt] (axis cs:3.6,0) rectangle (axis cs:4.4,9.71e-09);
\draw[draw=darkslategray51,fill=firebrick2022328,line width=0.48pt] (axis cs:4.6,0) rectangle (axis cs:5.4,3.79e-08);
\draw[draw=darkslategray51,fill=firebrick1631420,line width=0.48pt] (axis cs:5.6,0) rectangle (axis cs:6.4,4.47e-08);

\nextgroupplot[
axis line style={darkslategray38},
height=\figureheight,
major tick length=1ex,
tick align=outside,
tick pos=left,
title style={align=center},
title={$|H|$ 99th Pctile \\ {\tiny($\times 10^{-7}$)}},
width=\figurewidth,
x grid style={lightgray204},
xmin=-0.74, xmax=6.74,
xtick=\empty,
xlabel={Method},
ymin=0, ymax=1.26e-07,
scaled y ticks=false, %
ytick style={color=darkslategray38},
ytick={0,2e-08,4e-08,6e-08,8e-08,1e-07,1.2e-07,1.4e-07},
yticklabels={0.0,0.2,0.4,0.6,0.8,1.0,1.2,1.4}
]
\draw[draw=darkslategray51,fill=black17,line width=0.48pt] (axis cs:-0.4,0) rectangle (axis cs:0.4,9.52e-08);
\draw[draw=darkslategray51,fill=lightpink252186160,line width=0.48pt] (axis cs:0.6,0) rectangle (axis cs:1.4,1.86e-08);
\draw[draw=darkslategray51,fill=salmon251145113,line width=0.48pt] (axis cs:1.6,0) rectangle (axis cs:2.4,1.97e-08);
\draw[draw=darkslategray51,fill=tomato25010573,line width=0.48pt] (axis cs:2.6,0) rectangle (axis cs:3.4,2.15e-08);
\draw[draw=darkslategray51,fill=crimson2385843,line width=0.48pt] (axis cs:3.6,0) rectangle (axis cs:4.4,2.26e-08);
\draw[draw=darkslategray51,fill=firebrick2022328,line width=0.48pt] (axis cs:4.6,0) rectangle (axis cs:5.4,9.69e-08);
\draw[draw=darkslategray51,fill=firebrick1631420,line width=0.48pt] (axis cs:5.6,0) rectangle (axis cs:6.4,1.2e-07);
\end{groupplot}

\end{tikzpicture}
        \tikzexternalenable
    \end{subfigure}
    \hfill
    \begin{subfigure}[c]{0.445\textwidth}
        \centering
        \definecolor{color_baseline}{HTML}{1982C4}
\definecolor{color_red1}{HTML}{FC9272}
\definecolor{color_red2}{HTML}{FB6A4A}
\definecolor{color_red3}{HTML}{EF3B2C}
\definecolor{color_red4}{HTML}{CB181D}
\definecolor{color_red5}{HTML}{A50F15}
\definecolor{color_red6}{HTML}{67000D}

\newcommand{\sd}[1]{{\color{gray}\tiny$\pm$ #1}}

\tikzexternaldisable

\scalebox{.6}{
    \setlength{\tabcolsep}{1.2pt}
    \renewcommand{\arraystretch}{1.12}
    \begin{tabular}{lcccc}
        \toprule
         & Test Acc & mCA & $|H|_1 / |H|_5$ & Norm. \\
        Method & (\%) $\uparrow$ & (\%) $\uparrow$ & Ratio & $\mathrm{Ric}_F$ \\
        \midrule
        {\protect\tikz[baseline=-1ex]\protect\draw[color=color_baseline, fill=color_baseline, opacity=0.99, mark size=2.3pt, line width=2.3pt] plot[] (-0.0,0)--(-0.35,0);} BOLD & 91.65 \sd{0.12} & 70.56 \sd{0.18} & 53.53 & -2.12 \\
        {\protect\tikz[baseline=-1ex]\protect\draw[color=color_red1, fill=color_red1, opacity=0.99, mark size=2.3pt, line width=2.3pt] plot[] (-0.0,0)--(-0.35,0);} HAM.-SAM ($k=50$) & 91.84 \sd{0.10} & 71.59 \sd{0.15} & 2.20 & -2.00 \\
        {\protect\tikz[baseline=-1ex]\protect\draw[color=color_red2, fill=color_red2, opacity=0.99, mark size=2.3pt, line width=2.3pt] plot[] (-0.0,0)--(-0.35,0);} HAM.-SAM ($k=100$) & \textbf{92.31} \sd{0.11} & 72.18 \sd{0.14} & 2.08 & -1.98 \\
        {\protect\tikz[baseline=-1ex]\protect\draw[color=color_red3, fill=color_red3, opacity=0.99, mark size=2.3pt, line width=2.3pt] plot[] (-0.0,0)--(-0.35,0);} HAM.-SAM ($k=500$) & 92.15 \sd{0.13} & \textbf{72.39} \sd{0.16} & 4.85 & -1.96 \\
        {\protect\tikz[baseline=-1ex]\protect\draw[color=color_red4, fill=color_red4, opacity=0.99, mark size=2.3pt, line width=2.3pt] plot[] (-0.0,0)--(-0.35,0);} HAM.-SAM ($k=1000$) & 91.73 \sd{0.15} & 72.01 \sd{0.19} & 6.06 & -1.92 \\
        {\protect\tikz[baseline=-1ex]\protect\draw[color=color_red5, fill=color_red5, opacity=0.99, mark size=2.3pt, line width=2.3pt] plot[] (-0.0,0)--(-0.35,0);} HAM.-SAM ($k=5000$) & 91.05 \sd{0.21} & 70.97 \sd{0.22} & 5.38 & -1.80 \\
        {\protect\tikz[baseline=-1ex]\protect\draw[color=color_red6, fill=color_red6, opacity=0.99, mark size=2.3pt, line width=2.3pt] plot[] (-0.0,0)--(-0.35,0);} HAM.-SAM ($k=10000$) & 88.31 \sd{0.25} & 69.80 \sd{0.28} & 1.40 & -1.47 \\
        \bottomrule
    \end{tabular}
}

\tikzexternalenable

    \end{subfigure}

    \vspace{-3ex}

    \caption{\small{VGG on CIFAR-10. \emph{Left:} Training and test loss curves.
  \emph{Center:} Discrete Hessian statistics at convergence.
  \emph{Right:} Test accuracy, corruption robustness (mCA), Hessian spikiness
  ratio, and normalized Forman-Ricci curvature across methods and Hamming radii $k$.
  Results averaged over 4 seeds ($\pm$ std shown).}}
  \label{fig:vgg_cifar10}

  \vspace{-3.1ex}
\end{figure}

We use the VGG-16 / CIFAR-10 setup to validate the theoretical predictions, ablating over $k$ (\cref{fig:vgg_cifar10}).

\vspace{-1.8ex}

\paragraph{$k$-ablation.}
Test accuracy and corruption robustness both peak at moderate
$k \in [100, 500]$ and degrade for large $k$ (e.g., $k{=}10000$
falls below the BOLD baseline).  This matches the bias--sharpness
tradeoff of \cref{sec:choosing_k}.  Moderate $k$ probes meaningful
perturbations, while excessive $k$ forces the optimizer to defend
against unrealistically large Hamming balls.  Seed variance also grows
with $k$, as the interaction term $kh$ of \cref{thm:convergence_sam}
approaches the acceptance threshold and the tail of the selected set
becomes less stable across minibatches.

\vspace{-1.8ex}

\paragraph{Discrete Hessian spectrum.}
The standard deviation of sampled $|H_{ij}|$ drops by ${\sim}40\times$ from BOLD to Hamming-SAM $k{=}100$.
The spikiness ratio $|H|_1 / |H|_5$, the largest sampled $|H_{ij}|$ divided by the fifth largest,
drops from $53.5$ (BOLD) to between $1.4$ and $6.1$ across all Hamming-SAM radii.  This indicates a shift from a few dominant interactions to a nearly uniform spectrum.
Over $300$ random bit pairs, the largest interaction entry falls by $12\times$ during training to about $10^{-7}$ at convergence (\cref{app:hessian_verification}),
so the interaction term $kh$ in \cref{thm:convergence_sam} remains small.
Small interactions also make the loss change of a set of flips nearly additive, which keeps the submodularity ratio of \cref{thm:ascent} close to $1$, as confirmed empirically in \cref{app:gamma_k}.

\vspace{-1.8ex}

\paragraph{Forman-Ricci curvature.}
Normalized $\mathrm{Ric}_F$ (\cref{sec:ricci_estimation}) increases (becomes less negative) at Hamming-SAM solutions.
This is the expected direction.  Curvature is high when the loss and the Boolean Variations vary little across the edges incident to a configuration, and reducing the sharpness $B_k$ raises a lower bound on the average curvature (\cref{thm:avg_curvature}).
The curvature is highest at $k{=}10000$, which has the flattest local neighborhood but the worst test accuracy.
This reflects the trade-off in $k$: accuracy peaks at moderate radii while
curvature keeps increasing, so curvature is a local diagnostic of flatness
rather than an independent predictor of generalization.

\vspace{-1.8ex}

\paragraph{Training dynamics.}
The loss curves (\cref{fig:vgg_cifar10}, left) illustrate the generalization mechanism: Hamming-SAM maintains higher train loss
throughout but achieves monotonically decreasing test loss, while standard BOLD stalls at a worse minimum after ${\sim}20$K steps.

\vspace{-1.8ex}

\subsection{Large-scale Evaluation on ImageNet}

\vspace{-2.4ex}

\begin{table}[H]
    \centering
    \begin{minipage}[c]{0.68\textwidth}
        \cref{tab:resnet_imagenet} reports ResNet-50 results on ImageNet.
Hamming-SAM improves over standard BOLD on both test accuracy and OOD
AUROC (evaluated on ImageNet-C~\citep{hendrycks2019robustness}),
with gains of $0.69$ points in top-1 accuracy and $2.65$ points in OOD AUROC,
consistent with the pattern on CIFAR.  In contrast,
STE-SAM degrades both metrics relative to STE, confirming that
latent-space sharpness-aware training does not scale reliably and
that the geometric pathologies of \cref{sec:pathology} grow stronger at
ImageNet scale.
    \end{minipage}\hfill
    \renewcommand{\arraystretch}{1.4}
    \begin{minipage}[c]{0.3\textwidth}
        \centering
        \caption{\small ImageNet Results.}
        \label{tab:resnet_imagenet}
        \newcommand{\sd}[1]{{\color{gray}\scriptsize $\pm$ #1}}

        \definecolor{darkslategray38}{RGB}{38,38,38}
        \definecolor{darkslategray51}{RGB}{51,51,51}
        \definecolor{darkslategray66}{RGB}{66,66,66}
        \definecolor{salmon234109113}{RGB}{107,84,138}
        \definecolor{lightgray204}{RGB}{204,204,204}
        \definecolor{dimgray10784138}{RGB}{234,109,113}
        \definecolor{sandybrown23019082}{RGB}{230,190,82}
        \definecolor{steelblue46125174}{RGB}{46,125,174}
        \definecolor{yellowgreen13318058}{RGB}{133,180,58}

        \tikzexternaldisable

        \resizebox{\linewidth}{!}{%
            \begin{tabular}{lcc}
                \toprule
                \textbf{Method} & \textbf{Test Acc}  & \textbf{OOD AUC} \\ \midrule
                \midrule
                {\protect\tikz[baseline=-.65ex] \protect\draw[fill=sandybrown23019082, draw=sandybrown23019082, line width=1.4pt] plot[mark size=2.0, only marks, mark=square*, mark options={xscale=1.5, yscale=1}] (0, 0);} STE             & 68.97 \sd{0.09}   & 77.79 \sd{0.13} \\
                {\protect\tikz[baseline=-.65ex] \protect\draw[fill=yellowgreen13318058, draw=yellowgreen13318058, line width=1.4pt] plot[mark size=2.0, only marks, mark=square*, mark options={xscale=1.5, yscale=1}] (0, 0);} STE-SAM         & 68.53 \sd{0.15}   & 77.12 \sd{0.18} \\
                \midrule
                {\protect\tikz[baseline=-.65ex] \protect\draw[fill=steelblue46125174, draw=steelblue46125174, line width=1.4pt] plot[mark size=2.0, only marks, mark=square*, mark options={xscale=1.5, yscale=1}] (0, 0);} BOLD            & 69.03 \sd{0.11}   & 78.22 \sd{0.08} \\
                {\protect\tikz[baseline=-.65ex] \protect\draw[fill=dimgray10784138, draw=dimgray10784138, line width=1.4pt] plot[mark size=2.0, only marks, mark=square*, mark options={xscale=1.5, yscale=1}] (0, 0);} Hamming-SAM     & \textbf{69.72} \sd{0.14}   & \textbf{80.87} \sd{0.11} \\ \bottomrule
            \end{tabular}%
        }

        \tikzexternalenable
    \end{minipage}
\end{table}

\vspace{-1.8ex}

\vspace{-1.0ex}

\section{Conclusion}
\label{sec:conclusion}

\vspace{-1.1ex}

We introduced Hamming-SAM, the first sharpness-aware optimizer that
operates natively on the Boolean hypercube, eliminating the geometric
pathologies of latent-space approaches.  The theoretical framework is
unified by the discrete interaction Hessian, which governs the quality
of the adversarial ascent, the price of the one-shot approximation used
in practice, and the convergence rate of the descent.  Experiments across a
wide range of architectures and datasets demonstrate
consistent improvements in both clean accuracy and out-of-distribution
robustness.  We believe the Hamming-ball formulation and the
interaction-Hessian analysis developed here may find broader applications
wherever optimization over binary parameter spaces arises.

\newpage

\section*{Acknowledgements and Disclosures}

The author would like to thank the anonymous reviewers for their constructive comments,
as well as Jean-Claude Belfiore, Li Xu, Van Minh Nguyen, Li Rong, and Yun Yaw Chu for their assistance and support.
The author is an employee of Huawei Technologies France S.A.S.U., which supported this work.
The author declares no other competing interests.

{
\small

\bibliographystyle{abbrvnat}
\bibliography{IEEEabrv,bib/main_biblio.bib}

}

\newpage

{\bf \Large{Appendix}}

\setcounter{section}{0}
\renewcommand{\thesection}{\Alph{section}}

\begin{spacing}{0.2}
    \addtocontents{toc}{\protect\setcounter{tocdepth}{2}} %
    \renewcommand{\contentsname}{\textbf{Table of Contents}\vskip3pt\hrule}
    \tableofcontents
    \vskip5pt\hrule
\end{spacing}

\newpage

\section{Notations}
\label{app:notations}

The main notation used throughout the paper is summarized in
\cref{tab:notation}.

\begin{table}[H]
    \caption{Summary of main notations.}
    \label{tab:notation}
    \renewcommand{\arraystretch}{1.08}
    \centering\small
    \begin{tabular}{ll}
        \toprule
        \textbf{Symbol}                 & \textbf{Description}                                                                                                                \\
        \midrule
        \midrule
        \multicolumn{2}{l}{\textcolor{Brown}{\emph{Setup}}}                                                                                                               \\
        $n$                             & Number of Boolean parameters                                                                                                        \\
        $m$                             & Number of training samples ($|\mathcal{D}|$ or $|\mathcal{S}_1|$)                                                                   \\
        $\mathcal{D}$                   & Training dataset                                                                                                                     \\
        $w \in \{-1,+1\}^n$             & Boolean weight vector                                                                                                               \\
        $w^{(i)}$                       & $w$ with bit $i$ negated                                                                                                            \\
        $w^{(ij)}$                      & $w$ with bits $i$ and $j$ negated                                                                                                   \\
        $w^t \oplus \mathbf{1}_S$       & $w^t$ with all bits in set $S$ flipped                                                                                              \\
        \midrule
        \multicolumn{2}{l}{\textcolor{Brown}{\emph{Loss functions}}}                                                                                                      \\
        $\mathcal{L}(w)$                & Population (true) loss: $\mathbb{E}_{(x,y)\sim\mathcal{P}}[\ell(f_w(x), y)]$                                                        \\
        $\hat{\mathcal{L}}(w)$          & Empirical loss: $\frac{1}{m}\sum_{i=1}^{m}\ell(f_w(x_i), y_i)$                                                                      \\
        $\mathcal{L}_{\mathrm{adv}}(w)$ & Adversarial loss: $\max_{w' \in \mathcal{B}_k(w)} \hat{\mathcal{L}}(w')$                                                            \\
        $B$                             & Global loss bound: $\max_{w'} \hat{\mathcal{L}}(w')$                                                                                \\
        $B_k$                           & Sharpness: $\mathcal{L}_{\mathrm{adv}}(w) - \hat{\mathcal{L}}(w)$                                                                   \\
        \midrule
        \multicolumn{2}{l}{\textcolor{Brown}{\emph{Hamming geometry}}}                                                                                                    \\
        $d_H(w, w')$                    & Hamming distance between $w$ and $w'$                                                                                               \\
        $\mathcal{B}_k(w)$              & Hamming ball: $\{w' : d_H(w,w') \leq k\}$                                                                                           \\
        $V_k$                           & Hamming ball volume: $\sum_{j=0}^{k}\binom{n}{j}$                                                                                   \\
        $k$                             & Hamming radius (adversarial probe strength)                                                                                         \\
        \midrule
        \multicolumn{2}{l}{\textcolor{Brown}{\emph{Discrete calculus}}}                                                                                                   \\
        $f'_i(w)$                       & BOLD variation: $-w_i[\hat{\mathcal{L}}(w^{(i)}) - \hat{\mathcal{L}}(w)] = \hat{\mathcal{L}}(w|_{w_i=+1}) - \hat{\mathcal{L}}(w|_{w_i=-1})$                                               \\
        $\mathrm{BV}_i(w)$              & Boolean Variation (unsigned): $\hat{\mathcal{L}}(w^{(i)}) - \hat{\mathcal{L}}(w)$                                                   \\
        $H_{ij}(w)$                     & Interaction Hessian: $\hat{\mathcal{L}}(w^{(ij)}) - \hat{\mathcal{L}}(w^{(i)}) - \hat{\mathcal{L}}(w^{(j)}) + \hat{\mathcal{L}}(w)$ \\
        $h$                             & Hessian bound: $\max_{w,\,i \neq j} |H_{ij}(w)|$                                                                                    \\
        \midrule
        \multicolumn{2}{l}{\textcolor{Brown}{\emph{Ascent and convergence}}}                                                                                              \\
        $f(S)$                          & Set function: $\hat{\mathcal{L}}(w^t \oplus \mathbf{1}_S) - \hat{\mathcal{L}}(w^t)$                                                 \\
        $\gamma_k$                      & Submodularity ratio ($=1$ for submodular $f$)                                                                                       \\
        $\alpha$                        & Generalized curvature ($=0$ for modular $f$)                                                                                        \\
        $\eta$                          & Descent threshold (flip when $\mathrm{BV}_{i^*} < -\eta$)                                                                           \\
        \midrule
        \multicolumn{2}{l}{\textcolor{Brown}{\emph{PAC-Bayes}}}                                                                                                           \\
        $\mathcal{S}_0, \mathcal{S}_1$  & Data split for prior / posterior training                                                                                           \\
        $w_0$                           & Prior center (trained on $\mathcal{S}_0$)                                                                                           \\
        $d$                             & Adaptation distance: $d_H(\hat{w}, w_0)$                                                                                            \\
        $p$                             & Bernoulli flip probability in the prior                                                                                             \\
        $\beta_k$                       & Tail mass outside $\mathcal{B}_k(w)$ ($\leq e^{-k/6}$)                                                                              \\
        $\delta$                        & Confidence parameter (bound holds w.p.\ $\geq 1-\delta$)                                                                            \\
        \midrule
        \multicolumn{2}{l}{\textcolor{Brown}{\emph{Geometric and information-theoretic perspectives}}}                                                                                                        \\
        $\mathrm{Ric}_F$                & Forman-Ricci curvature on the Hamming graph                                                                                         \\
        $\beta$                         & Temperature in the loss-weighted Hamming graph                                                                                      \\
        $\mathrm{BDM}(w)$               & Block Decomposition Method (algorithmic complexity)                                                                                 \\
        $\lambda$                       & BDM regularization coefficient                                                                                                      \\
        \bottomrule
    \end{tabular}
\end{table}

\newpage
\section{Preliminaries: The BOLD Framework}
\label{sec:bold_background}

We briefly review the Boolean Logic Deep Learning (BOLD)
framework~\citep{nguyen2024bold}, which we use as the backend for
Hamming-SAM.
BOLD removes real-valued latent weights entirely.  Network parameters live
natively in $\{-1,+1\}^n$ (equivalently $\{\mathrm{T},\mathrm{F}\}^n$ in
Boolean logic).

\subsection{Boolean Neuron}

A BOLD neuron with input size $m$, weights $w_0, w_1, \ldots, w_m \in
    \{-1,+1\}$, and Boolean or real inputs computes the pre-activation:
\begin{equation}
    s = w_0 + \textstyle\sum_{i=1}^{m} \mathrm{L}(w_i, x_i),
    \label{eq:bold_neuron}
\end{equation}
where $\mathrm{L}$ is a Boolean logic gate (typically XNOR) and the summation counts the number of TRUE outputs.
The framework is flexible, as it allows Boolean linear layers to be connected through activation layers, normalization layers, or other types of layers.

\subsection{Boolean Variation and Chain Rule}
\label{app:bold_bv}

The BOLD variation~\citep{nguyen2024bold} of $\hat{\mathcal{L}}$ with respect to $w_i$ is
\begin{equation}
  f'_i(w) = \mathbf{xnor}\bigl(\delta(w_i \to \neg w_i),\,
  \delta\hat{\mathcal{L}}(w \to w^{(i)})\bigr),
\end{equation}
where $\delta(\cdot \to \cdot)$ denotes the direction of change and $w^{(i)}$ is
$w$ with bit $i$ negated.  For a real-valued loss it evaluates to
\begin{equation}
  f'_i(w) = -w_i\bigl[\hat{\mathcal{L}}(w^{(i)}) - \hat{\mathcal{L}}(w)\bigr]
  = \hat{\mathcal{L}}(w|_{w_i=+1}) - \hat{\mathcal{L}}(w|_{w_i=-1})
\end{equation}
(\cref{sec:bv_bridge}), the partial derivative of pseudo-Boolean function
theory~\citep{crama2011boolean, odonnell2014analysis}.  It compares the two
values of bit $i$ and does not depend on the current value of $w_i$.  It plays
the role of the gradient.  We have $f'_i > 0$ when $\hat{\mathcal{L}}$ increases with
$w_i$, and flipping bit $i$ decreases the loss when $f'_i$ and $w_i$ have the
same sign.

BOLD also provides a chain rule for compositions
(Theorem~3.12 of~\cite{nguyen2024bold}), under the assumptions stated there,
which propagates variations through a deep Boolean network and returns a signal
for all $n$ weights in a single backward pass.  Evaluating $\mathrm{BV}$ exactly
would instead require $n+1$ forward passes.  We treat the backward signal as an
estimate.  Flipping bit $i$ moves $w_i$ by $-2w_i$, so a first-order expansion
gives $\mathrm{BV}_i(w) \approx -2w_i\,\partial\hat{\mathcal{L}}/\partial w_i$.  For the
multilinear extension of $\hat{\mathcal{L}}$ this relation is exact, since its
partial derivative with respect to $w_i$ equals $f'_i$ divided by $2$.  The
backward pass differentiates the network's own real-valued computation instead,
so the estimate is exact only when the two agree.  The positive factor $2$ does
not affect the algorithm, because Phase~1 ranks bits and Phase~2 tests signs.
\cref{app:ascent} quantifies how an oracle of this kind affects the ascent
guarantee.

\subsection{Accumulator-Based Optimizer}
\label{sec:bold_optim}

The per-sample atomic variation $q_{i,k} =
    \mathrm{xnor}(\delta\mathcal{L}/\delta x^{l+1}_k, x^l_k)$ (Eq.~5
of~\citep{nguyen2024bold}) is aggregated over the batch into an integer signal
$q_i$ (Eq.~7 of~\citep{nguyen2024bold}). The simplest update rule flips $w_i$
when $\mathrm{xnor}(q_i, w_i) = \mathrm{T}$, that is, when the aggregated signal
indicates that flipping decreases the loss.

In practice, BOLD uses an \emph{accumulator} $a_i^t$ that integrates signals
across iterations:
\begin{equation}
    a_i^{t+1} = r^t a_i^t + \eta^t q_i^{t+1},
    \label{eq:bold_accum}
\end{equation}
where $\eta^t$ is a learning rate and $r^t = N_{\mathrm{unchanged}}^t / N_{\mathrm{total}}$ is a \emph{plasticity ratio} that down-weights the accumulator when many bits are flipping (high plasticity) and preserves it when the network is stable (low plasticity).
A bit is flipped when $\mathbf{xnor}(a_i, w_i) = \mathrm{T}$, that is, when
$a_i w_i > 0$.  The reference implementation uses the thresholded form
$a_i w_i \geq 1$ (Algorithm~8 of~\cite{nguyen2024bold}), which we also use, and
the accumulator resets to zero on flip.  This provides an implicit, adaptive
convergence threshold (analyzed in \cref{app:accumulator_threshold}).

\subsection{Notation Bridge: Unsigned Loss Change}
\label{sec:bv_bridge}

For the theoretical analysis we work with the loss change of a single flip,
\begin{equation}
  \mathrm{BV}_i(w) = \hat{\mathcal{L}}(w^{(i)}) - \hat{\mathcal{L}}(w),
\end{equation}
because it is the change caused by the move the algorithm makes.  It removes
sign factors from the set function $f(S) = \hat{\mathcal{L}}(w^t \oplus \mathbf{1}_S) - \hat{\mathcal{L}}(w^t)$,
the interaction Hessian and the convergence analysis.

In the variation of the loss $\hat{\mathcal{L}}$ defined by BOLD,
$f'_i(w) = \mathbf{xnor}\bigl(\delta(w_i \to \neg w_i),\, \delta\hat{\mathcal{L}}(w \to w^{(i)})\bigr)$,
the second argument is numeric and equals the finite difference
$\hat{\mathcal{L}}(w^{(i)}) - \hat{\mathcal{L}}(w)$.  With the embedding
$e(\mathrm{T}) = +1$, $e(\mathrm{F}) = -1$ and
$\mathbf{xnor}(a, x) = e(a)\,x$ for a logic value $a$ and a number $x$
(Proposition~A.3 of~\cite{nguyen2024bold}),
\begin{equation}
  f'_i(w) = e\bigl(\delta(w_i \to \neg w_i)\bigr)\cdot
  \bigl[\hat{\mathcal{L}}(w^{(i)}) - \hat{\mathcal{L}}(w)\bigr].
\end{equation}
Now $e(\delta(w_i \to \neg w_i)) = -w_i$.  If $w_i = +1$, the direction is down,
$\delta = \mathrm{F}$ and $e = -1$.  If $w_i = -1$, the direction is up,
$\delta = \mathrm{T}$ and $e = +1$.  Therefore
\begin{equation}
  f'_i(w) = -w_i\,\mathrm{BV}_i(w)
  = \hat{\mathcal{L}}(w|_{w_i=+1}) - \hat{\mathcal{L}}(w|_{w_i=-1}),
  \qquad
  \mathrm{BV}_i(w) = -w_i\,f'_i(w).
\end{equation}
The variation compares the two values of bit $i$, while $\mathrm{BV}_i$ flips its
current value.  If $w_i = +1$, the flip goes from $+1$ to $-1$ and
$\mathrm{BV}_i = -f'_i$, and if $w_i = -1$ it goes from $-1$ to $+1$ and
$\mathrm{BV}_i = f'_i$.
The product $f'_i \cdot w_i = -\mathrm{BV}_i$ is positive exactly when flipping
bit $i$ decreases the loss, while positive $\mathrm{BV}_i$ means that flipping it
increases the loss.  Since $w_i^2 = 1$, either quantity recovers the other.
At second order the same sign factors give
$H_{ij}(w) = w_iw_j\bigl[f'_i(w|_{w_j=+1}) - f'_i(w|_{w_j=-1})\bigr]$.  In \cref{alg:hamming_sam}, Phase~2
reduces to BOLD's flip rule in the $f'_i$ convention.  A bit is flipped when
$\mathbf{xnor}(f'_i, w_i) = \mathrm{T}$, equivalently when $\mathrm{BV}_i < 0$.

\section{Proofs for Pathological Behavior of Latent-Weight SAM}
\label{app:pathology_proofs}

The main motivation for a native Boolean SAM comes from
the geometric pathologies of training binarized networks
via latent-weight approximations. In the standard approach,
high-precision latent weights $\tilde{w} \in \mathbb{R}^n$ are
maintained and projected onto the Boolean space through a
non-differentiable sign function,
$w = \mathrm{sign}(\tilde{w}) \in \{-1,+1\}^n$~\citep{hubara2016binarized}.
Gradients are approximated using the STE.  When continuous SAM is applied
to these latent weights, it searches for an adversarial perturbation
$\epsilon$ within an $\ell_p$-norm ball of radius $\rho$ to maximize
the loss:
\begin{equation}
  \max_{\|\epsilon\|_p \leq \rho}\; \mathcal{L}(\tilde{w} + \epsilon).
  \label{eq:continuous_sam}
\end{equation}
This approach has two failures.

\subsection{The Vanishing Perturbation Effect}

The sign function $\mathrm{sign}(\cdot)$ acts as a quantization
barrier.  For a latent weight $\tilde{w}_i$ that has been optimized to
high magnitude, and therefore represents high confidence, a perturbation
$\epsilon$ bounded by $\rho$ is frequently insufficient to cross the
zero threshold.  Consequently, the binarized network's functional output
remains identical:
$\mathrm{sign}(\tilde{w}_i + \epsilon_i) = \mathrm{sign}(\tilde{w}_i)$
for all $|\epsilon_i| \leq \rho < |\tilde{w}_i|$.  The adversarial
loss then equals the clean loss.  The sharpness-aware signal
vanishes, and the optimizer reverts to standard SGD.  This
is a binary-specific case of the perturbation
shrinkage reported by \cite{liu2021saq} for multi-bit quantized
networks.

\begin{mybox}
  \begin{repeatenv}{proposition}{\ref{prop:vanishing}}[Vanishing Perturbation]
    Let $\tilde{w} \in \mathbb{R}^n$ be latent weights and
    $w = \mathrm{sign}(\tilde{w})$.  Define
    $\mathcal{C}(\rho) = \{i \in [n] : |\tilde{w}_i| > \rho\}$ as the set
    of ``confident'' coordinates.  Then for any $\epsilon$ with
    $\|\epsilon\|_\infty \leq \rho$, the binarized adversarial point
    satisfies
    $\mathrm{sign}(\tilde{w}_i + \epsilon_i) = \mathrm{sign}(\tilde{w}_i)$
    for all $i \in \mathcal{C}(\rho)$.  In particular, if
    $|\mathcal{C}(\rho)| = n$ (all weights confident), then
    $\max_{\|\epsilon\|_\infty \leq \rho}
    \mathcal{L}(\mathrm{sign}(\tilde{w} + \epsilon)) =
    \mathcal{L}(\mathrm{sign}(\tilde{w}))$, and SAM reduces to standard
    gradient descent on the latent loss.
  \end{repeatenv}
\end{mybox}
\begin{proof}
  If $|\tilde{w}_i| > \rho \geq |\epsilon_i|$, then
  $\tilde{w}_i + \epsilon_i$ has the same sign as $\tilde{w}_i$.  The
  second claim follows by noting that the binarized output is identical
for all perturbations in the $\ell_\infty$-ball.
\end{proof}

\noindent The statement is given for the $\ell_\infty$-ball, which is the
weakest form.  Since $\|\epsilon\|_2 \leq \rho$ implies
$\|\epsilon\|_\infty \leq \rho$, the same conclusion holds for the
$\ell_2$-ball used by continuous SAM in practice.

\subsection{Euclidean--Hamming Geometric Mismatch}

The second failure comes from a geometric mismatch.  Flatness in the
continuous latent space does not imply robustness in the actual discrete
operating domain.  Two latent configurations that are close in
Euclidean distance may map to Boolean configurations that are far apart
in Hamming distance, and vice versa.

\begin{mybox}
  \begin{repeatenv}{proposition}{\ref{prop:distortion}}[Metric Distortion]
  The map $\pi : \mathbb{R}^n \to \{-1,+1\}^n$ defined by
  $\pi(\tilde{w}) = \mathrm{sign}(\tilde{w})$ can exhibit arbitrarily
  large metric distortion: for any $D > 0$, there exist
  $\tilde{u}, \tilde{v} \in \mathbb{R}^n$ with
  $\|\tilde{u} - \tilde{v}\|_2 < D^{-1}$ but
  $d_H(\pi(\tilde{u}), \pi(\tilde{v})) = n$, and conversely,
  $\tilde{u}', \tilde{v}'$ with
  $\|\tilde{u}' - \tilde{v}'\|_2 > D$ but
  $d_H(\pi(\tilde{u}'), \pi(\tilde{v}')) = 0$.
  \end{repeatenv}
\end{mybox}
\begin{proof}
For the first claim, take $\tilde{u}_i = 1/(4nD)$ and
$\tilde{v}_i = -1/(4nD)$ for all $i$.  Then
$\|\tilde{u} - \tilde{v}\|_2 = 1/(2D\sqrt{n}) < D^{-1}$ for every $n \geq 1$,
while every
coordinate of $\tilde{u}$ is positive and every coordinate of
$\tilde{v}$ is negative, so
$d_H(\pi(\tilde{u}), \pi(\tilde{v})) = n$.  For the second, take
$\tilde{u}'_i = 1$ and $\tilde{v}'_i = 1 + 2D/\sqrt{n}$.  Then
$\|\tilde{u}' - \tilde{v}'\|_2 = 2D > D$, and both vectors lie in the
positive orthant, so $\pi(\tilde{u}') = \pi(\tilde{v}')$.
\end{proof}

\noindent Both constructions place the two points on opposite sides of the
sign boundary, or strictly inside the same orthant, which is what
makes the distortion unbounded in either direction.  Euclidean proximity is
therefore neither necessary nor sufficient for Hamming proximity.

\section{Justifications for the Objective}
\label{app:objective_justifications}

\paragraph{Structural parallel with \cite{foret2021sam}.}
The foundational theorem of continuous SAM~\citep{foret2021sam}
establishes that, for any $\rho > 0$, with high probability over
$\mathcal{S} \sim \mathcal{D}^m$:
\begin{equation}
    \mathcal{L}_\mathcal{D}(w) \;\leq\;
    \underbrace{\textcolor{BrickRed}{\max_{\|\epsilon\|_2 \leq \rho}
        \hat{\mathcal{L}}_S(w + \epsilon)}}_{\text{sharpness-aware loss}}
    \;+\; \textcolor{NavyBlue}{h\!\bigl(\|w\|_2^2 / \rho^2\bigr)},
    \label{eq:foret_bound}
\end{equation}
where $h$ is a strictly increasing function incorporating the KL
divergence between a Gaussian posterior $Q = \mathcal{N}(w,\rho^2 I)$
and a grid of Gaussian priors.  The continuous SAM objective
$\min_w \max_{\|\epsilon\|_2\leq\rho} \hat{\mathcal{L}}_S(w+\epsilon)$
is then motivated directly by minimizing the right-hand side
of~\eqref{eq:foret_bound}.

Hamming-SAM's Eq.~\eqref{eq:hamming_sam_objective} follows the same derivation
on the Boolean hypercube, with the replacements below.
\begin{center}
    \small
    \begin{tabular}{lll}
        \toprule
        \textbf{Component}   & \textbf{Foret et al.\ (continuous)}                    &
        \textbf{Hamming-SAM (Boolean)}                                                                \\
        \midrule
        Parameter space      & $\mathbb{R}^n$                                         & $\{-1,+1\}^n$ \\
        Perturbation set     & $\ell_2$-ball $\{\epsilon: \|\epsilon\|_2 \leq \rho\}$
                             & Hamming ball $\mathcal{B}_k(w)$                                        \\
        Stochastic posterior & $Q = \mathcal{N}(w, \rho^2 I)$
                             & $Q = \mathrm{BernoulliFlip}(w, q)$                                     \\
        Complexity term      & $h(\|w\|_2^2/\rho^2)$ via Gaussian KL
                             & $d_H(\hat{w}, w_0)\!\cdot\!\log(1/p)$ via Bernoulli KL                 \\
        \bottomrule
    \end{tabular}
\end{center}

We now make this analogy precise.  We give two theoretical
justifications, a PAC-Bayes upper bound (the direct Boolean counterpart of
\cite{foret2021sam}) and a compression-based argument, followed by an
equivalent perspective of the objective as distributionally robust optimization.

\subsection{Derivation I: PAC-Bayes Upper Bound}
\label{app:deri_pac_bayes}

In continuous SAM, \cite{foret2021sam} use a Gaussian posterior $Q =
\mathcal{N}(w, \rho^2 I)$ so that $\mathbb{E}_{\epsilon \sim
Q}[\hat{\mathcal{L}}(w + \epsilon)]$ is controlled by
$\max_{\|\epsilon\|_2 \leq \rho} \hat{\mathcal{L}}(w + \epsilon)$
via tail concentration of the $\chi^2$ distribution.  We repeat
this argument on the Boolean hypercube using a Bernoulli bit-flip
distribution as the stochastic posterior.

Consider the bit-flip distribution $P$ centered at $w$ that
independently flips each bit with probability
$q \in (0, \tfrac{1}{2})$.  For any
$w' \in \{-1,+1\}^n$ with $d_H(w',w) = j$, the distribution
assigns mass $P(w') = q^j(1-q)^{n-j}$.  The expected empirical
loss under $P$ is:
\[
    \mathbb{E}_{w' \sim P}\bigl[\hat{\mathcal{L}}(w')\bigr]
    = \sum_{j=0}^{n} \sum_{\substack{w' : d_H(w',w)=j}}
    q^j(1-q)^{n-j}\,\hat{\mathcal{L}}(w').
\]

\begin{mybox}
    \begin{proposition}[Adversarial Loss as a Stochastic Upper Bound]
        \label{prop:pac_bayes_motivation}
        For any $w \in \{-1,+1\}^n$ and any $q \in (0,\tfrac{1}{2})$:
        \begin{equation}
            \mathbb{E}_{w' \sim P}\bigl[\hat{\mathcal{L}}(w')\bigr]
            \;\leq\;
            \underbrace{\Bigl[1 - \sum_{j=0}^{k}\tbinom{n}{j}q^j(1{-}q)^{n-j}
                    \Bigr]}_{\displaystyle \triangleq\;\beta_k}
            \cdot B
            \;+\; \bigl(1 - \beta_k\bigr)\cdot
            \mathcal{L}_{\mathrm{adv}}(w),
        \end{equation}
        where $B = \max_{w'} \hat{\mathcal{L}}(w')$ is the global loss bound
        and $\beta_k$ is the probability mass outside $\mathcal{B}_k(w)$.
        In particular, for $q \leq k/(2n)$, we have
        $\beta_k \leq e^{-k/6}$ (by a Chernoff bound), so the expected loss
        is dominated by $\mathcal{L}_{\mathrm{adv}}(w)$.
    \end{proposition}
\end{mybox}

\begin{proof}
    Split the expectation over the Hamming ball and its complement:
    \begin{align}
        \mathbb{E}_{w'\sim P}[\hat{\mathcal{L}}(w')]
         & = \underbrace{\sum_{\substack{w': \\d_H(w',w)\leq k}}
        P(w')\,\hat{\mathcal{L}}(w')}_{
        \leq\;(1-\beta_k)\cdot\mathcal{L}_{\mathrm{adv}}(w)}
        +\;\underbrace{\sum_{\substack{w':   \\d_H(w',w)>k}}
        P(w')\,\hat{\mathcal{L}}(w')}_{
        \leq\;\beta_k \cdot B}.
        \label{eq:split_expectation}
    \end{align}
    The first inequality holds because
    $\hat{\mathcal{L}}(w') \leq \mathcal{L}_{\mathrm{adv}}(w)$ for all
    $w' \in \mathcal{B}_k(w)$ and the total mass on $\mathcal{B}_k(w)$ is
    $1 - \beta_k$.  The second holds since
    $\hat{\mathcal{L}}(w') \leq B$ everywhere.

    For the tail bound, the number of bit-flips under $P$ is
$\mathrm{Bin}(n, q)$ with mean $\mu = nq = k/2$ when
$q = k/(2n)$.  The standard multiplicative Chernoff
bound~\citep{Boucheron2013concentration} at $\delta = 1$
(i.e., $\Pr[X > 2\mu]$) gives
\[
  \beta_k
  = \Pr[\mathrm{Bin}(n,q) > k]
  \leq \left(\frac{e}{4}\right)^{\!\mu}
  = \left(\frac{e}{4}\right)^{\!k/2}
  = e^{-\frac{\ln 4 - 1}{2}\,k}
  \leq e^{-k/6},
\]
since $(\ln 4 - 1)/2 \approx 0.193 \geq 1/6$.  Smaller $q$ only decreases
the tail mass, so the bound holds for all $q \leq k/(2n)$.
\end{proof}

Combining this with the standard PAC-Bayes theorem gives the
Boolean analogue of the generalization bound of \cite{foret2021sam}.

\begin{mybox}
    \begin{repeatenv}{theorem}{\ref{thm:discrete_foret}}[Discrete
        PAC-Bayes Sharpness Bound, Boolean Analogue of
        {\cite{foret2021sam}}]
        Let $P_0$ be a prior over $\{-1,+1\}^n$ independent of the
        training set $\mathcal{S}$ of size $m$, and let $P$ be the
        Bernoulli bit-flip distribution centered at $w$ with flip
        probability $q = k/(2n)$.  For any $\delta \in (0,1)$, with
        probability at least $1 - \delta$ over
        $\mathcal{S} \sim \mathcal{D}^m$:
        \begin{equation}
            \mathbb{E}_{w'\sim P}\bigl[\mathcal{L}(w')\bigr]
            \;\leq\;
            \underbrace{\textcolor{BrickRed}{(1 - \beta_k)\cdot\mathcal{L}_{\mathrm{adv}}(w)
            + \beta_k B}}_{\textcolor{BrickRed}{\text{sharpness-aware empirical term}}}
            \;+\;
            \underbrace{\textcolor{NavyBlue}{\sqrt{\frac{\mathrm{KL}(P \| P_0)
                        + \log(2\sqrt{m}/\delta)}{2m}}}}_{\textcolor{NavyBlue}{\text{complexity term}}},
            \label{eq:discrete_foret}
        \end{equation}
        where $\beta_k \leq e^{-k/6}$ and the complexity term depends on the
        choice of prior $P_0$.
    \end{repeatenv}
\end{mybox}

\begin{proof}
    By the PAC-Bayes--McAllester inequality~\citep{mcallester2003pac},
    for any posterior $Q$ and data-independent prior $P_0$:
    \[
      \mathbb{E}_{w'\sim Q}[\mathcal{L}(w')]
      \;\leq\;
      \mathbb{E}_{w'\sim Q}[\hat{\mathcal{L}}(w')]
      \;+\;
      \sqrt{\frac{\mathrm{KL}(Q\|P_0) + \log(2\sqrt{m}/\delta)}{2m}}.
    \]
    Set $Q = P$ (the Bernoulli bit-flip distribution centered at $w$).
    By \cref{prop:pac_bayes_motivation}, the expected
    empirical loss satisfies
    $\mathbb{E}_{w'\sim P}[\hat{\mathcal{L}}(w')]
    \leq (1-\beta_k)\,\mathcal{L}_{\mathrm{adv}}(w) + \beta_k B$.
    Substituting yields~\eqref{eq:discrete_foret}.
\end{proof}

\begin{remark}[One-to-one correspondence with {\cite{foret2021sam}}]
    \label{rem:foret_correspondence}
    The structure of \cref{thm:discrete_foret} mirrors
    \cite{foret2021sam} exactly.
    \emph{(i)}~The \textcolor{BrickRed}{sharpness-aware term}
    $(1-\beta_k)\mathcal{L}_{\mathrm{adv}}(w) + \beta_k B$ plays the
    role of $\max_{\|\epsilon\|_2\leq\rho}\hat{\mathcal{L}}(w+\epsilon)$,
    with $\beta_k \leq e^{-k/6}$ replacing the Gaussian tail.
    \emph{(ii)}~The \textcolor{NavyBlue}{complexity term} is driven by a
    Bernoulli KL in place of a Gaussian KL.
    \emph{(iii)}~Both bounds control the \emph{expected perturbed loss}
    $\mathbb{E}_{w'\sim Q}[\mathcal{L}(w')]$ rather than the
    deterministic loss at the center, so minimizing
    $\mathcal{L}_{\mathrm{adv}}(w)$ is motivated by its dominance on the
    right-hand side in both frameworks.  The key difference is
    that our bound operates directly on $\{-1,+1\}^n$.
\end{remark}

\subsection{Derivation II: Compression-Based Generalization}
\label{app:deri_compression}

A second justification comes from a volume argument on the
Boolean hypercube.  The main idea is that if $w$ has low adversarial loss,
then \emph{every} configuration in $\mathcal{B}_k(w)$ also has low loss.  A
flat solution therefore identifies not a single hypothesis but an entire ball
of them, and the log-volume of that ball is exactly what the complexity term
pays for.

\begin{mybox}
    \begin{theorem}[Hamming-Ball Compression Bound]
        \label{thm:compression}
        Suppose the Hamming-SAM solution $\hat{w}$ satisfies
        $\mathcal{L}_{\mathrm{adv}}(\hat{w}) \leq \epsilon$ on a training set
        of size $m$, and let $Q_{\hat{w}}$ denote the uniform distribution on
        $\mathcal{B}_k(\hat{w})$.  Then for any $\delta \in (0,1)$, with
        probability at least $1 - \delta$:
        \begin{equation}
            \mathbb{E}_{w' \sim Q_{\hat{w}}}\bigl[\mathcal{L}(w')\bigr]
            \;\leq\; \textcolor{BrickRed}{\epsilon}
            + \textcolor{NavyBlue}{\sqrt{\frac{n\log 2 - \log V_k
                    + \log(2\sqrt{m}/\delta)}{2m}}},
            \qquad V_k = \sum_{j=0}^{k}\binom{n}{j}.
            \label{eq:compression_bound}
        \end{equation}
    \end{theorem}
\end{mybox}

\begin{proof}
Take the prior $P_0$ to be uniform on $\{-1,+1\}^n$, which is independent of
the training set, and the posterior $Q_{\hat{w}}$ to be uniform on
$\mathcal{B}_k(\hat{w})$.  Both are supported on the hypercube, and
\[
  \mathrm{KL}(Q_{\hat{w}} \| P_0)
  \;=\; \sum_{w' \in \mathcal{B}_k(\hat{w})} \frac{1}{V_k}
  \log\frac{1/V_k}{2^{-n}}
  \;=\; \log\frac{2^n}{V_k}
  \;=\; n \log 2 - \log V_k .
\]
The adversarial guarantee gives $\hat{\mathcal{L}}(w') \leq \epsilon$ for
\emph{every} $w' \in \mathcal{B}_k(\hat{w})$, hence
$\mathbb{E}_{w'\sim Q_{\hat{w}}}[\hat{\mathcal{L}}(w')] \leq \epsilon$.
Substituting both into the PAC-Bayes--McAllester
inequality~\citep{mcallester2003pac}
yields~\eqref{eq:compression_bound}.
\end{proof}

\noindent Three remarks follow.

\emph{The object bounded.}  The statement controls the Gibbs predictor drawn
uniformly from the flat basin.  A posterior spread over the ball is what makes
the $V_k$ saving available, since a point-mass posterior against a uniform prior
has $\mathrm{KL} = n\log 2$.  Drawing $w' \sim Q_{\hat{w}}$ is also the deployed
model after up to $k$ bit-flips, so~\eqref{eq:compression_bound} bounds the
loss under bit-level corruption of the weights, a realistic failure mode for
binary networks on edge hardware.  If the population loss varies by at most
$\xi$ over $\mathcal{B}_k(\hat{w})$, the same bound holds for
$\mathcal{L}(\hat{w})$ up to $\xi$, and \cref{thm:pac_bayes_full} bounds
$\mathcal{L}(\hat{w})$ directly, with a point-mass posterior and a
data-dependent prior.

\emph{No covering is needed.}  The reduction comes from the
posterior being spread over $V_k$ configurations that the adversarial
guarantee certifies simultaneously, so the flatness of $\hat{w}$ is what pays
for the complexity saving.

\emph{Roles of the three justifications.}
\cref{thm:discrete_foret} and~\eqref{eq:compression_bound} both control a
perturbed loss and motivate the objective, differing in the perturbation
distribution (Bernoulli flips versus uniform on the ball).
\cref{thm:pac_bayes_full} supplies the deterministic guarantee.  The three are
complementary in what they bound, not merely in how they are proved.

The term $\log V_k$ is the complexity saving gained from flatness, since the
bound counts a ball of configurations rather than a single point.
Because $\log V_k \approx k\log(n/k)$ grows more slowly than $n\log 2$, the
saving is small in relative terms at the scales we study (for ResNet-18 with
$k=500$, roughly $5\times 10^{3}$ nats against $7.7\times10^{6}$), and with a
uniform prior the bound is vacuous when $n \gg m$.  The data-dependent prior of
\cref{app:pac_bayes} counts complexity per adapted bit instead.

\begin{mybox}
    \begin{corollary}[Hamming-SAM Reduces Effective Hypothesis Complexity]
        \label{cor:complexity_reduction}
        Minimizing $\mathcal{L}_{\mathrm{adv}}(w)$ with Hamming radius $k$
        replaces the complexity $n \log 2$ of an arbitrary Boolean
        configuration by $n\log 2 - \log V_k$.  For $k \ll n$, Stirling's
        approximation gives
        $\log V_k \approx k\log(n/k) + k - \tfrac{1}{2}\log(2\pi k)$ nats.
    \end{corollary}
\end{mybox}

\subsection{An Equivalent Perspective: Distributionally Robust Optimization}
\label{app:deri_dro}

The min-max objective~\eqref{eq:hamming_sam_objective} can be written as a
Distributionally Robust Optimization (DRO) problem, which connects Hamming-SAM
to the robust-optimization literature.

\begin{mybox}
    \begin{proposition}[Hamming-SAM as DRO]
        \label{prop:dro}
        The Hamming-SAM objective is equivalent to a DRO problem over discrete
        ambiguity sets:
        \begin{equation}
            \mathcal{L}_{\mathrm{adv}}(w)
            = \max_{w' \in \mathcal{B}_k(w)} \hat{\mathcal{L}}(w')
            = \max_{\mu \in \Delta(\mathcal{B}_k(w))}
            \mathbb{E}_{w' \sim \mu}\bigl[\hat{\mathcal{L}}(w')\bigr],
            \label{eq:dro}
        \end{equation}
        where $\Delta(\mathcal{B}_k(w))$ denotes the set of all probability
        distributions supported on $\mathcal{B}_k(w)$.  Moreover, the maximizing
        distribution $\mu^*$ is the Dirac measure
        $\mu^* = \delta_{w^*}$ where
        $w^* = \arg\max_{w' \in \mathcal{B}_k(w)} \hat{\mathcal{L}}(w')$.
    \end{proposition}
\end{mybox}

\begin{proof}
    For any $\mu \in \Delta(\mathcal{B}_k(w))$,
    $\mathbb{E}_{w'\sim\mu}[\hat{\mathcal{L}}(w')]
        \leq \max_{w' \in \mathcal{B}_k(w)} \hat{\mathcal{L}}(w')
        = \mathcal{L}_{\mathrm{adv}}(w)$.  Equality is achieved by $\mu^* =
        \delta_{w^*}$, which is a valid element of $\Delta(\mathcal{B}_k(w))$
    since $w^* \in \mathcal{B}_k(w)$.
\end{proof}

This perspective gives a direct connection to the robustness literature.
Hamming-SAM guards against the \emph{worst-case realization} within the ambiguity
set $\mathcal{B}_k(w)$.  In the continuous DRO framework, this set corresponds to a
Wasserstein ball, which is here replaced by its natural
discrete counterpart, the Hamming ball.  Given a uniform deviation bound between
empirical and population risk on the ball, the adversarial loss upper-bounds the
population risk of the minimizer.

\begin{mybox}
    \begin{theorem}[DRO Generalization via Hamming Ambiguity Sets]
        \label{thm:dro_gen}
        Let $\hat{w}$ minimize $\mathcal{L}_{\mathrm{adv}}(w)$
        over $\{-1,+1\}^n$, and let $\mathcal{L}(w) = \mathbb{E}_{(x,y)\sim
            \mathcal{P}}[\ell(f_w(x), y)]$ denote the population risk.  Suppose
        the sample loss satisfies
        $|\hat{\mathcal{L}}(w) - \mathcal{L}(w)| \leq \epsilon_m$ uniformly
        over $\mathcal{B}_k(\hat{w})$ with probability at least $1-\delta$.
        Then:
        \begin{equation}
            \mathcal{L}(\hat{w}) \;\leq\;
            \mathcal{L}_{\mathrm{adv}}(\hat{w}) + \epsilon_m
            \;\leq\; \min_{w \in \{-1,+1\}^n}
            \mathcal{L}_{\mathrm{adv}}(w) + \epsilon_m.
            \label{eq:dro_gen}
        \end{equation}
        In particular, if there exists a ``true'' parameter $w^{\star}$ with
        $\mathcal{L}(w') \leq \epsilon^{\star}$ for all
        $w' \in \mathcal{B}_k(w^{\star})$, then
        $\mathcal{L}(\hat{w}) \leq \epsilon^{\star} + 2\epsilon_m$.
    \end{theorem}
\end{mybox}

\begin{proof}
    The first inequality is immediate:
    $\mathcal{L}(\hat{w}) = \hat{\mathcal{L}}(\hat{w}) + [\mathcal{L}(\hat{w})
        - \hat{\mathcal{L}}(\hat{w})]
        \leq \hat{\mathcal{L}}(\hat{w}) + \epsilon_m
        \leq \mathcal{L}_{\mathrm{adv}}(\hat{w}) + \epsilon_m$, since
    $\hat{w} \in \mathcal{B}_k(\hat{w})$.  The second follows because
    $\hat{w}$ minimizes $\mathcal{L}_{\mathrm{adv}}$.

    For the ``true parameter'' claim, note $\mathcal{L}_{\mathrm{adv}}(w^{\star}) =
        \max_{w' \in \mathcal{B}_k(w^{\star})} \hat{\mathcal{L}}(w') \leq \max_{w' \in
            \mathcal{B}_k(w^{\star})} [\mathcal{L}(w') + \epsilon_m] \leq \epsilon^{\star}
        + \epsilon_m$. Since $\hat{w}$ minimizes $\mathcal{L}_{\mathrm{adv}}$,
    $\mathcal{L}_{\mathrm{adv}}(\hat{w}) \leq \mathcal{L}_{\mathrm{adv}}(w^{\star})
        \leq \epsilon^{\star} + \epsilon_m$. Combining: $\mathcal{L}(\hat{w}) \leq
        \epsilon^{\star} + 2\epsilon_m$.
\end{proof}

\section{Choosing the Hamming Radius $k$: Full Analysis}
\label{app:choosing_k}

The Hamming radius $k$ is the central hyperparameter of Hamming-SAM, controlling
both the strength of the adversarial probe and the tightness of the
generalization guarantee. We now provide a formal analysis of its effect on the
generalization bound, derive the optimal $k$ in closed form for the compression
bound, and give practical guidelines.

\subsection{The Bias--Sharpness Decomposition}

The generalization bounds of \cref{sec:method} all share a common structure.
The population loss is bounded by the sum of a \emph{sharpness term} (which
increases with $k$) and a \emph{complexity term} (which decreases with $k$). To
make this tradeoff explicit, we decompose the compression
bound~\eqref{eq:compression_bound} as:

\begin{equation}
    \mathbb{E}_{w' \sim Q_{\hat{w}}}\bigl[\mathcal{L}(w')\bigr] \;\leq\;
    \underbrace{\textcolor{BrickRed}{\mathcal{L}_{\mathrm{adv}}(\hat{w};k)}}_{\displaystyle
    \textcolor{BrickRed}{\text{sharpness: }\uparrow\text{ in }k}}
    \;+\;
    \underbrace{\textcolor{NavyBlue}{\sqrt{\frac{n\log 2
                - \log V_k + \log(2\sqrt{m}/\delta)}{2m}}}}_{\displaystyle
        \textcolor{NavyBlue}{\text{complexity: }\downarrow\text{ in }k}},
    \label{eq:k_decomposition}
\end{equation}
where $V_k = \sum_{j=0}^{k}\binom{n}{j}$ is the Hamming ball volume,
$Q_{\hat{w}}$ is uniform on $\mathcal{B}_k(\hat{w})$,
and we write $\mathcal{L}_{\mathrm{adv}}(\hat{w};k)$ to emphasize the
dependence on $k$.

\paragraph{Small $k$ (under-regularization).}
When $k = 0$, the Hamming ball is a singleton and
$\mathcal{L}_{\mathrm{adv}}(w; 0) = \hat{\mathcal{L}}(w)$, so Hamming-SAM reduces to
standard empirical risk minimization. The complexity term is $\sqrt{n\log 2 /
        (2m)}$, which is vacuous for $n \gg m$. The optimizer can find sharp minima
that happen to have low training loss but generalize poorly.

\paragraph{Large $k$ (over-regularization).}
As $k$ grows, $\mathcal{L}_{\mathrm{adv}}(w;k)$ increases because the adversary
can search over more configurations. In the extreme $k = n$, the Hamming ball
is the entire hypercube and $\mathcal{L}_{\mathrm{adv}}(w;n) = \max_{w'}
    \hat{\mathcal{L}}(w')$ for all $w$, so the objective becomes constant and
uninformative. The complexity term vanishes ($\log V_n = n\log 2$), but the
sharpness term is maximally pessimistic.

\subsection{Optimal $k$ from the Compression Bound}

To find the value of $k$ that minimizes the right-hand side
of~\eqref{eq:k_decomposition}, we must model how
$\mathcal{L}_{\mathrm{adv}}(w;k)$ grows with $k$. We consider a natural
``smoothness'' assumption on the discrete loss landscape.

\begin{repeatenv}{definition}{\ref{def:smooth_landscape}}[$(L, \sigma)$-Smooth Boolean Landscape]
    The empirical risk $\hat{\mathcal{L}}$ is $(L, \sigma)$-smooth on the
  Boolean hypercube if, for all $w \in \{-1,+1\}^n$ and all $k \geq 1$,
  $\mathcal{L}_{\mathrm{adv}}(w; k) - \hat{\mathcal{L}}(w) \leq L \cdot k^\sigma$,
  where $L > 0$ and $\sigma > 0$.
\end{repeatenv}

The exponent $\sigma$ captures the local geometry.  For $\sigma < 1$, the
worst-case loss grows sublinearly, because most additional flips hit redundant
directions.  For $\sigma > 1$, it grows superlinearly because of
positive interactions, as in the pairwise term $\binom{k}{2}h$ of
\cref{rem:ladv}, which gives $\sigma = 2$.

\begin{mybox}
  \begin{repeatenv}{proposition}{\ref{prop:optimal_k}}[Approximate
    Optimal $k$ for the Compression Bound]
    Under $(L, \sigma)$-smoothness with $k \ll n$, the leading-order
    approximation~\eqref{eq:rhs_approx} of the right-hand side
    of~\eqref{eq:k_decomposition} has the stationary point
    \[
        k^* \approx
        \begin{cases}
        \displaystyle
        \left(\frac{\log n}{2L\sigma\sqrt{2mn\log 2}}\right)^{\frac{1}{\sigma-1}},
        & \text{if } \sigma \neq 1, \\[10pt]
        \displaystyle
        n\exp\!\left(-2L\sqrt{2mn\log 2}\right),
        & \text{if } \sigma = 1.
        \end{cases}
    \]
    For $\sigma \geq 1$ it is a local minimum.  For $\sigma < 1$ it is a local
    maximum whenever $\log(n/k^*) > 1 + 1/(1-\sigma)$, and the bound is then
    minimized at the boundary of the admissible range of $k$.
  \end{repeatenv}
\end{mybox}

\begin{proof}
    For $k \ll n$, Stirling's approximation (see, e.g., \citep{Feller1968}) gives
    $\log V_k = \log\sum_{j=0}^{k}\binom{n}{j} \approx k\log(n/k)$
    (dominated by the largest term $\binom{n}{k}$).  Substituting the
    smoothness assumption into~\eqref{eq:k_decomposition}:
    \begin{equation}
        \mathrm{RHS}(k)
        \approx \hat{\mathcal{L}}(w) + Lk^\sigma
        + \sqrt{\frac{n\log 2 - k\log(n/k)}{2m}}.
        \label{eq:rhs_approx}
    \end{equation}
    Differentiating with respect to $k$ and setting to zero:
    \[
        \frac{d}{dk}\mathrm{RHS}(k)
        = L\sigma k^{\sigma - 1}
        - \frac{\log(n/k) - 1}{2\sqrt{2m(n\log 2 - k\log(n/k))}}
        = 0.
    \]
    For the leading-order behavior (neglecting the $-1$ and the $k\log(n/k)$ in the
    denominator relative to $n\log 2$):
    \[
        L\sigma k^{\sigma - 1}
        \approx \frac{\log(n/k)}{2\sqrt{2mn\log 2}}.
    \]
    Solving for $k$ with $\log(n/k) \approx \log n$ for $k \ll n$:
\begin{equation}
  k^{\sigma-1}
  \;=\; \frac{\log n}{2L\sigma\sqrt{2mn\log 2}}.
\end{equation}
For $\sigma \neq 1$, raising both sides to the power
$1/(\sigma - 1)$:
\begin{equation}
  k^* \;\approx\;
  \left(
    \frac{\log n}
         {2L\sigma\sqrt{2mn\log 2}}
  \right)^{\!1/(\sigma-1)}.
  \label{eq:optimal_k_derived}
\end{equation}
\end{proof}

For $\sigma = 1$ the first-order condition reads
$L = (\log(n/k)-1)/(2\sqrt{2mn\log 2})$, whose solution is the stated value at
leading order.  For the second-order behavior,
$\mathrm{RHS}''(k) \approx L\sigma(\sigma-1)k^{\sigma-2} + 1/(2k\sqrt{2mn\log 2})$.
Substituting the first-order condition
$L\sigma k^{\sigma-1} = (\log(n/k)-1)/(2\sqrt{2mn\log 2})$ gives
\[
  \mathrm{RHS}''(k^*) \approx
  \frac{(\sigma-1)\bigl(\log(n/k^*)-1\bigr) + 1}{2k^*\sqrt{2mn\log 2}},
\]
which is positive for $\sigma \geq 1$ (since $\log(n/k^*) > 1$ when $k^* \ll n$)
and negative for $\sigma < 1$ under the stated condition.

Since $L$ and $\sigma$ are landscape-dependent quantities that are not known
a priori, this formula serves as a qualitative description of the trade-off
rather than a numerical prescription.  The complexity term itself suggests a
scale-free parametrization of the radius.  For $k = \alpha n$ with
$\alpha \leq 1/2$, the standard entropy bounds on binomial sums give
$\log V_k = n\,H(\alpha) + O(\log n)$, where $H$ is the binary entropy in nats,
so the Hamming ball removes a fraction $H(\alpha)/\log 2$ of the description
length of $\{-1,+1\}^n$ regardless of the network size.  This motivates setting
$k = \lfloor \alpha \cdot n \rfloor$ when applying the method across
model sizes (\cref{sec:choosing_k}).

\subsection{Volume of the Hamming Ball}

The combinatorial structure of the Hamming ball plays a central role in the
analysis. The following bounds are useful for understanding the behavior of
$k$.

\begin{mybox}
    \begin{lemma}[Hamming Ball Volume Bounds]
        \label{lem:hamming_volume}
        For $1 \leq k \leq n/2$:
        \begin{equation}
            \left(\frac{n}{k}\right)^k
            \;\leq\; V_k = \sum_{j=0}^{k}\binom{n}{j}
            \;\leq\; \left(\frac{en}{k}\right)^k.
            \label{eq:volume_bounds}
        \end{equation}
        In particular, $\log V_k = k\log(n/k) + \Theta(k)$.
    \end{lemma}
\end{mybox}

\begin{proof}
    For the upper bound, take any $x \in (0,1]$ and note that
    $x^{j-k} \geq 1$ for every $j \leq k$, so
    \[
      V_k \;=\; \sum_{j=0}^{k}\binom{n}{j}
      \;\leq\; \sum_{j=0}^{n}\binom{n}{j} x^{j-k}
      \;=\; x^{-k}(1+x)^n .
    \]
    Choosing $x = k/n$ and using $(1+k/n)^n \leq e^{k}$ gives
    $V_k \leq (n/k)^k e^k = (en/k)^k$.  The lower bound follows from
    $V_k \geq \binom{n}{k} \geq (n/k)^k$.
\end{proof}

This lemma shows how the complexity saving grows with $k$. For
$k = 1$ (single-bit perturbation), it is merely $\log n$
nats. For $k = \sqrt{n}$ it is
$\Theta(\sqrt{n}\log\sqrt{n})$, larger but still a vanishing fraction of
$n\log 2$. Only for $k = \Theta(n)$ does the saving become comparable to the
total.  At $k = n/2$,
$\log V_k \approx n\log 2 - O(1)$, nearly eliminating the complexity term,
though $\mathcal{L}_{\mathrm{adv}}$ is then uninformative.
The ``sweet spot'' lies in the range $k = \Theta(\sqrt{n})$ to $k =
    \Theta(n^{2/3})$, where the complexity reduction is large enough to matter but
the Hamming ball is still small enough that $\mathcal{L}_{\mathrm{adv}}$
remains informative.

\subsection{Relationship to Continuous SAM's $\rho$}

The Hamming radius $k$ is the Boolean analogue of the perturbation radius
$\rho$ in continuous SAM~\citep{foret2021sam}. The following proposition makes
this correspondence quantitative.

\begin{mybox}
    \begin{proposition}[Correspondence Between $k$ and $\rho$]
        \label{prop:k_rho}
        Consider a continuous weight vector $\tilde{w} \in \mathbb{R}^n$ and
        its binarization $w = \mathrm{sign}(\tilde{w})$.  A perturbation
        $\epsilon$ with $\|\epsilon\|_2 = \rho$ flips, on average,
        \begin{equation}
            \mathbb{E}\bigl[d_H(
                \mathrm{sign}(\tilde{w}),
                \mathrm{sign}(\tilde{w}+\epsilon)
                )\bigr]
            \;=\; \sum_{i=1}^{n}
            \Phi\!\left(\frac{-|\tilde{w}_i|}{\rho/\sqrt{n}}\right),
            \label{eq:expected_flips}
        \end{equation}
        where $\Phi$ is the standard normal CDF and $\epsilon$ is drawn
        uniformly on the $\ell_2$-sphere of radius $\rho$.

        For the special case of ``marginally confident'' weights $|\tilde{w}_i| = c$
        for all $i$ (constant magnitude), this simplifies to:
        \begin{equation}
            k_{\mathrm{equiv}} \;=\; n \cdot \Phi\!\left(
            \frac{-c\sqrt{n}}{\rho}\right)
            \;\approx\; n \cdot \frac{\rho}{c\sqrt{2\pi n}}
            \exp\!\left(-\frac{c^2 n}{2\rho^2}\right)
            \label{eq:k_equiv}
        \end{equation}
        for $c\sqrt{n} \gg \rho$ (the ``confident weight'' regime where
        continuous SAM suffers from the vanishing perturbation effect).
    \end{proposition}
\end{mybox}

\begin{proof}
    Each coordinate $\epsilon_i$ of a uniform point on the $\ell_2$-sphere
    has marginal distribution
    $\mathcal{N}(0, \rho^2/n)$.  Bit $i$ flips when
    $\mathrm{sign}(\tilde{w}_i + \epsilon_i) \neq
        \mathrm{sign}(\tilde{w}_i)$, which requires $|\epsilon_i| >
        |\tilde{w}_i|$ with opposite sign.  This has probability
    $\Phi(-|\tilde{w}_i|/(\rho/\sqrt{n}))$.  Summing over $i$ and using
    linearity of expectation gives~\eqref{eq:expected_flips}.  The
    Gaussian tail approximation
    $\Phi(-x) \approx \phi(x)/x$ for $x \gg 1$
    yields~\eqref{eq:k_equiv}.
\end{proof}

\begin{remark}[The vanishing perturbation effect, quantified]
    Equation~\eqref{eq:k_equiv} shows the exponential decay of
    $k_{\mathrm{equiv}}$ in the ratio $c\sqrt{n}/\rho$.  For a typical
    trained network with $c = 2$ (moderate confidence), $n = 10^6$, and
    the standard SAM choice $\rho = 0.05$, we obtain $c\sqrt{n}/\rho = 2000/0.05
        = 40{,}000$, giving $k_{\mathrm{equiv}} \approx 0$, so continuous SAM
    flips essentially zero bits.  In contrast, Hamming-SAM with $k = 100$
    always probes a 100-dimensional subspace.  This makes precise the
    vanishing perturbation argument of \cref{sec:pathology}.
\end{remark}

\subsection{Practical Guidelines}

Based on the theoretical analysis above and preliminary experiments, we
recommend the following approach for setting $k$:

\begin{enumerate}
     \item \textbf{Starting point:}
          A radius between $10^{-5}n$ and $10^{-3}n$ is a reasonable starting
          range for tuning.  Our experiments use a fixed grid of absolute radii
          $\{50, 100, 500, 1000\}$.

    \item \textbf{Validation-based tuning:}
          Perform a grid search, e.g.\
          $k \in \{n/100000,\; n/10000,\; n/1000\}$ (i.e.,
          $\alpha \in \{10^{-5}, 10^{-4}, 10^{-3}\}$), monitoring
          the generalization gap on a held-out set.  The optimal $k$
          typically lies in the range where the sharpness $B_k =
              \mathcal{L}_{\mathrm{adv}}(w;k) - \hat{\mathcal{L}}(w)$ is
          non-trivial (say, $B_k \in [0.01, 0.5]$ for cross-entropy loss)
          but not so large that the adversarial loss dominates.
          In our experiments we instead search a fixed grid of absolute radii,
          $k \in \{50, 100, 500, 1000\}$, which spans $\alpha$ from roughly
          $3\times10^{-6}$ to $2\times10^{-3}$ across the models considered.

    \item \textbf{Sharpness-calibrated $k$:}
          Monitor $B_k$ during training.  If $B_k < 0.01$ persistently, $k$
          is too small and the optimizer is not probing enough of the landscape.
          If $B_k > 1.0$ persistently, $k$ is too large and the adversarial
          probe is finding configurations far outside the relevant basin.
          Adjust $k$ to maintain $B_k$ in a moderate range, or use the
          adaptive schedule $k(t) = \max(1, \lfloor k_0 e^{-\lambda t}\rfloor)$.
\end{enumerate}

\section{Details on the Connection to the Discrete Interaction Hessian}
\label{sec:hessian_app}

A second-order analysis gives more insight into the curvature of the
discrete loss landscape. In continuous optimization, sharpness is characterized
by the eigenvalues of the Hessian
$\nabla^2\mathcal{L}$~\citep{eigenvalue_sam2024}. For Boolean networks, the
analogous object is the \emph{discrete interaction Hessian}, whose entries
capture the pairwise interaction between bit-flips.

\begin{definition}[Discrete Interaction Hessian]
    \label{def:hessian}
    For an empirical risk $\hat{\mathcal{L}}$ and Boolean weight vector
    $w \in \{-1,+1\}^n$, the entry $H_{ij}(w)$ is defined as:
    \begin{equation}
        H_{ij}(w) = \hat{\mathcal{L}}(w^{(ij)}) - \hat{\mathcal{L}}(w^{(i)})
        - \hat{\mathcal{L}}(w^{(j)}) + \hat{\mathcal{L}}(w),
        \label{eq:discrete_hessian}
    \end{equation}
    where $w^{(i)}$ denotes $w$ with the $i$-th bit flipped, and
    $w^{(ij)}$ denotes $w$ with both bits $i$ and $j$ flipped.
\end{definition}

This quantity is the second-order flip difference of a pseudo-Boolean function,
a classical object in combinatorial
optimization~\citep{crama2011boolean, boros2002pseudo}.  In terms of the partial
derivative,
$H_{ij}(w) = w_iw_j\bigl[f'_i(w|_{w_j=+1}) - f'_i(w|_{w_j=-1})\bigr]$, that is,
$w_iw_j$ times the second-order partial derivative.  At $w^t$ it is the
pairwise M\"obius coefficient of the set function $f$ (\cref{lem:mobius}).  Its
average over all contexts $S \subseteq [n]\setminus\{i,j\}$ is the Banzhaf
interaction index of the pair~\citep{grabisch1999axiomatic}, which equals
$4\,w^t_iw^t_j\,\widehat{\mathcal{L}}(\{i,j\})$, with $\widehat{\mathcal{L}}(\{i,j\})$
the Walsh--Hadamard coefficient in the normalization
of~\cite{odonnell2014analysis}.  The interaction Hessian is the pointwise
quantity, and it coincides with these global coefficients only when
$\hat{\mathcal{L}}$ has no interactions of order three or more involving $i$
and $j$.

The following lemma clarifies the relationship between the discrete Hessian of
$\hat{\mathcal{L}}$ and the second-order structure of
the risk-increase set function $f$ of \cref{sec:ascent}.

\begin{mybox}
    \begin{lemma}[Hessian of $\hat{\mathcal{L}}$ as Second-Order Derivative of $f$]
        \label{lem:hessian_bridge}
        Let $f(S) = \hat{\mathcal{L}}(w^t \oplus \mathbf{1}_S) -
            \hat{\mathcal{L}}(w^t)$ be the risk-increase set function.
        For any $S \subseteq [n]$ and $i, j \notin S$ with $i \neq j$, the
        second-order discrete derivative of $f$ at context $S$ equals the
        interaction Hessian of $\hat{\mathcal{L}}$ evaluated at
        $u = w^t \oplus \mathbf{1}_S$:
        \begin{equation}
            \Delta_{ij} f(S) \;\triangleq\;
            f(S \cup \{i,j\}) - f(S \cup \{i\}) - f(S \cup \{j\}) + f(S)
            \;=\; H_{ij}(w^t \oplus \mathbf{1}_S).
            \label{eq:hessian_bridge}
        \end{equation}
    \end{lemma}
\end{mybox}

\begin{proof}
    Substituting the definition $f(S) = \hat{\mathcal{L}}(w^t \oplus
        \mathbf{1}_S) - \hat{\mathcal{L}}(w^t)$ into each of the four terms:
    \begin{align}
        \Delta_{ij} f(S)
         & = \bigl[\hat{\mathcal{L}}(w^t \oplus \mathbf{1}_{S\cup\{i,j\}})
            - \hat{\mathcal{L}}(w^t)\bigr]
        - \bigl[\hat{\mathcal{L}}(w^t \oplus \mathbf{1}_{S\cup\{i\}})
        - \hat{\mathcal{L}}(w^t)\bigr] \nonumber                           \\
         & \qquad
        - \bigl[\hat{\mathcal{L}}(w^t \oplus \mathbf{1}_{S\cup\{j\}})
            - \hat{\mathcal{L}}(w^t)\bigr]
        + \bigl[\hat{\mathcal{L}}(w^t \oplus \mathbf{1}_S)
            - \hat{\mathcal{L}}(w^t)\bigr].
    \end{align}
    The four copies of $-\hat{\mathcal{L}}(w^t)$ cancel.  Writing
    $u = w^t \oplus \mathbf{1}_S$ and noting that
    $w^t \oplus \mathbf{1}_{S \cup \{i\}} = u^{(i)}$,
    $w^t \oplus \mathbf{1}_{S \cup \{j\}} = u^{(j)}$, and
    $w^t \oplus \mathbf{1}_{S \cup \{i,j\}} = u^{(ij)}$:
    \[
        \Delta_{ij} f(S) = \hat{\mathcal{L}}(u^{(ij)})
        - \hat{\mathcal{L}}(u^{(i)}) - \hat{\mathcal{L}}(u^{(j)})
        + \hat{\mathcal{L}}(u) = H_{ij}(u).  \qedhere
    \]
\end{proof}

This identity shows that $H_{ij}$ of the empirical loss, evaluated at the
``context point'' $w^t \oplus \mathbf{1}_S$, is exactly the pairwise
interaction of $f$ at context $S$. The constant $\hat{\mathcal{L}}(w^t)$ in the
definition of $f$ drops out because second-order differences annihilate
additive constants.  The same structure reappears in \cref{app:ascent}, where
the M\"obius expansion of $f$ separates its first- and second-order terms and
thereby bounds the deficit of the one-shot top-$k$ ascent.
The connection to submodularity follows directly.

\begin{mybox}
    \begin{proposition}[Hessian--Submodularity Connection]
        \label{prop:hessian_submod}
        The set function $f(S) = \hat{\mathcal{L}}(w^t \oplus \mathbf{1}_S) - \hat{\mathcal{L}}(w^t)$ is submodular if and only if $H_{ij}(w^t \oplus \mathbf{1}_S) \leq 0$ for all
        $S \subseteq [n]$ and all $i, j \notin S$ with $i \neq j$.
        Equivalently, negative off-diagonal entries correspond to diminishing
        returns (flipping both bits gives less loss increase than the sum of
        individual flips), and positive entries correspond to reinforcing
        interactions (flipping both gives more).
    \end{proposition}
\end{mybox}

\begin{proof}
    The submodularity condition requires
    $f(S \cup \{i\}) - f(S) \geq f(S \cup \{i,j\}) - f(S \cup \{j\})$
    for all $S$, $i$, $j$.  Rearranging:
    $f(S \cup \{i,j\}) - f(S \cup \{i\}) - f(S \cup \{j\}) + f(S) \leq 0$.
    By Lemma~\ref{lem:hessian_bridge}, the left-hand side equals
    $H_{ij}(w^t \oplus \mathbf{1}_S)$.
\end{proof}

\begin{figure}[t]
    \centering

    \tikzexternaldisable

    \tikzexternaldisable

\definecolor{cFlat}{rgb}{0.0, 0.6, 0.5}   %
\definecolor{cSharp}{rgb}{0.5, 0.2, 0.7}  %
\definecolor{cMid}{rgb}{0.9, 0.6, 0.1}    %
\definecolor{cGray}{rgb}{0.4, 0.4, 0.4}   %

\begin{tikzpicture}[>=Stealth, font=\small,
      nd/.style={draw, circle, thick, minimum size=24pt,
                 inner sep=0pt, font=\footnotesize\bfseries}]
    \node[anchor=south, font=\bfseries\footnotesize] at (2.0,3.5)
      {Submodular ($H_{12} < 0$)};
    \node[anchor=south, font=\footnotesize, cFlat!70!black] at (2.0,3.05)
      {diminishing returns};
    \node[nd, fill=cFlat!40]  (a00) at (0,0)   {.20};
    \node[nd, fill=cMid!40]   (a10) at (4,0)   {.50};
    \node[nd, fill=cMid!40]   (a01) at (0,2.4) {.45};
    \node[nd, fill=cMid!55]   (a11) at (4,2.4) {.60};
    \draw[thick] (a00)--(a10) node[midway, below, font=\tiny]{flip bit 1: $+.30$};
    \draw[thick] (a00)--(a01) node[midway, left, font=\tiny]{flip bit 2: $+.25$};
    \draw[thick] (a10)--(a11) node[midway, right, font=\tiny]{flip bit 2: $+.10$};
    \draw[thick] (a01)--(a11) node[midway, above, font=\tiny]{flip bit 1: $+.15$};
    \node[font=\tiny, cGray, yshift=-2pt] at (0,-0.55) {$w$};
    \node[font=\tiny, cGray, yshift=-2pt] at (4,-0.55) {$w^{(1)}$};
    \node[font=\tiny, cGray, yshift=2pt] at (0,2.95)  {$w^{(2)}$};
    \node[font=\tiny, cGray, yshift=2pt] at (4,2.95)  {$w^{(12)}$};
    \node[draw, rounded corners, fill=cFlat!8, font=\tiny, align=left]
      at (2.0,-1.4) {$H_{12} = .60 - .50 - .45 + .20 = \mathbf{-.15}$\\
        Flipping both gives \emph{less} than sum of individual gains};

    \begin{scope}[xshift=7.5cm]
      \node[anchor=south, font=\bfseries\footnotesize] at (2.0,3.5)
        {Supermodular ($H_{12} > 0$)};
      \node[anchor=south, font=\footnotesize, cSharp!70!black] at (2.0,3.05)
        {synergistic interaction};
      \node[nd, fill=cFlat!40]  (b00) at (0,0)   {.20};
      \node[nd, fill=cMid!25]   (b10) at (4,0)   {.30};
      \node[nd, fill=cMid!25]   (b01) at (0,2.4) {.28};
      \node[nd, fill=cSharp!55] (b11) at (4,2.4) {.78};
      \draw[thick] (b00)--(b10) node[midway, below, font=\tiny]{flip bit 1: $+.10$};
      \draw[thick] (b00)--(b01) node[midway, left, font=\tiny]{flip bit 2: $+.08$};
      \draw[thick] (b10)--(b11) node[midway, right, font=\tiny, xshift=-45pt]{flip bit 2: $+.48$};
      \draw[thick] (b01)--(b11) node[midway, above, font=\tiny]{flip bit 1: $+.50$};
      \node[font=\tiny, cGray, yshift=-2pt] at (0,-0.55) {$w$};
      \node[font=\tiny, cGray, yshift=-2pt] at (4,-0.55) {$w^{(1)}$};
      \node[font=\tiny, cGray, yshift=2pt] at (0,2.95)  {$w^{(2)}$};
      \node[font=\tiny, cGray, yshift=2pt] at (4,2.95)  {$w^{(12)}$};
      \node[draw, rounded corners, fill=cSharp!8, font=\tiny, align=left]
        at (2.0,-1.4) {$H_{12} = .78 - .30 - .28 + .20 = \mathbf{+.40}$\\
          Flipping both gives \emph{more} than sum of individual gains};
    \end{scope}
  \end{tikzpicture}

    \tikzexternalenable

    \caption{\small\textbf{Discrete interaction Hessian on $\{-1,+1\}^2$.}
        Each vertex shows a loss value.  \emph{Left:}~Submodular case
        ($H_{12}{=}{-}0.15$): the marginal gain of flipping bit~2 decreases
        from $0.25$ to $0.10$ after bit~1 is flipped (diminishing returns).
        \emph{Right:}~Supermodular case ($H_{12}{=}{+}0.40$): the marginal
        gain of flipping bit~2 \emph{increases} from $0.08$ to $0.48$ after
        bit~1 is flipped.  In this case, the greedy ascent can
        underestimate the worst case.}
    \label{fig:discrete_hessian}

    \vspace{-2.5ex}
\end{figure}
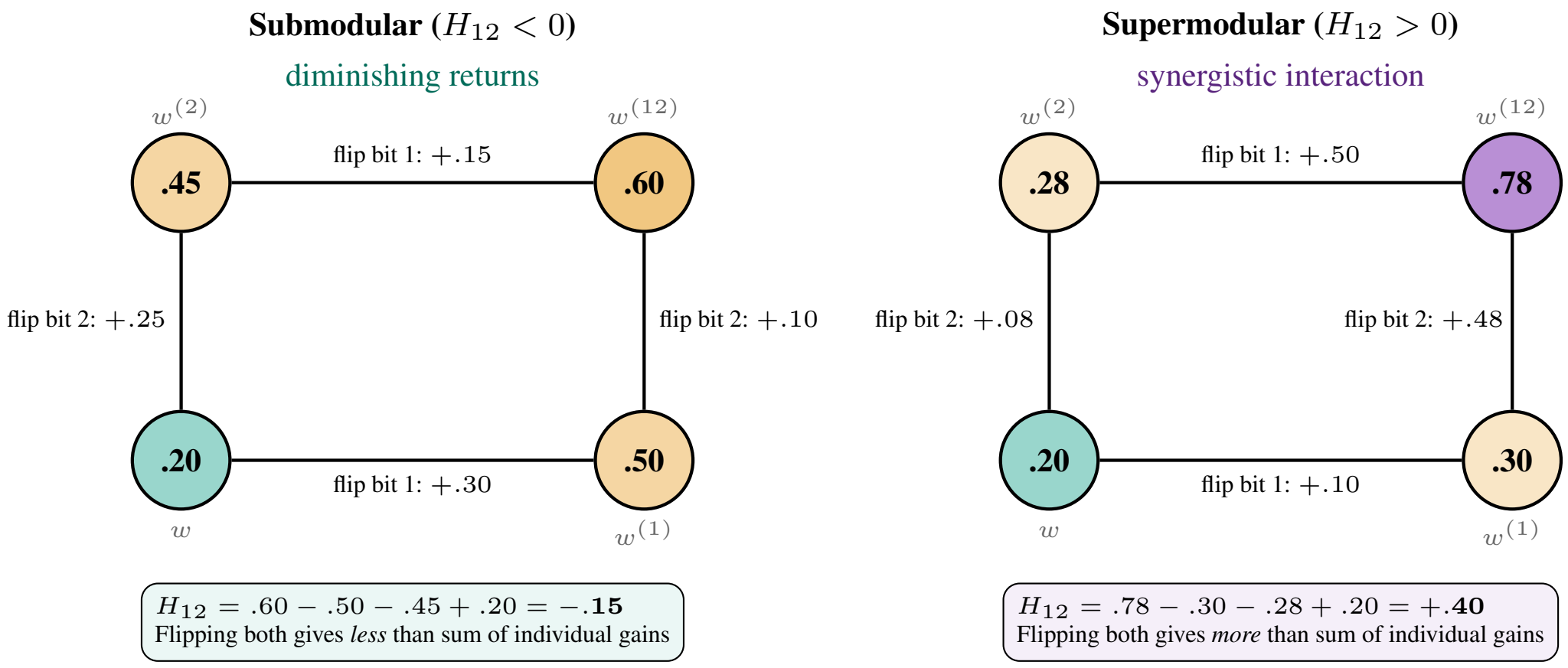

If $H_{ij}(w) = 0$ for all $i \neq j$, the loss function is modular and weights
contribute independently. Positive values indicate reinforcing flips,
and negative values indicate redundancy. Hamming-SAM
convergence regions should therefore tend to have lower $|H_{ij}|$ than typical
points, a prediction supported by the convergence condition $\eta > kh$
(\cref{thm:convergence_sam}), which favors regions with small $h$, and by the
connection to $\gamma_k$ in the ascent guarantee.  \cref{sec:exp_landscape}
and \cref{app:theory_verification} confirm this: on every architecture
measured, the sampled $|H_{ij}|$ statistics at Hamming-SAM solutions lie below
those at BOLD solutions at convergence.

\paragraph{Spectral Gap of the Hamming Graph.}
The flatness of a discrete minimum can further be quantified by the spectral
gap of the graph Laplacian restricted to the loss-weighted Hamming
neighborhood. The hypercube $\{-1,+1\}^n$ is a regular graph where nodes are
configurations and edges connect those with Hamming distance one. It is well
known that the eigenvalues of the hypercube Laplacian are $\{2j : j = 0, 1,
    \ldots, n\}$ with multiplicities $\binom{n}{j}$~\citep{chung1997spectral}. A
large spectral gap in the local subgraph weighted by the loss indicates a
well-connected region where the loss changes slowly, which gives faster mixing
and better robustness. By minimizing the adversarial loss in
$\mathcal{B}_k(w)$, Hamming-SAM implicitly regularizes for these spectral
properties.

\section{Detailed Proof of Approximation Guarantees for the Ascent Step}
\label{app:ascent}

A key requirement for the Hamming-SAM framework is a
guarantee that the greedy ascent identifies a near-worst-case
configuration within $\mathcal{B}_k(w)$.  In continuous SAM, the ascent
is justified as a first-order Taylor
approximation~\citep{foret2021sam}.  In the discrete domain, the
quality of the greedy search depends on the approximate submodularity of
the risk-increase function.

The problem of finding $\hat{w}$ is a cardinality-constrained
maximization of the non-negative set function
\begin{equation}
  f(S) = \hat{\mathcal{L}}(w^t \oplus \mathbf{1}_S;\mathcal{D})
        - \hat{\mathcal{L}}(w^t;\mathcal{D}),
  \quad S \subseteq [n],\; |S| \leq k,
  \label{eq:set_function}
\end{equation}
where $w^t \oplus \mathbf{1}_S$ denotes the configuration obtained by
flipping all bits indexed by $S$.  Note $f(\emptyset) = 0$ by
definition.  If $f$ were monotone and submodular, the classical greedy
algorithm would achieve an approximation ratio of
$(1 - 1/e)$~\citep{nemhauser1978submodular}.  However, neural networks
exhibit complex interactions, where flipping two bits together may increase
the loss far more than the sum of their individual
effects.  This violates submodularity.

We therefore use the framework of \cite{bian2017guarantees},
which extends greedy guarantees to non-submodular functions using two
parameters:

\begin{repeatenv}{definition}{\ref{def:submod_ratio}}[Submodularity Ratio~\citep{das2018approximate}]
The submodularity ratio $\gamma_k$ of a non-negative set function $f$
with respect to a parameter $k$ is the largest scalar $\gamma$ such
that, for all disjoint sets $L, S \subseteq [n]$ with $|L| \leq k$:
\begin{equation}
  \sum_{j \in L} \bigl[f(S \cup \{j\}) - f(S)\bigr]
  \;\geq\; \gamma \cdot \bigl[f(S \cup L) - f(S)\bigr].
  \label{eq:submod_ratio}
\end{equation}
\end{repeatenv}

\begin{definition}[Generalized Curvature~\citep{bian2017guarantees}]
  \label{def:curvature_app}
  Let $\rho_j(A) = f(A \cup \{j\}) - f(A)$.  The generalized curvature of a
  non-negative set function $f$ with $f(\emptyset) = 0$ is the smallest
  $\alpha \in [0,1]$ such that, for all $S, \Omega \subseteq \mathcal{V}_+$ and
  $i \in S \setminus \Omega$,
  \begin{equation}
    \rho_i\bigl((S\setminus\{i\}) \cup \Omega\bigr) \;\geq\; (1-\alpha)\,\rho_i(S\setminus\{i\}).
  \end{equation}
\end{definition}

Equivalently, $\alpha$ bounds how much the marginal gain of an element can
shrink when other elements are added to its context.  For modular functions,
$\alpha = 0$ (gains are constant).  For submodular functions, $\alpha$ can be as
large as $1$ (gains can vanish).

The submodularity ratio $\gamma \in (0,1]$ quantifies how close $f$ is
to submodular ($\gamma = 1$ for submodular functions).  The curvature
$\alpha \in [0,1]$ measures the maximum rate at which marginal gains
decrease ($\alpha = 0$ for modular functions, $\alpha = 1$ in the worst
case).

\begin{mybox}
  \begin{repeatenv}{theorem}{\ref{thm:ascent}}[Sequential Greedy Ascent Guarantee]
    Let $f$ be the set function defined in~\eqref{eq:set_function}, restricted to
subsets of $\mathcal{V}_+$, with $f(\emptyset) = 0$, and suppose $f$ is
non-negative and non-decreasing (monotone) on these subsets.  Let $\hat{S}$ be
the set of $k$ indices selected by the sequential greedy ascent, and
$S^* = \arg\max_{S \subseteq \mathcal{V}_+,\,|S| \leq k} f(S)$ the optimal set.  Then:
    \begin{equation}
      f(\hat{S}) \;\geq\;
      \frac{1}{\alpha}\left[1 -
      e^{-\gamma_k \alpha}\right] f(S^*).
      \label{eq:ascent_guarantee}
    \end{equation}
    In terms of the adversarial loss:
    \begin{equation}
      \hat{\mathcal{L}}(\hat{w}) - \hat{\mathcal{L}}(w^t)
      \;\geq\;
      \frac{1}{\alpha}\left[1 - e^{-\gamma_k \alpha}\right]
      \bigl[\hat{\mathcal{L}}(w^*) - \hat{\mathcal{L}}(w^t)\bigr],
    \end{equation}
    where $\hat{w} = w^t \oplus \mathbf{1}_{\hat{S}}$ and
    $w^* = w^t \oplus \mathbf{1}_{S^*}$.
  \end{repeatenv}
\end{mybox}

\begin{proof}
Let $f : 2^{[n]} \to \mathbb{R}_{\geq 0}$ be the set function
defined in~\eqref{eq:set_function} with $f(\emptyset) = 0$.
The greedy algorithm constructs a sequence
$S_0 = \emptyset \subset S_1 \subset \cdots \subset S_k$ where
at step $j$, we add
$i_j = \arg\max_{i \notin S_{j-1}}[f(S_{j-1} \cup \{i\}) - f(S_{j-1})]$.

Let $S^* = \arg\max_{|S| \leq k} f(S)$.  By the definition of
the submodularity ratio $\gamma_k$ (\cref{def:submod_ratio})
applied to $L = S^* \setminus S_{j-1}$ and $S = S_{j-1}$
(noting $|L| \leq k$):
\begin{equation}
  \sum_{i \in S^* \setminus S_{j-1}}
    \bigl[f(S_{j-1} \cup \{i\}) - f(S_{j-1})\bigr]
  \;\geq\; \gamma_k\bigl[f(S_{j-1} \cup S^*) - f(S_{j-1})\bigr].
  \label{eq:app_submod_ratio}
\end{equation}
Since $f$ is monotone, $f(S_{j-1} \cup S^*) \geq f(S^*)$, so:
\begin{equation}
  f(S_{j-1} \cup S^*) - f(S_{j-1})
  \;\geq\; f(S^*) - f(S_{j-1}).
  \label{eq:app_monotone}
\end{equation}
The greedy choice $i_j$ achieves at least $1/k$ of the sum
in~\eqref{eq:app_submod_ratio} (since
$|S^* \setminus S_{j-1}| \leq k$).
Combining~\eqref{eq:app_submod_ratio}
and~\eqref{eq:app_monotone}:
\begin{equation}
  f(S_j) - f(S_{j-1})
  \;\geq\; \frac{\gamma_k}{k}\bigl[f(S^*) - f(S_{j-1})\bigr].
  \label{eq:app_greedy_step}
\end{equation}
Define $\delta_j = f(S^*) - f(S_j)$.  Then:
\[
  \delta_j
  \;\leq\; \delta_{j-1} - \frac{\gamma_k}{k}\,\delta_{j-1}
  \;=\; \Bigl(1 - \frac{\gamma_k}{k}\Bigr)\delta_{j-1}.
\]
Iterating $k$ times from $\delta_0 = f(S^*)$:
\[
  \delta_k
  \;\leq\; \Bigl(1 - \frac{\gamma_k}{k}\Bigr)^k f(S^*)
  \;\leq\; e^{-\gamma_k}\,f(S^*).
\]
Therefore:
\begin{equation}
  f(\hat{S}) = f(S_k)
  \;\geq\; \bigl[1 - e^{-\gamma_k}\bigr]\,f(S^*).
  \label{eq:app_simple_bound}
\end{equation}

\paragraph{Tighter bound with curvature $\alpha$.}
The bound~\eqref{eq:app_simple_bound} uses only the
submodularity ratio and monotonicity.  A tighter result is
obtained by additionally incorporating the generalized
curvature $\alpha$ (\cref{def:curvature_app}), which bounds
how much marginal gains decrease relative to singleton gains:
$f(S \cup \{j\}) - f(S) \geq (1-\alpha)\,f(\{j\})$ for all
$S$ and $j \notin S$.  \cite{bian2017guarantees}
show that this yields the refined bound:
\begin{equation}
  f(\hat{S})
  \;\geq\; \frac{1}{\alpha}
    \bigl(1 - e^{-\gamma_k\alpha}\bigr)\,f(S^*),
  \label{eq:app_tight_bound}
\end{equation}
which is strictly tighter
than~\eqref{eq:app_simple_bound} for $\alpha < 1$.
For $\gamma_k = 1$ and $\alpha = 1$ (submodular), this
recovers the classical $(1-1/e)$ bound
of~\cite{nemhauser1978submodular}.  For $\alpha \to 0$
(modular), L'H\^{o}pital's rule gives
$f(\hat{S}) \geq \gamma_k\,f(S^*)$.
We refer to \cite{bian2017guarantees} for the
complete derivation.
\end{proof}

\begin{remark}
  When $\gamma_k = 1$ and $\alpha \to 0$ (the modular case), the bound
  recovers the exact optimum $f(\hat{S}) = f(S^*)$. When $\gamma_k = 1$
  and $\alpha = 1$ (the submodular case), it recovers the classical
  $(1-1/e)$ guarantee of~\cite{nemhauser1978submodular}.
\end{remark}

\paragraph{Assumptions and scope.}
\cref{thm:ascent} requires $f$ to be non-negative and non-decreasing (monotone) on
subsets of $\mathcal{V}_+$.  Every bit in $\mathcal{V}_+$ increases the loss when
flipped alone, but flipping other bits first changes its gain by one
interaction entry per flip, so
$\rho_i(S) = \mathrm{BV}_i(w^t \oplus \mathbf{1}_S) \geq \mathrm{BV}_i(w^t) - |S|\,h$,
with $h$ the interaction bound of \cref{thm:ascent_topk}.  Hence $f$ is monotone on
subsets of $\mathcal{V}_+$ of size at most $K$ whenever
$\min_{i \in \mathcal{V}_+}\mathrm{BV}_i(w^t) > (K-1)h$, which is the precise
form of \cref{rem:monotonicity}.  Monotonicity enters the proof through the
inequality $f(S_{j-1} \cup S^*) \geq f(S^*) + (1-\alpha)\,f(S_{j-1})$, obtained by
adding the elements of the greedy sequence to $S^*$ one at a time.  When $f$ is
not monotone, this step may fail and the ratio to $f(S^*)$ is no longer
guaranteed.  The greedy's own output $f(\hat{S})$ is unaffected.  Empirical
measurement of $\gamma_k$ during training (\cref{sec:exp_landscape}) quantifies
how close practical landscapes are to the submodular regime.

\subsection{Guarantee for the One-Shot Top-$k$ Ascent}
\label{app:topk}

This subsection develops the approximation guarantee for the one-shot
top-$k$ ascent that \cref{alg:hamming_sam} actually implements.  The result is
not a strengthening of \cref{thm:ascent} but a complement to it.
\cref{thm:ascent} bounds the sequential greedy under approximate
submodularity, and is therefore governed by the \emph{sign} structure of the
interaction Hessian, since $\gamma_k = 1$ if and only if $H_{ij} \leq 0$
(\cref{prop:hessian_submod}).  The result below bounds the top-$k$ rule under
a bound on interaction \emph{magnitude}, $|H_{ij}| \leq h$, and requires
neither monotonicity nor submodularity.  Neither statement implies the other.
\cref{sec:ascent} sets out their respective roles, and
\cref{app:sequential_greedy} measures the empirical gap between the two
procedures.

Throughout this subsection we color the
\textcolor{NavyBlue}{first-order (Boolean Variation)},
\textcolor{BrickRed}{pairwise (interaction Hessian)} and
\textcolor{Mulberry}{higher-order residual} contributions consistently, so
that the role of each can be tracked across the expansion, the bound and its
corollaries.

The two procedures also differ in cost.  The sequential greedy re-evaluates
marginal gains after each selection and therefore requires $k$ backward passes
per optimization step, which is intractable at the radii we use.  The top-$k$
rule obtains all $n$ marginal gains in a single backward pass.  The theorem
below quantifies the cost of this speed-up.

\paragraph{Higher-order discrete derivatives.}
For $T \subseteq [n]$ with $|T| \geq 1$, let $\Delta_T(w^t)$ denote the
order-$|T|$ discrete derivative of $\hat{\mathcal{L}}$ at $w^t$, defined by
M\"obius inversion of $f$:
\begin{equation}
  \Delta_T(w^t) \;=\; \sum_{U \subseteq T} (-1)^{|T| - |U|} f(U).
  \label{eq:M\"obius_coeff}
\end{equation}
The first two orders recover the quantities already in use,
$\Delta_{\{i\}}(w^t) = \mathrm{BV}_i(w^t)$ and
$\Delta_{\{i,j\}}(w^t) = H_{ij}(w^t)$, by \cref{def:hessian}.  We write
$h_p = \max_{w,\,|T| = p} |\Delta_T(w)|$ for the largest order-$p$
interaction, so that $h_2 = h$.

\begin{mybox}
  \begin{lemma}[M\"obius Expansion of the Risk-Increase Function]
    \label{lem:mobius}
    For every $S \subseteq [n]$,
    \begin{equation}
      f(S) \;=\;
      \textcolor{NavyBlue}{\sum_{i \in S} \mathrm{BV}_i(w^t)}
      \;+\; \textcolor{BrickRed}{\sum_{\{i,j\} \subseteq S} H_{ij}(w^t)}
      \;+\; \textcolor{Mulberry}{R(S)},
      \qquad
      \textcolor{Mulberry}{|R(S)| \;\leq\; \sum_{p=3}^{|S|} \binom{|S|}{p} h_p} .
      \label{eq:M\"obius_expansion}
    \end{equation}
  \end{lemma}
\end{mybox}

\begin{proof}
  Any set function with $f(\emptyset) = 0$ admits the unique multilinear
  (M\"obius) representation $f(S) = \sum_{\emptyset \neq T \subseteq S}
  \Delta_T(w^t)$ with coefficients given
  by~\eqref{eq:M\"obius_coeff}~\citep{crama2011boolean}.  Partitioning the sum
  by $|T| \in \{1\}$, $\{2\}$ and $\{3,\ldots,|S|\}$ gives the three terms
  of~\eqref{eq:M\"obius_expansion}, where the order-one coefficients are the
  Boolean Variations and the order-two coefficients are the interaction
  Hessian entries.  The residual $R(S)$ collects the remaining
  $\sum_{p \geq 3} \binom{|S|}{p}$ coefficients, each bounded in magnitude by
  $h_p$, and the triangle inequality gives the stated bound.
\end{proof}

\begin{mybox}
  \begin{repeatenv}{theorem}{\ref{thm:ascent_topk}}[Top-$k$ Ascent Guarantee]
    Let $\hat{S}_{\mathrm{tk}}$ be the set selected by the top-$k$ rule of
    \cref{alg:hamming_sam}, restricted to bits with $\mathrm{BV}_i > 0$, and let
    $S^* = \arg\max_{|S| \leq k} f(S)$.  Suppose $|H_{ij}(w)| \leq h$ for all
    $w$ and $i \neq j$, and $|R(S)| \leq R_{\max}$ for all $|S| \leq k$.  Then
    \begin{equation}
      f(\hat{S}_{\mathrm{tk}})
      \;\geq\; f(S^*) \;-\; \textcolor{BrickRed}{k(k-1)\,h}
      \;-\; \textcolor{Mulberry}{2R_{\max}}.
      \label{eq:topk_guarantee}
    \end{equation}
  \end{repeatenv}
\end{mybox}

\begin{proof}
  Apply \cref{lem:mobius} to $S^*$ and to $\hat{S}_{\mathrm{tk}}$ and
  subtract:
  \begin{equation}
    f(S^*) - f(\hat{S}_{\mathrm{tk}})
    = \underbrace{\textcolor{NavyBlue}{\Bigl[\sum_{i \in S^*}\mathrm{BV}_i
        - \sum_{i \in \hat{S}_{\mathrm{tk}}}\mathrm{BV}_i\Bigr]}}_{\textcolor{NavyBlue}{(\mathrm{A})\;\leq\;0}}
    + \underbrace{\textcolor{BrickRed}{\Bigl[\sum_{\{i,j\} \subseteq S^*} H_{ij}
        - \sum_{\{i,j\} \subseteq \hat{S}_{\mathrm{tk}}} H_{ij}\Bigr]}}_{\textcolor{BrickRed}{(\mathrm{B})\;\leq\;k(k-1)h}}
    + \underbrace{\textcolor{Mulberry}{\bigl[R(S^*) - R(\hat{S}_{\mathrm{tk}})\bigr]}}_{\textcolor{Mulberry}{(\mathrm{C})\;\leq\;2R_{\max}}}.
  \end{equation}

  \textcolor{NavyBlue}{\emph{Term (A), first order.}}  Let $P = \{i \in S^* : \mathrm{BV}_i > 0\}$.  Discarding
  the non-positive entries of $S^*$ can only increase the sum, so
  $\sum_{i \in S^*}\mathrm{BV}_i \leq \sum_{i \in P}\mathrm{BV}_i$.  Since
  $|P| \leq |S^*| \leq k$ and $\hat{S}_{\mathrm{tk}}$ consists of the $k$
  largest strictly positive Boolean Variations, $\sum_{i \in
  P}\mathrm{BV}_i \leq \sum_{i \in \hat{S}_{\mathrm{tk}}}\mathrm{BV}_i$.
  Hence $(\mathrm{A}) \leq 0$.  The restriction to $\mathrm{BV}_i > 0$ is
  needed here: without it, a top-$k$ set padded with negative entries could
  have a smaller sum than a strictly smaller optimal set.

  \textcolor{BrickRed}{\emph{Term (B), pairwise.}}  Each expansion contains at most $\binom{k}{2}$ pairwise
  coefficients, each of magnitude at most $h$, so
  $(\mathrm{B}) \leq 2\binom{k}{2}h = k(k-1)h$.

  \textcolor{Mulberry}{\emph{Term (C), higher order.}}  By assumption $|R(S)| \leq R_{\max}$ for both sets, so
  $(\mathrm{C}) \leq 2R_{\max}$ by the triangle inequality.

  Summing the three bounds gives $f(S^*) - f(\hat{S}_{\mathrm{tk}}) \leq
  \textcolor{BrickRed}{k(k-1)h} + \textcolor{Mulberry}{2R_{\max}}$, which
  is~\eqref{eq:topk_guarantee}.
\end{proof}

\begin{remark}[Relation to \cref{thm:ascent}]
  \label{rem:topk_vs_seq}
  The bound is \emph{additive} in the interaction strength, whereas
  \cref{thm:ascent} is \emph{multiplicative} in the submodularity ratio.  The
  two are complementary.  \cref{thm:ascent} certifies a constant fraction of
  the optimum however large $f(S^*)$ is, but needs monotonicity and degrades
  as $\gamma_k \to 0$.  \cref{thm:ascent_topk} needs neither, but is
  informative only while $k(k-1)h$ stays small relative to $f(S^*)$.  The
  ascent is therefore near-optimal whenever \emph{either} the landscape is
  approximately submodular \emph{or} its pairwise interactions are weak.
\end{remark}

\begin{remark}[Size of the residual]
  \label{rem:residual}
  The residual collects interactions of order three and higher and enters
  \cref{thm:ascent_topk} only through the assumed bound $R_{\max}$.  If
  higher-order interactions decay as $h_p \leq h^{p-1}$ for $p \geq 3$, the
  leading residual term is $\binom{k}{3}h^2 = O(k^3h^2)$, which is dominated by
  the pairwise term $k(k-1)h$ whenever $kh \ll 1$.
\end{remark}

\paragraph{Inexact oracles.}
\cref{sec:bv_main} notes that the backward pass returns a first-order
estimate $\widetilde{\mathrm{BV}}_i$ rather than the exact finite
difference.  The guarantee degrades gracefully in the oracle error.

\begin{mybox}
  \begin{proposition}[Top-$k$ Ascent with an Inexact Oracle]
    \label{prop:topk_inexact}
    Suppose the ascent ranks bits by a signal $\widetilde{\mathrm{BV}}$ with
    $\max_i |\widetilde{\mathrm{BV}}_i - \mathrm{BV}_i| \leq \varepsilon$, and
    let $\hat{S}_{\mathrm{tk}}$ be the resulting selection, restricted to bits
    with $\widetilde{\mathrm{BV}}_i > 0$.  Then under the assumptions of
    \cref{thm:ascent_topk},
    \begin{equation}
      f(\hat{S}_{\mathrm{tk}})
      \;\geq\; f(S^*) \;-\; \textcolor{NavyBlue}{2k\varepsilon}
      \;-\; \textcolor{BrickRed}{k(k-1)h} \;-\; \textcolor{Mulberry}{2R_{\max}}.
    \end{equation}
  \end{proposition}
\end{mybox}

\begin{proof}
  Only \textcolor{NavyBlue}{term (A)} changes.  Let $P = \{i \in S^* : \mathrm{BV}_i > 0\}$ and
  $P^{+} = \{i \in P : \widetilde{\mathrm{BV}}_i > 0\}$, so $|P^{+}| \leq |P|
  \leq k$.  Then
  $\sum_{S^*}\mathrm{BV}_i \leq \sum_{P}\mathrm{BV}_i
   \leq \sum_{P}\widetilde{\mathrm{BV}}_i + k\varepsilon
   \leq \sum_{P^{+}}\widetilde{\mathrm{BV}}_i + k\varepsilon
   \leq \sum_{\hat{S}_{\mathrm{tk}}}\widetilde{\mathrm{BV}}_i + k\varepsilon
   \leq \sum_{\hat{S}_{\mathrm{tk}}}\mathrm{BV}_i + 2k\varepsilon$,
  where the first step discards non-positive entries, the second and last use
  the oracle bound on at most $k$ terms, the third discards entries with
  $\widetilde{\mathrm{BV}}_i \leq 0$, and the fourth uses that
  $\hat{S}_{\mathrm{tk}}$ maximizes $\sum \widetilde{\mathrm{BV}}$ over
  positive-signal sets of size at most $k$.  Hence
  $\textcolor{NavyBlue}{(\mathrm{A}) \leq 2k\varepsilon}$, and
  \textcolor{BrickRed}{(B)} and \textcolor{Mulberry}{(C)} are bounded as
  before.
\end{proof}

\paragraph{Controlling the adversarial objective.}
The same expansion bounds the sharpness itself, which is the quantity the
outer minimization targets.

\begin{mybox}
  \begin{repeatenv}{remark}{\ref{rem:ladv}}[Sharpness Composition]
    Under the assumptions of \cref{thm:ascent_topk}, for every $w$,
    \begin{equation}
      \mathcal{L}_{\mathrm{adv}}(w) - \hat{\mathcal{L}}(w)
      \;=\; \max_{|S| \leq k} f(S)
      \;\leq\; \textcolor{NavyBlue}{k\,B_1(w)}
      + \textcolor{BrickRed}{\binom{k}{2}h}
      + \textcolor{Mulberry}{R_{\max}},
      \label{eq:sharpness_composition}
    \end{equation}
    where $B_1(w) = \max_i \max(\mathrm{BV}_i(w), 0)$ is the one-bit
    sharpness.
  \end{repeatenv}
\end{mybox}

\begin{proof}
  Apply \cref{lem:mobius} to an arbitrary $S$ with $|S| \leq k$.  The
  \textcolor{NavyBlue}{first-order} term satisfies $\sum_{i \in S}\mathrm{BV}_i \leq
  \sum_{i \in S}\max(\mathrm{BV}_i, 0) \leq \textcolor{NavyBlue}{k B_1(w)}$, the
  \textcolor{BrickRed}{second-order} term is at most
  $\textcolor{BrickRed}{\binom{k}{2}h}$, and the
  \textcolor{Mulberry}{residual} is at most $\textcolor{Mulberry}{R_{\max}}$.  Taking
  the maximum over $S$ gives~\eqref{eq:sharpness_composition}.
\end{proof}

\noindent We note two consequences.  First, since
\cref{thm:convergence_sam} drives $\hat{\mathcal{L}}(w^t)$ down monotonically,
\eqref{eq:sharpness_composition} keeps $\mathcal{L}_{\mathrm{adv}}(w^t)$ within
an explicit band above it, and the band is observable at run time because
$B_1$ costs one backward pass.  Full min-max convergence therefore reduces to
controlling $B_1$, which Phase~2 targets directly.  Second, at $k = 1$ both
the pairwise and residual terms vanish identically and the bound reads
$\mathcal{L}_{\mathrm{adv}}(w) - \hat{\mathcal{L}}(w) \leq B_1(w)$, which is
an identity, since the one-bit sharpness is by definition the largest loss
increase over $\mathcal{B}_1(w)$.  The same degeneracy explains why the
deficit $k(k-1)h$ of \cref{thm:ascent_topk} vanishes at $k = 1$.  The top-$1$
rule is then exactly optimal, and the sequential and one-shot procedures coincide.

\section{Details of Convergence Analysis}
\label{app:convergence}

We analyze convergence in two stages.  We first study a simplified
base-point descent variant with a clean Lyapunov guarantee, and then the
full SAM-style algorithm with Phase~2 evaluated at the adversarial
point.

\begin{repeatenv}{definition}{\ref{def:stationary}}[$\eta$-Discrete Stationarity]
A configuration $w \in \{-1,+1\}^n$ is $\eta$-stationary if
for every $i \in [n]$:
\begin{equation}
  \mathrm{BV}_i(w) \triangleq
  \hat{\mathcal{L}}(w^{(i)}) -
  \hat{\mathcal{L}}(w) \geq -\eta,
\end{equation}
so that no single bit-flip can reduce the empirical loss by more than
$\eta$.  A configuration is \emph{exactly stationary} when
$\eta = 0$.
\end{repeatenv}

Note that this characterizes local minima of the
empirical loss $\hat{\mathcal{L}}$, not of the adversarial loss
$\mathcal{L}_{\mathrm{adv}}$.  The overall Hamming-SAM
objective~\eqref{eq:hamming_sam_objective} targets
$\mathcal{L}_{\mathrm{adv}}$, and stationarity of $\hat{\mathcal{L}}$ is a
necessary but not sufficient condition, since a point with low
$\hat{\mathcal{L}}$ may still have high worst-case loss in
$\mathcal{B}_k$.  \cref{thm:convergence,thm:convergence_sam}
guarantee convergence of $\hat{\mathcal{L}}(w^t)$, and
\cref{rem:ladv} bounds the remaining gap to
$\mathcal{L}_{\mathrm{adv}}$ explicitly along the trajectory.

\begin{mybox}
  \begin{repeatenv}{theorem}{\ref{thm:convergence}}[Finite-Time Convergence (Base-Point Descent)]
    Suppose the empirical risk $\hat{\mathcal{L}} : \{-1,+1\}^n \to
    [0, B]$ is bounded.  Consider a variant of \cref{alg:hamming_sam}
    where Phase~2 evaluates $\mathrm{BV}_i$ at $w^t$ (rather than at the
    adversarial point $\hat{w}$), selecting
    $i^* = \arg\min_i\,\mathrm{BV}_i(w^t)$ and flipping only when
    $\mathrm{BV}_{i^*}(w^t) < -\eta$ for a threshold $\eta > 0$.  Then
    this variant converges to an $\eta$-stationary point in at most
    \begin{equation}
      T \;\leq\; \left\lfloor\frac{B}{\eta}\right\rfloor
      \label{eq:convergence_bound}
    \end{equation}
    iterations.
  \end{repeatenv}
\end{mybox}

\begin{proof}
Define the Lyapunov function
$\Phi(w^t) = \hat{\mathcal{L}}(w^t)$.

\textbf{Monotone decrease.} At each iteration, Phase~2
accepts a bit-flip at position $i^*$ only if
$\mathrm{BV}_{i^*}(w^t) < -\eta$.  Since
$\hat{\mathcal{L}}(w^{t+1}) = \hat{\mathcal{L}}((w^t)^{(i^*)}) =
\hat{\mathcal{L}}(w^t) + \mathrm{BV}_{i^*}(w^t) \leq
\hat{\mathcal{L}}(w^t) - \eta$, the Lyapunov function decreases by at
least $\eta$ at each accepted step.

\textbf{Finite termination.} Since $\hat{\mathcal{L}} \geq 0$ and
$\Phi(w^0) \leq B$, the sequence $\{\Phi(w^t)\}$ can decrease at most
$\lfloor B/\eta \rfloor$ times, after which no further bit-flip
satisfies the threshold condition, and the algorithm terminates at an
$\eta$-stationary point.
\end{proof}

\noindent The base-point variant does not exploit Phase~1, since the
descent direction is computed at $w^t$ rather than at the adversarial point.
We now show that the full SAM-style algorithm also converges, provided
the discrete Hessian is bounded.  The key observation is that the
discrete Hessian measures how much the Boolean Variation signal
changes when moving between $\hat{w}$ and $w^t$.

\begin{mybox}
  \begin{repeatenv}{theorem}{\ref{thm:convergence_sam}}[Finite-Time Convergence (SAM-Style Descent)]
    Suppose $\hat{\mathcal{L}}:\{-1,+1\}^n \to [0,B]$ and the discrete
    interaction Hessian is bounded: $|H_{ij}(w)| \leq h$ for all
    $w \in \{-1,+1\}^n$ and $i \neq j$.  In \cref{alg:hamming_sam},
    suppose Phase~2 selects $i^* \notin \mathcal{I}$ (the descent bit is
    distinct from the ascent set) and accepts the flip only when
    $\mathrm{BV}_{i^*}(\hat{w}) < -\eta$ with $\eta > k\cdot h$.  Then
    each accepted step satisfies
    \begin{equation}
      \hat{\mathcal{L}}(w^{t+1}) \;\leq\; \hat{\mathcal{L}}(w^t)
      - (\eta - kh),
    \end{equation}
    and convergence occurs in at most
    $T \leq \lfloor B/(\eta - kh) \rfloor$ iterations.
  \end{repeatenv}
\end{mybox}

\begin{proof}
Since $\hat{w} = w^t$ with the $k$ bits
$\mathcal{I} = \{j_1,\ldots,j_k\}$ flipped and $i^* \notin \mathcal{I}$,
we telescope $\mathrm{BV}_{i^*}$ from $\hat{w}$ back to $w^t$.  Define
$u^0 = \hat{w}$ and $u^l = (u^{l-1})^{(j_l)}$ (restoring bit $j_l$),
so $u^k = w^t$.  For each $l$, since $i^* \neq j_l$:
\begin{align}
  \mathrm{BV}_{i^*}(u^l) - \mathrm{BV}_{i^*}(u^{l-1})
  &= \bigl[\hat{\mathcal{L}}((u^l)^{(i^*)}) -
     \hat{\mathcal{L}}(u^l)\bigr]
   - \bigl[\hat{\mathcal{L}}((u^{l-1})^{(i^*)}) -
     \hat{\mathcal{L}}(u^{l-1})\bigr] \nonumber \\
  &= \textcolor{BrickRed}{H_{i^*,j_l}(u^{l-1})}.
  \label{eq:bv_step}
\end{align}
The second equality holds because $u^l = (u^{l-1})^{(j_l)}$, so
$(u^l)^{(i^*)} = (u^{l-1})^{(j_l,i^*)}$, and the four terms match
the definition of $H_{i^*,j_l}(u^{l-1})$.
Summing over $l = 1,\ldots,k$:
\begin{equation}
  \mathrm{BV}_{i^*}(w^t) = \textcolor{NavyBlue}{\mathrm{BV}_{i^*}(\hat{w})}
  + \textcolor{BrickRed}{\sum_{l=1}^{k} H_{i^*,j_l}(u^{l-1})}.
  \label{eq:bv_telescope}
\end{equation}
Each $|H_{i^*,j_l}(u^{l-1})| \leq h$, so
$\mathrm{BV}_{i^*}(w^t) < -\eta + kh < 0$ when $\eta > kh$.
Therefore $\hat{\mathcal{L}}(w^{t+1}) = \hat{\mathcal{L}}(w^t) +
\mathrm{BV}_{i^*}(w^t) < \hat{\mathcal{L}}(w^t) - (\eta - kh)$,
and the Lyapunov function $\Phi(w^t) = \hat{\mathcal{L}}(w^t)$
decreases by at least $\eta - kh > 0$ at each accepted step.
Since $\Phi \in [0,B]$, at most $\lfloor B/(\eta - kh)\rfloor$
decreases are possible.
\end{proof}

\begin{remark}[The BOLD accumulator as an adaptive threshold]
  \label{rem:accum_threshold}
  \cref{thm:convergence} requires a threshold $\eta > 0$ such
  that bits are flipped only when the base-point signal exceeds
  $\eta$.  The BOLD accumulator~\citep{nguyen2024bold} used in our
  practical implementation (\cref{rem:practical_descent}) plays an
  analogous role.  It maintains a per-bit state
  $a_i^t$ that integrates the adversarial signed variation across
  iterations and flips bit $i$ only when $a_i^t \cdot w_i^t \geq 1$, which
  is equivalent to requiring the current signal to exceed an effective
  threshold $\eta_{\mathrm{eff}}(i,t) > 0$ that the plasticity ratio raises
  in high-flux regions and lowers near convergence.  The thresholded quantity
  is the learning-rate-scaled accumulated backward signal, and the threshold
  adapts over training.
  \cref{app:accumulator_threshold} derives $\eta_{\mathrm{eff}}$, proves its
  positivity, and analyzes its behavior across training phases.
\end{remark}

Note that the accumulator-based variant may flip multiple bits
simultaneously per step, introducing non-linear interactions (positive
Hessian entries $H_{ij}$).  The following proposition quantifies when
simultaneous flips are safe.

\begin{mybox}
  \begin{repeatenv}{proposition}{\ref{prop:multi_bit}}[Multi-Bit Descent under Bounded Hessian]
    Under the bounded-Hessian assumption $|H_{ij}(w)| \leq h$, suppose the
    accumulator simultaneously flips a set $S = \{i_1,\ldots,i_m\}
    \subseteq [n]$ of bits in $w^t$, each satisfying
    $\mathrm{BV}_{i_l}(w^t) < -\delta$ for some $\delta > 0$.  If
    $\delta > \tfrac{(m-1)h}{2}$, then the joint flip decreases the
    empirical loss:
    \begin{equation}
      \hat{\mathcal{L}}(w^t \oplus \mathbf{1}_S) - \hat{\mathcal{L}}(w^t)
      \;\leq\; \textcolor{NavyBlue}{-m\delta}
      + \textcolor{BrickRed}{\tbinom{m}{2}h} \;<\; 0.
    \end{equation}
  \end{repeatenv}
\end{mybox}

\begin{proof}
Flip the bits sequentially: define $S_0 = \emptyset$ and
$S_l = \{i_1,\ldots,i_l\}$.  The $l$-th marginal gain is
\begin{align}
  f(S_l) - f(S_{l-1})
  &= \mathrm{BV}_{i_l}(w^t \oplus \mathbf{1}_{S_{l-1}}).
\end{align}
Using the telescoping identity (the same argument as
\cref{thm:convergence_sam}, applied to each earlier flip
$j \in S_{l-1}$):
\begin{equation}
  \mathrm{BV}_{i_l}(w^t \oplus \mathbf{1}_{S_{l-1}})
  = \mathrm{BV}_{i_l}(w^t)
  + \sum_{j \in S_{l-1}} H_{i_l,j}(\cdot)
  \leq \mathrm{BV}_{i_l}(w^t) + (l{-}1)h.
\end{equation}
Summing over $l = 1,\ldots,m$ and using
$\sum_{l=1}^{m}(l{-}1) = \binom{m}{2}$:
\[
  f(S) = \sum_{l=1}^{m} \bigl[f(S_l) - f(S_{l-1})\bigr]
  \leq \sum_{l=1}^{m}\bigl[\mathrm{BV}_{i_l}(w^t) + (l{-}1)h\bigr]
  < -m\delta + \tbinom{m}{2}h. \qedhere
\]
\end{proof}

\begin{remark}[Convergence of the accumulator variant]
  \label{rem:accum_convergence}
  Combining \cref{thm:convergence_sam} (which gives each accepted
  bit $\mathrm{BV}_{i}(w^t) < -(\eta - kh)$) with
  \cref{prop:multi_bit} (setting $\delta = \eta - kh$), the
  accumulator variant with $m$ simultaneous flips per step converges
  provided
  \begin{equation}
    \eta > \bigl(k + \tfrac{m-1}{2}\bigr)h.
    \label{eq:multi_bit_condition}
  \end{equation}
  This shows a three-way interaction between the Hamming radius $k$ (adversarial
  probe strength), the number of simultaneous flips $m$ (optimizer
  aggressiveness), and the Hessian bound $h$ (landscape curvature).  For
  typical BOLD training where $m \ll n$ and $h$ is moderate, this is a
  mild strengthening of the single-bit condition $\eta > kh$.  When the
  accumulator flips many bits ($m$ large) or the landscape has strong
  interactions ($h$ large), the condition requires either a larger threshold
  $\eta$ or a smaller Hamming radius $k$.  The condition applies to multi-bit
  steps whose flipped bits satisfy the premise of \cref{prop:multi_bit}.
\end{remark}

\section{The BOLD Accumulator as an Adaptive Threshold}
\label{app:accumulator_threshold}

This appendix provides a detailed analysis of how the BOLD
accumulator~\citep{nguyen2024bold} used in our practical implementation
(\cref{rem:practical_descent}) gives rise to an implicit,
adaptive threshold that plays the role of the fixed threshold in
\cref{thm:convergence}, applied to the accumulated backward signal
rather than to the exact Boolean Variation.  We show that the accumulator dynamics
induce a per-bit, per-step effective threshold
$\eta_{\mathrm{eff}}(i,t) > 0$ that automatically adjusts its
sensitivity over the course of training.

\subsection{Accumulator Dynamics}
\label{sec:accumulator_dynamics}

The BOLD accumulator maintains, for each bit $i \in [n]$, a scalar
state $a_i^t \in \mathbb{R}$ and a plasticity ratio
$r_i^t \in [0, 1]$.  At each Hamming-SAM iteration $t$, after Phase~1
produces the adversarial point $\hat{w}^t$
a backward pass at $\hat{w}^t$ yields the signal $g^t_i$, an estimate of
$f'_i(\hat{w}^t)$, which does not depend on the current value of bit $i$
(\cref{sec:bv_bridge}).  For
$i \notin \mathcal{I}$, $g^t_i\,w^t_i$ estimates $-\mathrm{BV}_i(\hat{w}^t)$.  For an
ascent bit, it estimates
$-\mathrm{BV}_i(w^t \oplus \mathbf{1}_{\mathcal{I}\setminus\{i\}}) \leq -\mathrm{BV}_i(w^t) + (k-1)h$.
The accumulator updates as:

\paragraph{Step 1: Accumulation.}
\begin{equation}
  a_i^{t+1} = r_i^t \cdot a_i^t
             + \eta_{\mathrm{lr}} \cdot g_i^t,
  \label{eq:accum_update}
\end{equation}
where $\eta_{\mathrm{lr}} > 0$ is the learning rate.

\paragraph{Step 2: Flip decision.}
Define the flip indicator set:
\begin{equation}
  \mathcal{F}^{t+1} = \bigl\{i \in [n] :
    a_i^{t+1} \cdot w_i^t \geq 1\bigr\}.
  \label{eq:flip_condition}
\end{equation}
For each $i \in \mathcal{F}^{t+1}$, the weight is flipped:
$w_i^{t+1} = -w_i^t$.  For $i \notin \mathcal{F}^{t+1}$,
$w_i^{t+1} = w_i^t$.

\paragraph{Step 3: Reset and plasticity.}
After flipping:
\begin{align}
  a_i^{t+1} &\leftarrow 0 \quad \text{for all } i \in \mathcal{F}^{t+1},
  \label{eq:accum_reset} \\
  r_i^{t+1} &= 1 - \bar{f}^{t+1},
  \quad \text{where } \bar{f}^{t+1}
    = \frac{|\mathcal{F}^{t+1} \cap \mathcal{G}_i|}{|\mathcal{G}_i|}
  \label{eq:plasticity_update}
\end{align}
is the fraction of bits in the same parameter tensor $\mathcal{G}_i$
that flipped at step $t+1$.

\subsection{Derivation of the Effective Threshold}

We now derive the effective threshold $\eta_{\mathrm{eff}}(i,t)$ by
analyzing the flip condition~\eqref{eq:flip_condition}.

\begin{mybox}
  \begin{proposition}[Effective Threshold]
    \label{prop:effective_threshold}
    At iteration $t$, bit $i$ is flipped if and only if the signal
$g^t_i \cdot w^t_i$ satisfies
    \begin{equation}
      g_i^t \cdot w_i^t \;\geq\;
      \underbrace{\frac{1 - r_i^t \cdot a_i^t \cdot w_i^t}
      {\eta_{\mathrm{lr}}}}_{\displaystyle
      \triangleq\; \eta_{\mathrm{eff}}(i, t)}.
      \label{eq:eta_eff_derivation}
    \end{equation}
    Moreover, $\eta_{\mathrm{eff}}(i, t) > 0$ for all $i$ and $t$.
  \end{proposition}
\end{mybox}

\begin{proof}
Substituting~\eqref{eq:accum_update}
into~\eqref{eq:flip_condition}:
\begin{align}
  a_i^{t+1} \cdot w_i^t &\geq 1 \nonumber \\
  \bigl(r_i^t \cdot a_i^t + \eta_{\mathrm{lr}} \cdot g_i^t\bigr)
    \cdot w_i^t &\geq 1 \nonumber \\
  r_i^t \cdot a_i^t \cdot w_i^t
    + \eta_{\mathrm{lr}} \cdot g_i^t \cdot w_i^t &\geq 1 \nonumber \\
  g_i^t \cdot w_i^t &\geq
    \frac{1 - r_i^t \cdot a_i^t \cdot w_i^t}{\eta_{\mathrm{lr}}}.
\end{align}

For strict positivity of $\eta_{\mathrm{eff}}$, we need
$r_i^t \cdot a_i^t \cdot w_i^t < 1$.  This holds by a simple
induction argument:

\emph{Base case.}  At initialization, $a_i^0 = 0$ for all $i$, so
$r_i^0 \cdot a_i^0 \cdot w_i^0 = 0 < 1$.

\emph{Inductive step.}  Suppose
$r_i^t \cdot a_i^t \cdot w_i^t < 1$ at the start of iteration $t$.
Two cases arise:
\begin{itemize}
  \item If $i \in \mathcal{F}^{t+1}$ (bit $i$ flips): the
    accumulator resets to $a_i^{t+1} = 0$
    by~\eqref{eq:accum_reset}, so
    $r_i^{t+1} \cdot a_i^{t+1} \cdot w_i^{t+1} = 0 < 1$.
  \item If $i \notin \mathcal{F}^{t+1}$ (bit $i$ does not flip): by~\eqref{eq:flip_condition},
we have $a_i^{t+1} \cdot w_i^t < 1$.  Since $w_i^{t+1} = w_i^t$ and
$r_i^{t+1} \in [0, 1]$, two subcases arise.
If $a_i^{t+1} \cdot w_i^t \geq 0$, then
$r_i^{t+1} \cdot a_i^{t+1} \cdot w_i^{t+1}
\leq a_i^{t+1} \cdot w_i^t < 1$.
If $a_i^{t+1} \cdot w_i^t < 0$, then
$r_i^{t+1} \cdot a_i^{t+1} \cdot w_i^{t+1}
\leq 0 < 1$.
\end{itemize}
In both cases the invariant is maintained, so
$\eta_{\mathrm{eff}}(i,t) > 0$ for all $i, t$.
\end{proof}

\subsection{Adaptive Behavior Across Training Phases}

The plasticity ratio $r_i^t$ adjusts $\eta_{\mathrm{eff}}$ according to the
training phase.  We analyze three regimes.

\paragraph{Early training (high flip rate).}
When many bits flip per step, $\bar{f}_{\mathcal{G}_i}^t$ is large and
$r_i^t = 1 - \bar{f}_{\mathcal{G}_i}^t$ is small.  Consequently, the carry-over
term $r_i^t \cdot a_i^t \cdot w_i^t$ is heavily damped, and:
\begin{equation}
  \eta_{\mathrm{eff}}(i, t)
  \approx \frac{1 - 0}{\eta_{\mathrm{lr}}}
  = \frac{1}{\eta_{\mathrm{lr}}}.
  \label{eq:eta_early}
\end{equation}
The threshold is at its \emph{maximum}.  The optimizer requires a strong
single-step signal before flipping any bit.  This prevents
oscillation during the initial phase when the loss landscape is noisy
and the sharpness $B_k$ is large.

\paragraph{Mid training (moderate flip rate).}
As training progresses, fewer bits flip and $r_i^t$ increases toward
$1$.  The accumulator begins to retain history across steps:
$a_i^t$ builds up over $\sim 1/(1 - r_i^t)$ iterations.  If the
signal consistently favors flipping bit $i$, that is,
$g_i^t \cdot w_i^t > 0$ across multiple steps, the accumulated
product $r_i^t \cdot a_i^t \cdot w_i^t$ grows, and
$\eta_{\mathrm{eff}}(i,t)$ \emph{decreases}.  The optimizer becomes
progressively more willing to act on weaker but persistent sharpness
signals.

\paragraph{Near convergence (low flip rate).}
When $\bar{f}_{\mathcal{G}_i}^t \approx 0$, we have $r_i^t \approx 1$ and the
accumulator is nearly a running sum.  After $T$ consistent steps with
average signal $\bar{g}_i$:
\begin{equation}
  a_i^t \cdot w_i^t \approx T \cdot \eta_{\mathrm{lr}}
    \cdot \bar{g}_i \cdot w_i^t,
\end{equation}
so $\eta_{\mathrm{eff}}(i,t)
\approx (1 - T\,\eta_{\mathrm{lr}}\,\bar{g}_i \cdot w_i^t)
/ \eta_{\mathrm{lr}} \to 0^+$ as $T$ grows.  The optimizer can
therefore detect very small sharpness improvements, which a
fixed $\eta$ would not allow.

\subsection{Formal Convergence Guarantee}

\begin{mybox}
  \begin{theorem}[Properties of Accumulator-Based Hamming-SAM]
    \label{thm:accum_convergence}
    Let $\hat{\mathcal{L}} : \{-1,+1\}^n \to [0, B]$ be bounded, and
    suppose \cref{alg:hamming_sam} with Phase~2 replaced by the
    accumulator update~\eqref{eq:accum_update}--\eqref{eq:plasticity_update}.
    Assume the bounded-Hessian condition $|H_{ij}(w)| \leq h$ for all
    $w, i \neq j$.  Part~(i) holds unconditionally, and parts~(ii) and~(iii)
    hold when every flipped bit satisfies $\mathrm{BV}_i(w^t) < -\delta$.
    Then:
    \begin{enumerate}
      \item[\emph{(i)}] Every bit that flips satisfies
      $g_{i^*}^t \cdot w_{i^*}^t \geq \eta_{\mathrm{eff}}(i^*, t) > 0$
      (\cref{prop:effective_threshold}),
      so each accepted flip is justified by strictly positive accumulated
      evidence.
      \item[\emph{(ii)}] If at step $t$ the accumulator flips a set
      $\mathcal{F}^{t+1}$ of $m$ bits, each with
      $\mathrm{BV}_i(w^t) < -\delta$ for $\delta > 0$, and
      $\delta > \tfrac{(m-1)h}{2}$, then
      $\hat{\mathcal{L}}(w^{t+1}) < \hat{\mathcal{L}}(w^t)$
      (\cref{prop:multi_bit}).
      \item[\emph{(iii)}] The total number of flips across all bits and
      all iterations is finite.  Specifically, if the conditions of (ii) hold
      at every flip step with per-flip decrease at least
      $\delta_{\min} > 0$, then the total
      number of flip events is at most
      $\lfloor B / \delta_{\min} \rfloor$.
    \end{enumerate}
  \end{theorem}
\end{mybox}

  \begin{proof}\mbox{}

\emph{Part~(i).}  By \cref{prop:effective_threshold},
$\eta_{\mathrm{eff}}(i,t) > 0$ for all $i, t$ (the invariant
$r_i^t a_i^t w_i^t < 1$ is maintained by the reset mechanism).
Therefore every flip is justified by accumulated backward signal
exceeding a positive threshold.

\emph{Part~(ii).}  This follows directly from
\cref{prop:multi_bit}.  The $m$ bits flipped simultaneously
each have $\mathrm{BV}_i(w^t) < -\delta$, and the bounded-Hessian
assumption controls the interaction cost at
$\binom{m}{2}h$.  When $\delta > (m{-}1)h/2$, the joint loss change
$\hat{\mathcal{L}}(w^{t+1}) - \hat{\mathcal{L}}(w^t) \leq
-m\delta + \binom{m}{2}h < 0$.

When the SAM-style evaluation is used (Boolean Variation computed at $\hat{w}$
rather than $w^t$), \cref{thm:convergence_sam} gives
$\mathrm{BV}_i(w^t) < -(\eta - kh)$ for each accepted bit, so
$\delta = \eta - kh$ and the multi-bit condition becomes
$\eta > (k + \tfrac{m-1}{2})h$.

\emph{Part~(iii).}  Each flip step that satisfies the condition in (ii)
decreases $\hat{\mathcal{L}}(w^t)$ by at least $\delta_{\min} > 0$.
Since $\hat{\mathcal{L}} \in [0, B]$, the total number of such
steps is at most $\lfloor B / \delta_{\min} \rfloor$.  The accumulator
reset ensures no bit flips indefinitely: after each flip, the threshold
returns to $1/\eta_{\mathrm{lr}}$, requiring fresh evidence for any
subsequent flip of the same bit.
\end{proof}

\begin{remark}[Comparison with fixed-$\eta$ convergence]
  \cref{thm:convergence} guarantees convergence of
  $\hat{\mathcal{L}}(w^t)$ in $\lfloor B/\eta \rfloor$ steps with a
  fixed threshold $\eta$.  \cref{thm:convergence_sam} extends
  this to the SAM-style algorithm under $\eta > kh$.
  \cref{thm:accum_convergence} further extends to the multi-bit
  accumulator under $\eta > (k + \tfrac{m-1}{2})h$.  The three results
  form a hierarchy, from base-point descent (simplest, strongest bound)
  $\to$ SAM-style single-bit (uses adversarial information, requires
  bounded Hessian) $\to$ accumulator multi-bit (closest to the practical
  algorithm, under the premise of part~(ii)).
  The decrease guarantee of part~(ii) applies at a flip step whose
  instantaneous signal exceeds $(k + \tfrac{m-1}{2})h$.  Because
  $\eta_{\mathrm{eff}}(i,t)$ can fall below this level near convergence,
  flips triggered by accumulated evidence alone are not covered by it.
  Ascent bits do not satisfy the premise $\mathrm{BV}_i(w^t) < -\delta$, and
  they receive a non-flip signal whenever $\mathrm{BV}_i(w^t) > (k-1)h$
  (\cref{sec:accumulator_dynamics}).
\end{remark}

\section{Detailed Proof of Generalization Bound with Data-Dependent Priors}
\label{app:pac_bayes}

We derive the PAC-Bayes generalization bound of
\cref{sec:pac_bayes} using data-dependent
priors~\citep{parrado2012pac, dziugaite2021role}.  Split the training
data as $\mathcal{D} = \mathcal{S}_0 \sqcup \mathcal{S}_1$.  Train a
1,9xreference configuration $w_0$ on $\mathcal{S}_0$ and define the prior as
a product distribution that independently flips each bit of $w_0$ with
probability $p \in (0, \tfrac{1}{2})$:
1\begin{equation}
  P(w) = \prod_{i=1}^n P_i(w_i), \quad
  P_i(w_i) = \begin{cases}
    1 - p & \text{if } w_i = (w_0)_i, \\
    p     & \text{if } w_i \neq (w_0)_i.
  \end{cases}
  \label{eq:prior}
1\end{equation}
The posterior $Q$ is the degenerate (deterministic) distribution
concentrating on the Hamming-SAM solution $\hat{w}$, which is a valid
choice in the PAC-Bayes framework~\citep{catoni2007pac}.  Because $Q$ is a
point mass, $\mathbb{E}_{Q}[\mathcal{L}] = \mathcal{L}(\hat{w})$, so the
bound below controls the loss of the returned configuration itself.

\subsection{KL Divergence Computation}

\begin{mybox}
  \begin{repeatenv}{lemma}{\ref{lem:kl}}[KL for Bernoulli Bit-Flip Prior]
    Let $d = d_H(\hat{w}, w_0)$ denote the Hamming distance between
    $\hat{w}$ and the prior center $w_0$.  The KL divergence between the
    deterministic posterior $Q = \delta_{\hat{w}}$ and the prior $P$
    in~\eqref{eq:prior} is:
    \begin{equation}
      \mathrm{KL}(Q \| P)
      = d \cdot \log\frac{1}{p}
      + (n - d) \cdot \log\frac{1}{1-p}.
      \label{eq:kl_correct}
    \end{equation}
  \end{repeatenv}
\end{mybox}

\begin{proof}
Since $Q = \delta_{\hat{w}}$ is a point mass:
\begin{align}
  \mathrm{KL}(Q \| P)
  &= \log\frac{1}{P(\hat{w})}
  = -\sum_{i=1}^n \log P_i(\hat{w}_i).
\end{align}
Partition the coordinates into
$\mathcal{A} = \{i : \hat{w}_i = (w_0)_i\}$ (agreement,
$|\mathcal{A}| = n-d$) and
$\mathcal{F} = \{i : \hat{w}_i \neq (w_0)_i\}$ (flipped,
$|\mathcal{F}| = d$).  Then:
\begin{align}
  \mathrm{KL}(Q \| P)
  &= -\sum_{i \in \mathcal{A}} \log(1-p)
     -\sum_{i \in \mathcal{F}} \log p
  = (n - d)\log\frac{1}{1-p} + d\log\frac{1}{p}.
\end{align}
\end{proof}

\paragraph{Optimization over $p$.}
Taking the derivative and setting it to zero:
\[
  \frac{\partial}{\partial p}\mathrm{KL}(Q\|P)
  = -\frac{d}{p} + \frac{n-d}{1-p} = 0
  \quad\Longrightarrow\quad
  p^* = \frac{d}{n}.
\]
At this optimum, $\mathrm{KL}(Q\|P) = n \cdot h(d/n)$, where $h$ is
the binary entropy.  For $d \ll n$, this scales as
$d\log(n/d) + O(d)$, which is much smaller than $n \log 2$.

\subsection{Generalization Bound}

\begin{mybox}
  \begin{theorem}[PAC-Bayes Bound with Data-Dependent Bernoulli Prior]
    \label{thm:pac_bayes_full}
    Let the loss take values in $[0,1]$, such as the 0-1 classification error.
    Let $w_0$ be a configuration trained on $\mathcal{S}_0$, and $P$ be the
    bit-flip prior~\eqref{eq:prior} centered at $w_0$ with flip probability
    $p \in (0, \tfrac{1}{2})$.  Let $\hat{w}$ be the Hamming-SAM solution
    trained on $\mathcal{S}_1$, with $d = d_H(\hat{w}, w_0)$.  Then for any
    $\delta \in (0,1)$, with probability at least $1 - \delta$ over the
    draw of $\mathcal{S}_1$:
    \begin{equation}
      \mathcal{L}(\hat{w}) \;\leq\;
      \textcolor{BrickRed}{\hat{\mathcal{L}}_{\mathcal{S}_1}(\hat{w})}
      + \textcolor{NavyBlue}{\sqrt{\frac{d\,\log\frac{1}{p}
      + (n-d)\,\log\frac{1}{1-p}
      + \log\frac{2\sqrt{m}}{\delta}}{2m}}},
      \label{eq:pac_bayes_full}
    \end{equation}
    where $m = |\mathcal{S}_1|$.
  \end{theorem}
\end{mybox}

\begin{proof}
We apply the PAC-Bayes--KL inequality of \cite{catoni2007pac} in its
McAllester form~\citep{mcallester2003pac}.  For any prior $P$
(independent of $\mathcal{S}_1$) and posterior $Q$,
\[
  \mathrm{KL}_{\mathrm{Ber}}\!\bigl(
    \mathbb{E}_Q[\hat{\mathcal{L}}]
    \;\big\|\;
    \mathbb{E}_Q[\mathcal{L}]
  \bigr)
  \leq \frac{\mathrm{KL}(Q\|P) + \log(2\sqrt{m}/\delta)}{m}.
\]
Applying Pinsker's inequality to the left-hand side and substituting
the KL from \cref{lem:kl} yields~\eqref{eq:pac_bayes_full}, using
$\mathbb{E}_Q[\mathcal{L}] = \mathcal{L}(\hat{w})$ for the point-mass
posterior.  The
prior $P$ is constructed from $\mathcal{S}_0$ and is therefore
independent of $\mathcal{S}_1$, satisfying the measurability condition.
\end{proof}

\begin{remark}[Optimizing the flip probability $p$]
\label{rem:optimize_p}
The bound~\eqref{eq:pac_bayes_full} holds for any fixed
$p \in (0, \tfrac{1}{2})$ chosen independently of
$\mathcal{S}_1$.  To optimize $p$ post hoc (after observing
$d = d_H(\hat{w}, w_0)$, which depends on $\mathcal{S}_1$),
we apply a union bound over a dyadic grid
$\mathcal{P} = \{2^{-1}, 2^{-2}, \ldots, 2^{-\lceil\log_2 n\rceil}\}$
of $|\mathcal{P}| = \lceil\log_2 n\rceil$ candidate values.
Replacing $\delta$ by $\delta / |\mathcal{P}|$ in the bound and
selecting $p^* = \arg\min_{p \in \mathcal{P}}
\mathrm{KL}(Q\|P)$ yields:
\begin{equation}
  \mathcal{L}(\hat{w}) \;\leq\;
  \textcolor{BrickRed}{\hat{\mathcal{L}}_{\mathcal{S}_1}(\hat{w})} \;+\;
  \textcolor{NavyBlue}{\sqrt{\frac{\mathrm{KL}^* + \log(2\sqrt{m}\,
    \lceil\log_2 n\rceil\,/\,\delta)}{2m}}},
  \label{eq:pac_bayes_optimized}
\end{equation}
where $\mathrm{KL}^* = \min_{p \in \mathcal{P}}
[d\log(1/p) + (n{-}d)\log(1/(1{-}p))]$.  The additional
penalty $\log\lceil\log_2 n\rceil$ is negligible.
For $n = 442{,}368$ it adds $\log 19 \approx 2.9$ to the numerator, which is
below $0.1\%$ of the KL term for every bound with $d > 0$ beyond the first
epoch, and changes no reported bound by more than $0.001$ at epoch 1.
In practice, the grid optimum closely approximates the
continuous optimum $p^* = d/n$.
This is standard practice in PAC-Bayes
evaluation~\citep{dziugaite2017nonvacuous, dziugaite2021role}.
\end{remark}
\subsection{Empirical Evaluation}

We evaluate the bound using the data-split protocol described in
\cref{sec:pac_bayes}.  CIFAR-10 is split into $\mathcal{S}_0$ (first
25K training samples) and $\mathcal{S}_1$ (last 25K).  Each method
trains a prior $w_0$ on $\mathcal{S}_0$, then fine-tunes on
$\mathcal{S}_1$ starting from $w_0$.  Each method uses its
own prior.  Hamming-SAM fine-tunes from a Hamming-SAM prior, and standard BOLD
fine-tunes from a BOLD prior.  The Hamming distance $d = d_H(\hat{w},
w_0)$ and the bound are computed
at each epoch of fine-tuning, with the 0-1 error on $\mathcal{S}_1$ as the
empirical risk.  Following
\cref{rem:optimize_p}, we optimize $p$ over a dyadic
grid and apply the union-bound-corrected
bound~\eqref{eq:pac_bayes_optimized}.  The correction adds
$\log\lceil\log_2 n\rceil \approx 2.9$ to the numerator,
which is negligible relative to the KL term for every bound beyond the
first epoch and does not affect non-vacuousness of
any reported bound.

\paragraph{Results.}  \Cref{tab:pac_bayes_results} reports the
PAC-Bayes bound for the 3-layer MLP ($n = 442{,}368$, $m = 25{,}000$,
$\delta = 0.05$).  Two findings stand out.

First, {Hamming-SAM achieves non-vacuous bounds at every epoch},
while standard BOLD is vacuous after epoch~1.  At epoch~50, Hamming-SAM
yields a bound of $0.800$ with only $d = 644$ bits changed ($d/n =
0.15\%$), whereas BOLD requires $d = 4{,}233$ bits ($d/n = 0.96\%$) and
produces a vacuous bound of $1.197$.  Hamming-SAM flips $6.6\times$ fewer
bits during adaptation, which shows that flat basins are more stable
under the distribution shift from $\mathcal{S}_0$ to $\mathcal{S}_1$.

Second, {the Hamming-SAM prior generalizes well without
fine-tuning}.  At epoch~1 ($d = 0$, before any adaptation), the
Hamming-SAM prior already achieves $52.1\%$ accuracy on $\mathcal{S}_1$
versus $48.0\%$ for the BOLD prior, an advantage of $4.1$ points from the
flat basin alone.  The tightest bound ($0.495$) occurs at this point,
reflecting that the Hamming-SAM solution found on $\mathcal{S}_0$ transfers
directly to $\mathcal{S}_1$ with no modification.

\begin{table}[t]
  \caption{PAC-Bayes bounds for 3-layer MLP on CIFAR-10 ($n = 442{,}368$,
    $m = 25{,}000$, $\delta = 0.05$).  Each method uses its own prior
    trained on $\mathcal{S}_0$, and fine-tuning is on $\mathcal{S}_1$.
    KL is the minimum over the dyadic grid of \cref{rem:optimize_p}.
    Non-vacuous bounds (${}<1.0$) are \textbf{bolded}.}
  \label{tab:pac_bayes_results}
  \centering\small
  \begin{tabular}{llrrrrr}
    \toprule
    \textbf{Method} & \textbf{Epoch} & \textbf{Acc on $\mathcal{S}_1$} & $d$ & $d/n$ (\%) & \textbf{KL} & \textbf{Bound} \\
    \midrule
    \multirow{6}{*}{BOLD}
      & 1  & 48.0\% &     2 & 0.00 &      27 & \textbf{0.548} \\
      & 10 & 47.6\% & 2,161 & 0.49 & 13,706  & 1.048 \\
      & 20 & 48.9\% & 3,213 & 0.73 & 19,034  & 1.128 \\
      & 30 & 49.4\% & 3,902 & 0.88 & 22,372  & 1.175 \\
      & 40 & 49.4\% & 4,186 & 0.95 & 23,747  & 1.195 \\
      & 50 & 49.6\% & 4,233 & 0.96 & 23,975  & 1.197 \\
    \midrule
    \multirow{6}{*}{Hamming-SAM ($k{=}500$)}
      & 1  & 52.1\% &     0 & 0.00 &       1 & \textbf{0.495} \\
      & 10 & 50.9\% &   385 & 0.09 &  3,101  & \textbf{0.740} \\
      & 20 & 51.2\% &   490 & 0.11 &  3,828  & \textbf{0.765} \\
      & 30 & 51.4\% &   581 & 0.13 &  4,459  & \textbf{0.785} \\
      & 40 & 51.2\% &   633 & 0.14 &  4,813  & \textbf{0.799} \\
      & 50 & 51.3\% &   644 & 0.15 &  4,881  & \textbf{0.800} \\
    \bottomrule
  \end{tabular}
\end{table}

\paragraph{Scaling behavior.}
For small $p$, the dominant KL term is $d \cdot \log(1/p)$,
since $\log(1/(1{-}p)) \approx p$ is negligible.  Near the optimum
$p \approx d/n$, the bound scales as $O(d\log(n/d))$, so the complexity
depends on the \emph{adaptation distance} $d$ rather than the total
number of parameters $n$.  For example, for the Hamming-SAM
result at epoch~50, the grid optimum $p = 2^{-9}$ gives
$\mathrm{KL} \approx 644 \cdot \log 512 +
441{,}724 \cdot \log(512/511) \approx 4{,}881$, and a generalization
gap of $\sqrt{(4{,}881 + 11.7)/50{,}000} \approx 0.31$.  The uniform prior
would give $\mathrm{KL} = 442{,}368 \cdot \log 2 \approx 306{,}600$,
yielding a vacuous bound exceeding $2.4$.

The results confirm the mechanism described in \cref{sec:pac_bayes}:
lower sharpness $B_k$ yields smaller adaptation distance $d$, which in
turn yields a tighter bound.  The Hamming-SAM prior, having converged to a
flat basin on $\mathcal{S}_0$, is stable under the distribution shift to
$\mathcal{S}_1$, and the optimizer needs only 644 bit-flips ($0.15\%$ of
parameters) to adapt.  The BOLD prior lies in a sharper basin and
requires $4{,}233$ bit-flips ($0.96\%$) to reach a reasonable solution on
$\mathcal{S}_1$, and even then the bound remains vacuous.

\section{Forman-Ricci Curvature: Full Analysis}
\label{app:ricci}

The preceding sections established that Hamming-SAM finds weight
configurations with uniformly low loss across a Hamming ball.
In this appendix, we ask a structural question: \emph{what geometric property of the
loss landscape distinguishes a truly flat basin from a narrow valley
that only has low worst-case loss?}  In continuous optimization, this
question is answered by Ricci curvature, which measures whether
neighboring geodesics converge (positive curvature, robust basins) or
diverge (negative curvature, brittle saddle regions).  We develop the
discrete analogue for the Boolean hypercube using Forman-Ricci
curvature~\citep{forman2003bochner}, providing a combinatorial
characterization of flatness that complements the algebraic
(PAC-Bayes) and combinatorial (compression) perspectives of
\cref{sec:method}.

\subsection{The Loss-Weighted Hamming Graph}

The key construction is to embed the loss landscape into the geometry of
the Hamming graph by assigning weights that reflect the local loss
structure.

\begin{definition}[Loss-Weighted Hamming Graph]\label{def:weighted_hamming}
Given a Boolean weight space $\{-1,+1\}^n$ and empirical risk
$\hat{\mathcal{L}}$, define the loss-weighted Hamming graph
$G_\beta = (V, E, \omega)$ with:
\begin{itemize}
  \item \emph{Node weights:}
    $\omega(w) = \exp(-\beta\,\hat{\mathcal{L}}(w))$ for a temperature
    $\beta > 0$, concentrating mass on low-loss configurations.
  \item \emph{Edge weights:}
    $\omega(w, w^{(i)}) = \exp(-\beta\,|\mathrm{BV}_i(w)|)$, weighting
    edges by the stability of the corresponding bit-flip.  Stable
    flips with $|\mathrm{BV}_i| \approx 0$ receive weight close to
    $1$, while disruptive flips receive low weight.
\end{itemize}
The parameter $\beta$ controls the sensitivity of the geometry to loss
differences.  In the limit $\beta \to 0$, $G_\beta$ reduces to the
unweighted Hamming graph.  As $\beta \to \infty$, the geometry
concentrates entirely on global minima.
\end{definition}

\subsection{Forman-Ricci Curvature on Weighted Graphs}

Forman~\citep{forman2003bochner} introduced a combinatorial analogue of
Ricci curvature for CW-complexes that depends only on local
edge--node--edge adjacency.  Specialized to our weighted graph:

\begin{definition}[Forman-Ricci Curvature~\citep{forman2003bochner}]
\label{def:ricci}
For an edge $e = (w, w^{(j)})$ in the loss-weighted Hamming graph
$G_\beta$:
\begin{align}
  \mathrm{Ric}_F(e)
  &= \omega(e)\Biggl[
      \frac{\omega(w) + \omega(w^{(j)})}{\omega(e)} \nonumber \\
  &\qquad
    - \sum_{\substack{e' \sim w \\ e' \neq e}}
        \frac{\omega(w)}{\sqrt{\omega(e)\,\omega(e')}}
    - \sum_{\substack{e' \sim w^{(j)} \\ e' \neq e}}
        \frac{\omega(w^{(j)})}{\sqrt{\omega(e)\,\omega(e')}}
  \Biggr],
  \label{eq:forman_ricci}
\end{align}
where the sums range over edges incident to $w$ and $w^{(j)}$,
respectively, and each vertex has exactly $n-1$ other incident edges.
\end{definition}

The following lemma simplifies the curvature formula for the specific
structure of the loss-weighted Hamming graph, making the dependence on
the loss landscape explicit.

\begin{mybox}
  \begin{lemma}[Simplified Forman-Ricci on the Loss-Weighted Hamming Graph]
    \label{lem:ricci_simplified}
    Let $e_j = (w, w^{(j)})$ be an edge with weight
    $\omega_j \triangleq \omega(e_j) = \exp(-\beta|\mathrm{BV}_j(w)|)$.
    Then:
    \begin{equation}
      \mathrm{Ric}_F(e_j)
      = \textcolor{NavyBlue}{\omega(w) + \omega(w^{(j)})}
      - \textcolor{BrickRed}{\sum_{\substack{i=1 \\ i \neq j}}^{n}
      \left[
      \omega(w)\sqrt{\frac{\omega_j}{\omega_i}}
    + \omega(w^{(j)})\sqrt{\frac{\omega_j}{\omega'_i}}
    \right]},
    \label{eq:ricci_simplified}
\end{equation}
where $\omega_i = \exp(-\beta|\mathrm{BV}_i(w)|)$ and
$\omega'_i = \exp(-\beta|\mathrm{BV}_i(w^{(j)})|)$ are the edge
weights at $w$ and $w^{(j)}$, respectively.
\end{lemma}
\end{mybox}

\begin{proof}
The $n$-dimensional hypercube is $n$-regular, so each vertex has exactly
$n$ incident edges.  Expanding~\eqref{eq:forman_ricci}, the first term
is
$\omega(e)\cdot[\omega(w)+\omega(w^{(j)})]/\omega(e)
= \omega(w)+\omega(w^{(j)})$.  Each of the two sums ranges over $n-1$
edges (all neighbors except the edge $e_j$ itself).  Distributing the
$\omega(e)$ into the sums, the terms become
$\omega(w) \sqrt{\omega(e)/\omega(e')}$.  Substituting the
specific edge weights
$\omega(e') = \exp(-\beta|\mathrm{BV}_i(w)|) = \omega_i$ for edges
$e' \sim w$ and $\omega(e') = \omega'_i$ for edges $e' \sim w^{(j)}$
yields~\eqref{eq:ricci_simplified}.
\end{proof}

The structure of~\eqref{eq:ricci_simplified} shows a competition.
The \textcolor{NavyBlue}{positive term} rewards nodes with
low loss (high $\omega$), while the \textcolor{BrickRed}{negative sum}
penalizes disparity
between the edge $e_j$ and its neighboring edges.  Curvature is high
when both endpoint losses and all neighboring Boolean Variations are
approximately uniform, which is the signature of a flat basin.

\subsection{Curvature and Sharpness}

The following bounds relate the Forman-Ricci curvature to the sharpness $B_k$.
They provide a computable diagnostic of local flatness.

\begin{mybox}
  \begin{proposition}[Curvature Bounds under Low Sharpness]
    \label{prop:curvature}
    Let $w^*$ be a local minimum of the Hamming-SAM objective
    $\mathcal{L}_{\mathrm{adv}}(w) = \max_{w' \in \mathcal{B}_k(w)}
    \hat{\mathcal{L}}(w')$, and define the sharpness at $w^*$ as
    $B_k \triangleq \mathcal{L}_{\mathrm{adv}}(w^*) -
\hat{\mathcal{L}}(w^*)$.  Then for any edge
$e_j = (w^*, w^{*,(j)})$ with $j \in [n]$:
\begin{enumerate}
  \item[\emph{(i)}] \textbf{(Node weight concentration)}
  The node weight ratio satisfies
  $\omega(w^{*,(j)})/\omega(w^*) \geq \exp(-\beta B_k)$.

  \item[\emph{(ii)}] \textbf{(Flat-basin limit)}
    If in addition $w^*$ is a local minimum of $\hat{\mathcal{L}}$ and
    $|H_{ij}(w^*)| \leq h$ for all $i \neq j$, then as $B_k, h \to 0$ all
    edge curvatures $\mathrm{Ric}_F(e_j)$ converge to
    $\mathrm{Ric}_F^{\mathrm{unif}} = 2\omega^* - 2(n-1)\omega^*
    = 2\omega^*(1 - (n-1))$,
    the curvature of the unweighted hypercube rescaled by the
    node weight $\omega^* = \exp(-\beta\hat{\mathcal{L}}(w^*))$.
  \end{enumerate}
  \end{proposition}
\end{mybox}

\begin{proof}\mbox{}

\emph{Part~(i).}  Since $w^{*,(j)} \in \mathcal{B}_1(w^*) \subseteq
\mathcal{B}_k(w^*)$ for $k \geq 1$, the definition of the sharpness gives
$\hat{\mathcal{L}}(w^{*,(j)}) \leq
\mathcal{L}_{\mathrm{adv}}(w^*) = \hat{\mathcal{L}}(w^*) + B_k$.
Therefore:
\begin{align}
  \frac{\omega(w^{*,(j)})}{\omega(w^*)}
  &= \exp\bigl(-\beta[\hat{\mathcal{L}}(w^{*,(j)})
    - \hat{\mathcal{L}}(w^*)]\bigr) \nonumber \\
  &\geq \exp(-\beta B_k). \label{eq:weight_ratio}
\end{align}

\emph{Part~(ii).}  At a local minimum,
$0 \leq \mathrm{BV}_i(w^*) \leq B_1(w^*) \leq B_k$, and
$\mathrm{BV}_i(w^{*,(j)}) = \mathrm{BV}_i(w^*) + H_{ij}(w^*)$, so
$|\mathrm{BV}_i(w^{*,(j)})| \leq B_k + h$.  As $B_k, h \to 0$, all edge
weights $\omega_i, \omega'_i$ tend to $1$ and, by part~(i),
$\omega(w^{*,(j)}) \to \omega(w^*) = \omega^*$.  Substituting into
\cref{lem:ricci_simplified}:
\[
  \mathrm{Ric}_F(e_j)
  \;\to\; \omega^* + \omega^*
  - \sum_{i \neq j} \bigl[\omega^* + \omega^*\bigr]
  = 2\omega^* - 2(n-1)\omega^*.
\]
For the unweighted case ($\beta = 0$, so $\omega^* = 1$), this
reduces to $2 - 2(n-1) < 0$ for $n > 2$, recovering the known
negative curvature of the hypercube graph.
\end{proof}

\subsection{Average Curvature and Sharpness}

The individual-edge result can be aggregated into a statement
linking the average curvature to the Hamming-SAM objective.

\begin{mybox}
  \begin{theorem}[Average Curvature Lower Bound]
    \label{thm:avg_curvature}
    Let $w^*$ be a Hamming-SAM minimum with sharpness
$B_k$ and loss $\ell^* = \hat{\mathcal{L}}(w^*)$.
Assume further that $w^*$ is a local minimum of $\hat{\mathcal{L}}$, i.e.,
$\mathrm{BV}_i(w^*) \geq 0$ for all $i \in [n]$, and that
$|H_{ij}(w^*)| \leq h$ for all $i \neq j$.
Define the average Forman-Ricci curvature over all edges incident to $w^*$:
    \[
      \overline{\mathrm{Ric}}_F(w^*)
      \triangleq \frac{1}{n} \sum_{j=1}^{n}
      \mathrm{Ric}_F(w^*, w^{*,(j)}).
      \]
      Then:
      \begin{equation}
        \overline{\mathrm{Ric}}_F(w^*)
        \;\geq\;
        e^{-\beta \ell^*}\Bigl[
        \textcolor{NavyBlue}{1 + e^{-\beta B_k}}
        - \textcolor{BrickRed}{(n-1)\,e^{\beta B_k/2}\bigl(1 + e^{\beta h/2}\bigr)}\Bigr].
        \label{eq:avg_curvature_bound}
      \end{equation}
      In particular, the right-hand side is decreasing in $B_k$,
      so minimizing the Hamming-SAM sharpness $B_k$
      maximizes the curvature lower bound.
    \end{theorem}
\end{mybox}

\begin{proof}
For each edge $e_j = (w^*, w^{*,(j)})$, the positive term
in~\eqref{eq:ricci_simplified} satisfies
$\omega(w^*) + \omega(w^{*,(j)}) \geq
e^{-\beta\ell^*}(1 + e^{-\beta B_k})$ by part~(i) of
\cref{prop:curvature}.

For the negative terms, we seek an upper bound.  The edge weights satisfy
$\omega_j = e^{-\beta|\mathrm{BV}_j|} \leq 1$.  For the edges incident to
$w^*$, the local minimum assumption gives
$0 \leq \mathrm{BV}_i(w^*) \leq B_1(w^*) \leq B_k$, so
$\omega_i \geq e^{-\beta B_k}$ and
$\sqrt{\omega_j/\omega_i} \leq e^{\beta B_k/2}$.
For the edges incident to the neighbor $w^{*,(j)}$, flipping bit $j$ changes
the variation of bit $i$ by one interaction entry,
$\mathrm{BV}_i(w^{*,(j)}) = \mathrm{BV}_i(w^*) + H_{ij}(w^*)$, so
$|\mathrm{BV}_i(w^{*,(j)})| \leq B_k + h$, which gives
$\omega'_i \geq e^{-\beta(B_k+h)}$ and
$\sqrt{\omega_j/\omega'_i} \leq e^{\beta(B_k+h)/2}$.

Applying these bounds to the negative sums from \cref{lem:ricci_simplified}:
\begin{align*}
  \text{Negative Terms} &\leq \sum_{i \neq j} \omega(w^*)\, e^{\beta B_k/2}
  + \sum_{i \neq j} \omega(w^{*,(j)})\, e^{\beta(B_k+h)/2} \\
  &\leq (n-1)\, e^{-\beta \ell^*}\, e^{\beta B_k/2}\bigl(1 + e^{\beta h/2}\bigr),
\end{align*}
where the second step uses $\omega(w^{*,(j)}) \leq \omega(w^*) = e^{-\beta \ell^*}$,
since $w^*$ is a local minimum of the empirical loss and its neighbor has equal
or higher loss.

Subtracting this upper bound from the positive term lower bound yields
\[
  \mathrm{Ric}_F(e_j) \geq e^{-\beta \ell^*}\Bigl[1 + e^{-\beta B_k}
  - (n-1)\,e^{\beta B_k/2}\bigl(1 + e^{\beta h/2}\bigr)\Bigr].
\]
The right-hand side does not depend on $j$, so averaging over all $n$ edges
yields~\eqref{eq:avg_curvature_bound}.

\emph{Monotonicity in $B_k$:}  As $B_k, h \to 0$, the terms
$e^{-\beta B_k}$, $e^{\beta B_k/2}$ and $e^{\beta h/2}$ all tend to $1$, and the
bound reduces to $e^{-\beta \ell^*} [2 - 2(n-1)]$, exactly recovering the
curvature of the unweighted hypercube scaled by the node weight
(\cref{prop:curvature}(ii)).  As $B_k$ decreases, that is, as the basin becomes
flatter, the positive term $e^{-\beta B_k}$ grows and the subtracted penalty
$e^{\beta B_k/2}$ shrinks, monotonically increasing the curvature lower bound.
\end{proof}

\begin{remark}[On the local minimum assumption]
\label{rem:ricci_assumption}
The assumption that $w^*$ is a local minimum of
$\hat{\mathcal{L}}$ (not just of $\mathcal{L}_{\mathrm{adv}}$)
is used in the proof to ensure $0 \leq \mathrm{BV}_i(w^*) \leq B_k$ and
$\omega(w^{*,(j)}) \leq \omega(w^*)$ for all neighbors.
This is mild in practice.  The convergence analysis
(\cref{thm:convergence,thm:convergence_sam}) shows that
Hamming-SAM drives $\hat{\mathcal{L}}(w^t)$ to an
$\eta$-stationary point where
$\mathrm{BV}_i(w^t) \geq -\eta$ for all $i$, so the
assumption holds approximately at convergence with
tolerance $\eta$.
\end{remark}

\begin{remark}[Interpretation]
  \cref{thm:avg_curvature} formalizes the intuition that Hamming-SAM
  converges to regions of the Hamming graph with higher Forman-Ricci
  curvature.  Informally, high curvature on this graph means that the loss
  and the Boolean Variations vary little across the edges incident to a
  vertex, which is the combinatorial counterpart of a basin with slow loss
  variation.  In the ablation of \cref{sec:exp_landscape}, curvature is highest
  at $k{=}10000$, which has the flattest neighborhood but the lowest test
  accuracy, so curvature measures local flatness rather than
  generalization.
\end{remark}

\subsection{Estimating the Average Curvature}
\label{sec:ricci_estimation}

Evaluating $\overline{\mathrm{Ric}}_F$ exactly requires all $n$ edges at $w$ and
all $n-1$ edges at each neighbor.  We estimate it by Monte Carlo, after
projecting the weights onto $\{-1,+1\}^n$.  We sample $200$ edges $e_j$
uniformly, and for each, estimate the two negative sums of
\cref{lem:ricci_simplified} from the same $50$ uniformly sampled bits at $w$
and at $w^{(j)}$, rescaled by $(n-1)/50$ (random seed $42$).  All node and edge
weights use exact finite differences of the cross-entropy on a fixed batch of
$256$ training images, with $\beta = 1$.  The two-hop weights $\omega'_i$ are
measured directly at $w^{(j)}$, so the estimate does not rely on the
bounded-interaction assumption.  The values reported in
\cref{sec:exp_landscape} are this average divided by $(n-1)\,\omega(w)$, which
removes the dependence on the network size and on the loss level and equals
$-2$ in the flat-basin limit of \cref{prop:curvature}.

\subsection{Worked Example: Curvature on $\{-1,+1\}^3$}
\label{sec:ricci_example}

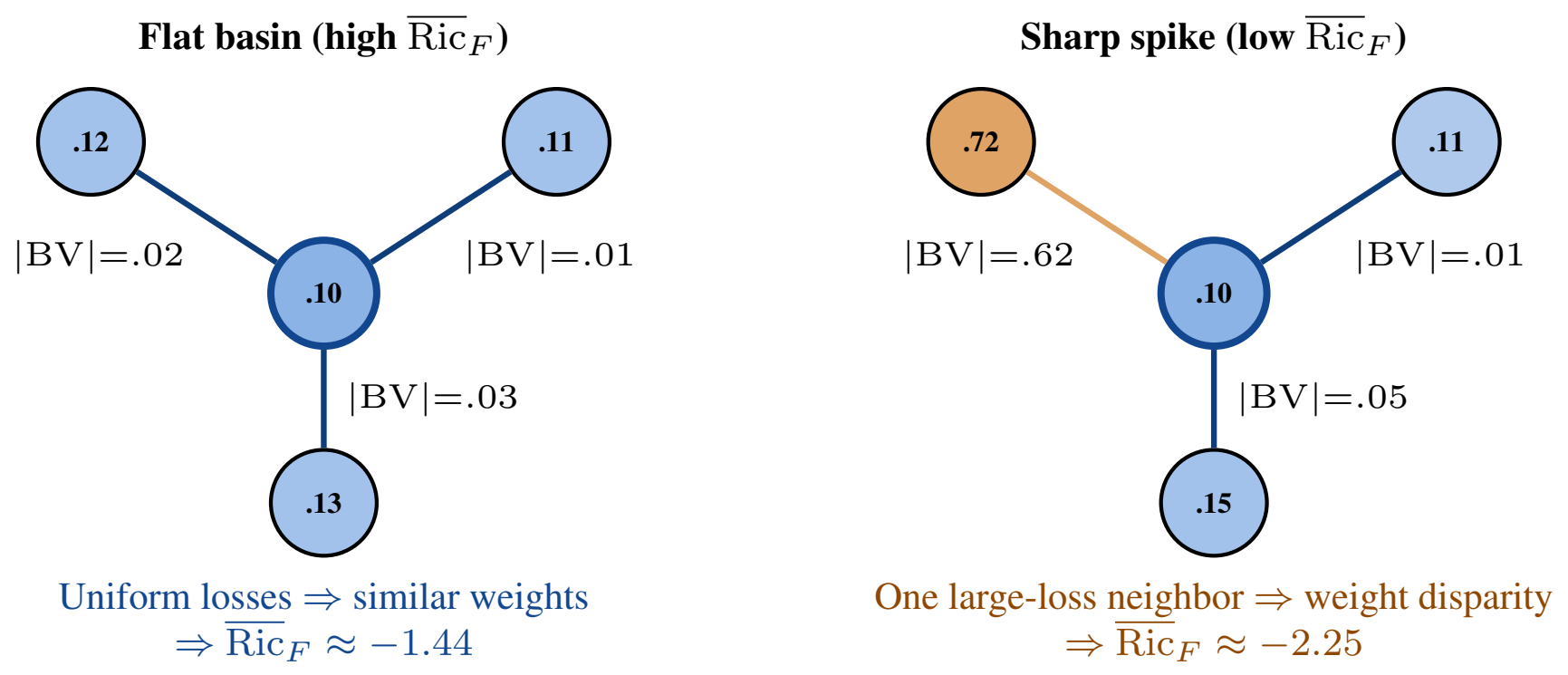
\begin{figure}[t]
  \centering

  \tikzexternaldisable

  \tikzexternaldisable

\definecolor{cFlat}{rgb}{0.1, 0.4, 0.8}  %
\definecolor{cSharp}{rgb}{0.8, 0.4, 0.0} %

\begin{tikzpicture}[>=Stealth, font=\small,
    vnd/.style={draw, circle, thick, minimum size=22pt,
                inner sep=0pt, font=\tiny\bfseries}]

    \node[anchor=south, font=\bfseries\footnotesize] at (2.0,3.4)
      {Flat basin (high $\overline{\mathrm{Ric}}_F$)};

    \begin{scope}[shift={(2.0,1.8)}, scale=0.85]
        \node[vnd, fill=cFlat!50, line width=1.5pt, draw=cFlat!70!black]
          (fc) at (0,0) {.10};
        \node[vnd, fill=cFlat!40] (fn1) at (-2.0, 1.3) {.12};
        \node[vnd, fill=cFlat!40] (fn2) at (2.0, 1.3) {.11};
        \node[vnd, fill=cFlat!40] (fn3) at (0, -1.8) {.13};

        \draw[very thick, cFlat!60!black] (fc)--(fn1)
          node[midway, above left, font=\tiny, black, scale=1.18, yshift=-14pt]{$|\mathrm{BV}|{=}.02$};
        \draw[very thick, cFlat!60!black] (fc)--(fn2)
          node[midway, above right, font=\tiny, black, scale=1.18, yshift=-14pt]{$|\mathrm{BV}|{=}.01$};
        \draw[very thick, cFlat!60!black] (fc)--(fn3)
          node[midway, right, font=\tiny, black, scale=1.18]{$|\mathrm{BV}|{=}.03$};
    \end{scope}

    \node[font=\footnotesize, align=center, cFlat!70!black] at (2.0,-0.6)
      {Uniform losses $\Rightarrow$ similar weights\\
       $\Rightarrow$ $\overline{\mathrm{Ric}}_F \approx -1.44$};

    \begin{scope}[xshift=6.5cm]
      \node[anchor=south, font=\bfseries\footnotesize] at (2.0,3.4)
        {Sharp spike (low $\overline{\mathrm{Ric}}_F$)};

      \begin{scope}[shift={(2.0,1.8)}, scale=0.85]
          \node[vnd, fill=cFlat!50, line width=1.5pt, draw=cFlat!70!black]
            (sc) at (0,0) {.10};
          \node[vnd, fill=cSharp!60] (sn1) at (-2.0, 1.3) {.72};
          \node[vnd, fill=cFlat!35]  (sn2) at (2.0, 1.3) {.11};
          \node[vnd, fill=cFlat!40]  (sn3) at (0, -1.8) {.15};

          \draw[very thick, cSharp!60] (sc)--(sn1)
            node[midway, above left, font=\tiny, black, scale=1.18, yshift=-14pt]{$|\mathrm{BV}|{=}.62$};
          \draw[very thick, cFlat!60!black] (sc)--(sn2)
            node[midway, above right, font=\tiny, black, scale=1.18, yshift=-14pt]{$|\mathrm{BV}|{=}.01$};
          \draw[very thick, cFlat!60!black] (sc)--(sn3)
            node[midway, right, font=\tiny, black, scale=1.18]{$|\mathrm{BV}|{=}.05$};
      \end{scope}

      \node[font=\footnotesize, align=center, cSharp!70!black] at (2.0,-0.6)
        {One large-loss neighbor $\Rightarrow$ weight disparity\\
         $\Rightarrow$ $\overline{\mathrm{Ric}}_F \approx -2.25$};
    \end{scope}
\end{tikzpicture}

  \tikzexternalenable

  \caption{\small \textbf{Forman-Ricci curvature on the loss-weighted Hamming
    graph} (one vertex and its $n{=}3$ neighbors, $\beta{=}3$).
    \emph{Left:}~Flat basin with losses $0.10$--$0.13$.  Nearly uniform
    weights give $\overline{\mathrm{Ric}}_F \approx -1.44$.
    \emph{Right:}~Sharp spike (one neighbor at $0.72$).  The low-weight edge
    lowers the curvature of the other two edges, giving
    $\overline{\mathrm{Ric}}_F \approx -2.25$.  Hamming-SAM favors the left
    configuration.}
  \label{fig:forman_ricci}
\end{figure}

To make the curvature analysis concrete, we compute $\mathrm{Ric}_F$ for the two
scenarios of \cref{fig:forman_ricci}, assuming no pairwise interaction, so that
the edges incident to $w^{(j)}$ carry the same $|\mathrm{BV}_i|$ as those
incident to $w$.  We use $\beta = 3$ so that the contrast is visible with three
neighbors.  The measurements of \cref{sec:exp_landscape} use $\beta = 1$
(\cref{sec:ricci_estimation}).

\paragraph{Flat basin.}  The center has loss $0.10$ and its neighbors
$0.12, 0.11, 0.13$, so $\omega(w) = e^{-0.30} \approx 0.741$.  All three edges
have similar weight, and \cref{lem:ricci_simplified} gives
$\mathrm{Ric}_F(e_1, e_2, e_3) \approx (-1.44, -1.53, -1.35)$, an average of
$\overline{\mathrm{Ric}}_F \approx -1.44$.

\paragraph{Sharp spike.}  Replacing the first neighbor by a loss of $0.72$
gives the large-variation edge a small weight, $\omega(e_1) = e^{-1.86} \approx 0.156$.
Because $\omega(e_1)$ enters the numerator of its own ratios, this edge becomes
mildly positively curved ($\mathrm{Ric}_F(e_1) \approx 0.15$).  It also appears in the
denominator of the ratios of the two other edges, whose curvatures drop to
$-3.73$ and $-3.16$.  The vertex average falls to
$\overline{\mathrm{Ric}}_F \approx -2.25$, about $56\%$ more negative than in the
flat basin.  It is the average over incident edges, the quantity bounded in
\cref{thm:avg_curvature} and estimated in \cref{sec:ricci_estimation}, that
separates the two neighborhoods.

\subsection{Discussion}

\begin{remark}[Novelty]
  While Forman-Ricci curvature has been applied to data
  graphs~\citep{topping2022understanding}, this is, to our knowledge, the
  first application to a neural network's \emph{parameter space}.  The
  loss-weighted Hamming graph construction
  (\cref{def:weighted_hamming}) is original to this work.
\end{remark}

\begin{remark}[Relation to other curvatures]
  Several discrete Ricci curvatures exist.  The discrete Bonnet--Myers theorem
  of~\cite{ollivier2009ricci}, which links positive curvature to diameter and
  random-walk mixing, is stated for Ollivier (coarse) Ricci curvature.  The
  Forman curvature used here is a different notion, which we adopt because it
  is computable directly from local node and edge weights.
\end{remark}

\section{BDM Regularization: Full Analysis and Empirical Evidence}
\label{app:bdm}

The theoretical framework developed so far characterizes flatness
through worst-case loss (\cref{sec:method}), graph curvature
(\cref{app:ricci}), and PAC-Bayes complexity
(\cref{sec:pac_bayes}).  In this appendix, we introduce a complementary
perspective from algorithmic information theory.  \emph{Simpler}
weight configurations, those with lower algorithmic complexity, should
generalize better, since they encode fewer arbitrary choices and are
therefore less likely to overfit.  For Boolean networks, this intuition
can be made concrete using the Block Decomposition Method (BDM), which
approximates Kolmogorov complexity for finite binary
strings~\citep{zenil2018decomposition}.  BDM is
non-differentiable, so gradient-based optimizers cannot use it.  The
gradient-free optimization of Hamming-SAM allows its direct use as a
training-time regularizer, addressing an open problem raised in recent
work~\citep{bnn_compression}.

\subsection{The Block Decomposition Method}

\begin{definition}[Block Decomposition Method~\citep{zenil2018decomposition}]
\label{def:bdm}
Let $s \in \{0,1\}^n$ be a binary string (obtained from a Boolean weight
vector $w \in \{-1,+1\}^n$ via the map $s_i = (1+w_i)/2$).  The BDM
partitions $s$ into non-overlapping blocks $b_1, b_2, \ldots, b_M$ of
fixed size $d$ (typically $d = 12$), with each block's algorithmic
complexity estimated by the Coding Theorem Method (CTM):
\begin{equation}
  \mathrm{CTM}(b) = -\log_2 \mathbb{P}_U(b),
  \label{eq:ctm}
\end{equation}
where $\mathbb{P}_U(b)$ is the probability that a random Turing machine
of bounded size produces $b$, precomputed and tabulated for all
$2^d$ strings of length $d$.  The BDM aggregates block complexities with
a logarithmic correction for repeated patterns:
\begin{equation}
  \mathrm{BDM}(s) = \sum_{u \in \mathcal{U}}
    \bigl[\mathrm{CTM}(u) + \log_2 n_u\bigr],
  \label{eq:bdm}
\end{equation}
where $\mathcal{U}$ is the set of distinct block values occurring in the
decomposition and $n_u$ is the
multiplicity of block value $u$.  Intuitively, $\mathrm{BDM}(s)$ is small
when $s$ consists of few distinct simple blocks repeated many times
(high compressibility), and large when $s$ contains many distinct
complex blocks (low compressibility).
\end{definition}

\begin{remark}[BDM as an approximation to Kolmogorov complexity]
  While Kolmogorov complexity $K(s)$ is uncomputable, BDM is computable and
  upper-bounds $K(s)$ up to logarithmic terms.  It is in turn bounded above
  by a block-entropy term, to which it reduces when all blocks look
  random~\citep{zenil2018decomposition}.  Between these two regimes, BDM
  captures repeated structure that Shannon entropy does not see.  This matters for Boolean
  weight matrices, which develop repeated motifs during training, as observed
  empirically by \cite{bnn_compression}.
\end{remark}

\subsection{BDM-Regularized Hamming-SAM Objective}

The key observation enabling BDM regularization is that Hamming-SAM
operates via discrete bit-flips, not gradients.  Since the BDM of a
binary string can be evaluated exactly for any configuration, and a
single bit-flip changes exactly one block of the decomposition, the
\emph{change} in BDM under a bit-flip is efficiently computable.

\begin{definition}[BDM-Regularized Objective]
\label{def:bdm_objective}
For a regularization parameter $\lambda \geq 0$, define the
BDM-regularized Hamming-SAM objective:
\begin{equation}
  \mathcal{L}_{\mathrm{adv}}^{\mathrm{BDM}}(w) \;\triangleq\;
  \textcolor{BrickRed}{\mathcal{L}_{\mathrm{adv}}(w)}
  + \textcolor{NavyBlue}{\lambda \cdot \mathrm{BDM}(w)},
  \label{eq:bdm_reg}
\end{equation}
and the corresponding BDM-augmented Boolean Variation:
\begin{equation}
  \mathrm{BV}_i^{\mathrm{BDM}}(w)
  \;\triangleq\; \textcolor{BrickRed}{\mathrm{BV}_i(w)}
  + \textcolor{NavyBlue}{\lambda\bigl[\mathrm{BDM}(w^{(i)}) - \mathrm{BDM}(w)\bigr]},
  \label{eq:bv_bdm}
\end{equation}
where $\mathrm{BV}_i(w) = \hat{\mathcal{L}}(w^{(i)}) -
\hat{\mathcal{L}}(w)$ is the standard Boolean Variation.
\end{definition}

The BDM-augmented Boolean Variation~\eqref{eq:bv_bdm} replaces the
standard $\mathrm{BV}$ in \cref{alg:hamming_sam} without any
other modification.  Phase~1 selects the at most $k$ bits with largest positive
$\mathrm{BV}_i^{\mathrm{BDM}}$ (adversarial probing that also seeks
high-complexity perturbations), and Phase~2 flips the bit with most
negative $\mathrm{BV}_i^{\mathrm{BDM}}$ at the adversarial point
(descent toward both low loss and low complexity).

\begin{mybox}
  \begin{proposition}[Efficient BDM Variation]
    \label{prop:bdm_efficient}
    For a weight vector $w \in \{-1,+1\}^n$ partitioned into
    non-overlapping blocks of size $d$, flipping bit $i$ changes exactly
    one block of the decomposition and therefore at most two entries of
    the multiplicity table.  Consequently:
    \begin{enumerate}
      \item[\emph{(i)}] $\mathrm{BDM}(w^{(i)}) - \mathrm{BDM}(w)$ can be
      computed in $O(1)$ time, given the precomputed CTM table and the
      current block decomposition.
      \item[\emph{(ii)}] The full vector of BDM differences
      $\bigl(\mathrm{BDM}(w^{(i)}) - \mathrm{BDM}(w)\bigr)_{i=1}^n$ can be
      computed in $O(n)$ time, so the BDM-augmented Boolean Variation adds
      only $O(n)$ operations to the backward pass that produces the standard
      Boolean Variations.
    \end{enumerate}
  \end{proposition}
\end{mybox}

\begin{proof}
\emph{Part~(i).}  The BDM partition divides $s \in \{0,1\}^n$ into
$M = \lceil n/d \rceil$ non-overlapping blocks, so position $i$ belongs
to exactly one block, whose value changes from some $u$ to some $u'$
under the flip.  By~\eqref{eq:bdm}, the only affected terms are those of
$u$ and $u'$: the multiplicity $n_u$ decreases by one and $n_{u'}$
increases by one, with $\mathrm{CTM}$ values read from the precomputed
table.  Updating the sum therefore requires a constant number of
operations.

\emph{Part~(ii).}  Each of the $n$ bits affects one block, so
computing all $n$ differences $\mathrm{BDM}(w^{(i)}) -
\mathrm{BDM}(w)$ requires $O(n)$ total operations.
\end{proof}

The same argument applies to the 2D spatial variant used in
\cref{sec:bdm_resnet}, where each bit belongs to exactly one $3 \times 3$
filter block.

\subsection{Convergence of BDM-Regularized Hamming-SAM}

We now check whether adding the BDM regularizer preserves
the convergence guarantees of \cref{thm:convergence}.  The
following result confirms this for the base-point descent rule.

\begin{mybox}
  \begin{proposition}[Convergence with BDM Regularization]
    \label{prop:bdm_convergence}
    Let $B_{\max} = \max_w \hat{\mathcal{L}}(w)$ and
$C_{\max} = \max_w \mathrm{BDM}(w)$.  If \cref{alg:hamming_sam}
is run with $\mathrm{BV}^{\mathrm{BDM}}$ in place of $\mathrm{BV}$
and accepts bit-flips only when
$\mathrm{BV}^{\mathrm{BDM}}_{i^*}(w^t) < -\eta$, evaluated at the base point as
in \cref{thm:convergence}, for threshold
$\eta > 0$, then the algorithm converges to an $\eta$-stationary
point of $\hat{\mathcal{L}}(w) + \lambda\,\mathrm{BDM}(w)$ in at most
\begin{equation}
  T \;\leq\; \left\lfloor
    \frac{B_{\max} + \lambda\,C_{\max}}{\eta}
  \right\rfloor
\end{equation}
iterations.
  \end{proposition}
\end{mybox}

\begin{proof}
Define the Lyapunov function
$\Phi^{\mathrm{BDM}}(w^t) = \hat{\mathcal{L}}(w^t)
+ \lambda\,\mathrm{BDM}(w^t)$.  The
BDM-augmented Boolean Variation
$\mathrm{BV}^{\mathrm{BDM}}_i(w) = \mathrm{BV}_i(w) +
\lambda[\mathrm{BDM}(w^{(i)}) - \mathrm{BDM}(w)]$ equals
exactly $\Phi^{\mathrm{BDM}}(w^{(i)}) - \Phi^{\mathrm{BDM}}(w)$.
Since the accepted flip is applied to $w^t$ and the acceptance test
$\mathrm{BV}^{\mathrm{BDM}}_{i^*}(w^t) < -\eta$ is evaluated at $w^t$, each
accepted step gives
$\Phi^{\mathrm{BDM}}(w^{t+1}) \leq \Phi^{\mathrm{BDM}}(w^t) - \eta$, as in
\cref{thm:convergence}.
Since $\Phi^{\mathrm{BDM}} \geq 0$ and
$\Phi^{\mathrm{BDM}}(w^0) \leq B_{\max} + \lambda\,C_{\max}$,
at most $\lfloor(B_{\max} + \lambda\,C_{\max})/\eta\rfloor$
accepted steps are possible.
\end{proof}

\subsection{Connection to the PAC-Bayes Framework}

The BDM regularizer can be understood as an \emph{algorithmic
complexity prior} within the PAC-Bayes framework.  The following
result makes this connection precise.

\begin{mybox}
  \begin{proposition}[BDM as a Coding-Length Prior]
    \label{prop:bdm_pac_bayes}
    Define a prior $P_{\mathrm{BDM}}$ over $\{-1,+1\}^n$ by
    $P_{\mathrm{BDM}}(w) \propto 2^{-\mathrm{BDM}(w)}$, assigning higher
    probability to algorithmically simpler configurations.  Then for the
    deterministic posterior $Q = \delta_{\hat{w}}$:
    \begin{equation}
      \mathrm{KL}(Q \| P_{\mathrm{BDM}})
      = \mathrm{BDM}(\hat{w}) \cdot \log 2 + \log Z_{\mathrm{BDM}},
      \label{eq:kl_bdm}
    \end{equation}
    where $Z_{\mathrm{BDM}} = \sum_{w} 2^{-\mathrm{BDM}(w)}$ is the
    normalizing constant.  For a loss with values in $[0,1]$, the PAC-Bayes
    bound with this prior is increasing in both
    $\hat{\mathcal{L}}(\hat{w})$ and $\mathrm{BDM}(\hat{w})$.
  \end{proposition}
\end{mybox}

\begin{proof}
For the deterministic posterior $Q = \delta_{\hat{w}}$:
\begin{align}
  \mathrm{KL}(Q \| P_{\mathrm{BDM}})
  &= -\log P_{\mathrm{BDM}}(\hat{w})
   = -\log\frac{2^{-\mathrm{BDM}(\hat{w})}}{Z_{\mathrm{BDM}}}
     \nonumber \\
  &= \mathrm{BDM}(\hat{w})\cdot\log 2 + \log Z_{\mathrm{BDM}}.
\end{align}
The prior does not depend on the data, so substituting into the McAllester
PAC-Bayes bound~\citep{mcallester2003pac} gives
\[
  \mathcal{L}(\hat{w}) \leq \hat{\mathcal{L}}(\hat{w})
  + \sqrt{\frac{\mathrm{BDM}(\hat{w})\cdot\log 2
    + \log Z_{\mathrm{BDM}}
    + \log(2\sqrt{m}/\delta)}{2m}}.
\]
The constant $\log Z_{\mathrm{BDM}}$ does not depend on $\hat{w}$, so the bound
depends on the solution only through $\hat{\mathcal{L}}(\hat{w})$ and
$\mathrm{BDM}(\hat{w})$, and is increasing in both.
\end{proof}

Minimizing $\hat{\mathcal{L}} + \lambda\,\mathrm{BDM}$ therefore decreases both
data-dependent terms of this bound, with $\lambda$ playing the role of the
local exchange rate between them.

\begin{remark}[Why BDM and not Shannon entropy?]
  One might consider regularizing with the Shannon entropy of the weights,
  $n\,H_{\mathrm{b}}(\hat{p})$ in bits, where $\hat{p}$ is the fraction of $+1$
  entries and $H_{\mathrm{b}}$ the binary entropy.
  However, Shannon entropy measures only first-order symbol frequencies
  and is maximized by a random bit-string, which is the opposite of
  what we want.  BDM captures \emph{structural} simplicity.  A weight
  matrix with repeated motifs (e.g., convolutional filter symmetries or
  tiled logic patterns) has low BDM but may have near-maximal Shannon
  entropy.  The learning-as-compression hypothesis~\citep{bnn_compression}
  specifically predicts that trained BNNs develop such structured motifs,
  making BDM the appropriate complexity measure.
\end{remark}

\begin{remark}[Resolving the differentiability barrier]
  \cite{bnn_compression} empirically tracked BDM during BNN training and
  observed that weight matrices converge toward lower algorithmic
  complexity, but explicitly concluded that BDM ``is inherently
  non-continuous and non-differentiable,'' making it ``unsuitable as a
  training-time regularizer.''  Hamming-SAM removes this barrier.
  Since it uses bit-flips rather than gradients, the BDM-augmented Boolean
  Variation~\eqref{eq:bv_bdm} provides exact, efficiently computable
  descent directions for the non-differentiable BDM objective.  To our
  knowledge, this is the first use of algorithmic complexity
  as a \emph{training-time} regularizer rather than a post-hoc
  diagnostic.
\end{remark}

\subsection{Empirical Evaluation}

We test whether a non-differentiable complexity penalty can be optimized
directly within Hamming-SAM, by augmenting its objective with the BDM of the
Boolean weights.  The penalty favors configurations built from simpler,
repeated structural motifs.

\subsubsection{Boolean MLP}

\paragraph{Experimental Setup.} We first evaluate this regularizer on a Boolean MLP trained on CIFAR-10.
We apply a global 1D BDM penalty to the concatenated Boolean weight vectors, controlled by a scaling hyperparameter $\lambda$,
alongside the neighborhood perturbation for Hamming-SAM ($k=50$). The empirical results are detailed in \cref{tab:mlp_bdm}.

\begin{table}[H]
    \centering
    \caption{\small Algorithmic complexity and generalization metrics for a Boolean MLP on CIFAR-10. BDM represents the global 1D algorithmic complexity approximation, while Shannon entropy (total bits, $n\,H_{\mathrm{b}}(\hat{p})$) tracks first-order bit statistics. Unique block values are measured against the $36{,}864$ blocks of length $d=12$ in the decomposition, of which at most $2^{12}=4{,}096$ can be distinct.}
    \label{tab:mlp_bdm}
    \centering

\newcommand{\sd}[1]{{\color{gray}\scriptsize $\pm$ #1}}

\resizebox{0.93\textwidth}{!}{%
    \begin{tabular}{lccccc}
        \toprule
        \textbf{Method}         & \textbf{$\lambda$} & \textbf{Test Acc (\%)}   & \textbf{BDM Score} & \textbf{Shannon Entropy} & \textbf{Unique Blocks} \\ \midrule
        Standard BOLD           & -                  & 51.58 \sd{0.12}          & 139,403 \sd{14}    & 442,368 \sd{5}           & 3,968           \\
        Hamming-SAM ($k=50$)       & 0                  & 52.11 \sd{0.09}          & 138,031 \sd{11}    & 442,360 \sd{4}           & 3,932           \\ \midrule
        Hamming-SAM + BDM          & $10^{-4}$          & 52.88 \sd{0.07}          & 138,107 \sd{9}     & 442,364 \sd{3}           & 3,936           \\
        \textbf{Hamming-SAM + BDM} & \textbf{$10^{-3}$} & \textbf{52.93} \sd{0.05} & 138,341 \sd{12}    & 442,362 \sd{4}           & 3,942           \\
        Hamming-SAM + BDM          & $10^{-2}$          & 51.57 \sd{0.11}          & 138,691 \sd{15}    & 442,365 \sd{6}           & 3,950           \\
        Hamming-SAM + BDM          & $10^{-1}$          & 51.27 \sd{0.14}          & 138,577 \sd{13}    & 442,365 \sd{5}           & 3,943           \\ \bottomrule
    \end{tabular}%
}

\end{table}

\paragraph{The effect of BDM regularization on generalization.}
Adding a moderate BDM penalty ($\lambda \in [10^{-4}, 10^{-3}]$)
improves test accuracy (peak 52.93\% at $\lambda = 10^{-3}$,
$+0.82$ points over unregularized Hamming-SAM).  Stronger
regularization ($\lambda \geq 10^{-2}$) degrades performance.
The final BDM scores remain close to the unregularized value, indicating that
the penalty acts mainly on the optimization trajectory by discouraging complex
intermediate configurations, much as weight decay can improve generalization
without producing the smallest-norm solution.

\paragraph{Flatness implicitly reduces algorithmic complexity.}
Comparing standard BOLD to unregularized Hamming-SAM ($\lambda = 0$),
sharpness-aware training reduces the BDM score from 139,403 to 138,031, about
$1\%$, and the number of unique blocks from 3,968 to 3,932, without any
explicit complexity penalty.  The effect is small but consistent with flat
basins having slightly more regular structure.

\paragraph{The blindness of statistical entropy.} Across all configurations in \cref{tab:mlp_bdm}, the Shannon entropy barely changes.
In contrast, the BDM score fluctuates and unique block counts decrease under regularization.
This occurs because the global bit-balance of the Boolean weights remains near 50/50.
As a statistical metric, Shannon entropy only measures the overall frequency of bits and ignores their arrangement.
BDM, however, is sensitive to the arrangement and reuse of local motifs.
The static entropy also indicates that the regularizer does not collapse the weights toward a degenerate configuration.

\subsubsection{2D Spatial BDM on ResNet-18}
\label{sec:bdm_resnet}

\paragraph{The global saturation limit.}
Applying a global 1D algorithmic complexity penalty to large convolutional networks has a practical limit.
The maximum number of distinct 1D blocks of length $12$ ($2^{12} = 4{,}096$) is quickly saturated by the large number of parameters.
Under global saturation, standard BDM and Shannon entropy become insensitive to structural shifts in the optimization trajectory.

\paragraph{Transitioning to 2D spatial regularization.}
To resolve this, we move from a global 1D penalty to a local 2D spatial BDM penalty.
Rather than flattening the entire network, we regularize the algorithmic complexity of the individual $3 \times 3$ convolutional filters during training, encouraging spatially smooth, low-complexity feature extractors.
We evaluate this approach on a Boolean ResNet-18 (\cref{tab:resnet_bdm}).

\begin{table}[H]
    \centering
    \caption{\small
      Generalization and global complexity metrics for a Boolean ResNet-18 on CIFAR-10. The Hamming-SAM + BDM models were optimized using a localized 2D spatial BDM penalty on the $3 \times 3$ convolutional filters. Global BDM, Shannon entropy (total bits) and unique block values (against the $929{,}792$ blocks of the decomposition) are reported post-training, illustrating the saturation of 1D global metrics in deep architectures.
    }
    \label{tab:resnet_bdm}
    \centering

\newcommand{\sd}[1]{{\color{gray}\scriptsize $\pm$ #1}}

\resizebox{0.98\textwidth}{!}{%
    \begin{tabular}{lcccccc}
        \toprule
        \textbf{Method}         & \textbf{$\lambda$} & \textbf{Test Acc (\%)}   & \textbf{Test Loss}        & \textbf{Global BDM} & \textbf{Shannon Entropy} & \textbf{Unique Blocks} \\ \midrule
        Standard BOLD           & -                  & 92.99 \sd{0.11}          & 0.566 \sd{0.012}          & 163,039 \sd{18}     & 11,149,042 \sd{85}       & 4,096           \\
        Hamming-SAM ($k=100$)      & 0                  & 93.22 \sd{0.08}          & 0.399 \sd{0.007}          & 163,108 \sd{14}     & 11,150,255 \sd{62}       & 4,096           \\ \midrule
        Hamming-SAM + BDM          & $10^{-5}$          & 93.35 \sd{0.06}          & 0.391 \sd{0.005}          & 163,077 \sd{12}     & 11,150,965 \sd{45}       & 4,096           \\
        \textbf{Hamming-SAM + BDM} & \textbf{$10^{-4}$} & \textbf{93.63} \sd{0.04} & \textbf{0.392} \sd{0.004} & 163,123 \sd{10}     & 11,150,910 \sd{38}       & 4,096           \\
        Hamming-SAM + BDM          & $10^{-3}$          & 93.42 \sd{0.07}          & 0.442 \sd{0.008}          & 163,148 \sd{15}     & 11,150,529 \sd{55}       & 4,096           \\
        Hamming-SAM + BDM          & $10^{-2}$          & 93.37 \sd{0.09}          & 0.476 \sd{0.010}          & 163,075 \sd{16}     & 11,149,457 \sd{70}       & 4,096           \\
        Hamming-SAM + BDM          & $10^{-1}$          & 93.24 \sd{0.12}          & 0.473 \sd{0.011}          & 163,039 \sd{20}     & 11,149,351 \sd{80}       & 4,096           \\ \bottomrule
    \end{tabular}%
}

\end{table}

\paragraph{Effect on generalization.}
Spatial BDM gives a clear regularization curve.
Hamming-SAM alone improves standard training from 92.99\% to 93.22\%.
Adding the spatial penalty improves accuracy further at every $\lambda$ tested, peaking at 93.63\% at $\lambda = 10^{-4}$, a gain of $0.41$ points over Hamming-SAM.
Test loss improves only for $\lambda \leq 10^{-4}$ and rises above the Hamming-SAM value for larger penalties.

\paragraph{Algorithmic versus statistical complexity at scale.}
The global evaluation metrics in \cref{tab:resnet_bdm} further show the difference between statistical and algorithmic complexity at scale.
Across all optimization strategies, the global Shannon entropy remains virtually static, varying by less than 0.02\%.
Because the network maintains a highly balanced ratio of $+1$ and $-1$ weights, statistical entropy assigns the networks nearly identical complexity.
The effect of the penalty is therefore not visible in global statistics, consistent with its acting locally on the $3 \times 3$ filters it targets.

\subsubsection{Discussion}

The two experiments show that a non-differentiable complexity penalty can be optimized directly within Hamming-SAM, improving test accuracy on top of sharpness-aware training.
Characterizing how the penalty shapes the weights, for instance by tracking per-filter complexity during training, is a natural direction for future work.

\section{Additional Results}
\label{app:extra_experiments}

\subsection{Tabular Version of Benchmark Results}

\cref{tab:benchmark_cifar10,tab:benchmark_cifar100} show the full clean and corruption error rates for all methods and architectures on CIFAR-10 and CIFAR-100, averaged over 4 seeds.
These complement the summary plots in \cref{sec:experiments}.

\begin{table}[H]
    \centering
    \caption{CIFAR-10 clean and corruption error rates (\%) across methods and architectures.  Results averaged over 4 seeds ($\pm$ std shown).}
    \label{tab:benchmark_cifar10}
    \definecolor{color_fp}{HTML}{6A4C93}
\definecolor{color_ste}{HTML}{FFCA3A}
\definecolor{color_stesam}{HTML}{8AC926}
\definecolor{color_bold}{HTML}{1982C4}
\definecolor{color_hamsam}{HTML}{FF595E}

\newcommand{\sd}[1]{{\color{gray}\scriptsize $\pm$ #1}}

\newcommand{\tmark}[1]{{\protect\tikz[baseline=-.65ex] \protect\draw[fill=#1, draw=#1, line width=1.4pt] plot[mark size=2.0, only marks, mark=square*, mark options={xscale=1.5, yscale=1}] (0, 0);}}

\tikzexternaldisable

\scalebox{.80}{
    \setlength{\tabcolsep}{4.5pt}
    \renewcommand{\arraystretch}{1.2}
    \begin{tabular}{llcccccc}
        \toprule
        \multirow{2}{*}{Model} & \multirow{2}{*}{Method} & \multirow{2}{*}{Clean Error (\%) $\downarrow$} & \multicolumn{5}{c}{Corruption Error (\%) $\downarrow$} \\
        \cmidrule(lr){4-8}
        & & & Level 1 & Level 2 & Level 3 & Level 4 & Level 5 \\
        \midrule
        \midrule
        \multirow{5}{*}{ResNet18} & \tmark{color_fp} FP & 5.38 \sd{0.18} & 11.90 \sd{0.10} & 17.53 \sd{0.33} & 23.27 \sd{0.22} & 30.87 \sd{0.18} & 43.18 \sd{0.19} \\
        & \tmark{color_ste} STE & 7.94 \sd{0.06} & 14.40 \sd{0.08} & 20.38 \sd{0.36} & 26.03 \sd{0.75} & 33.57 \sd{1.49} & 44.80 \sd{1.58} \\
        & \tmark{color_stesam} STE-SAM & 7.76 \sd{0.10} & 13.96 \sd{0.13} & 20.35 \sd{0.33} & 27.03 \sd{0.66} & 35.68 \sd{0.95} & 46.88 \sd{1.03} \\
        & \tmark{color_bold} BOLD & 6.99 \sd{0.15} & 14.08 \sd{0.12} & 20.69 \sd{0.43} & 27.40 \sd{0.56} & 35.44 \sd{0.90} & 46.94 \sd{0.64} \\
        & \tmark{color_hamsam} Hamming-SAM & \textbf{6.44} \sd{0.17} & \textbf{12.87} \sd{0.10} & \textbf{18.06} \sd{0.18} & \textbf{23.70} \sd{0.34} & \textbf{31.32} \sd{0.50} & \textbf{42.69} \sd{0.42} \\
        \midrule
        \multirow{5}{*}{VGG16} & \tmark{color_fp} FP & 7.07 \sd{0.20} & 13.60 \sd{0.16} & 19.47 \sd{0.10} & 25.60 \sd{0.08} & 33.18 \sd{0.18} & 44.85 \sd{0.44} \\
        & \tmark{color_ste} STE & 8.72 \sd{0.18} & 15.27 \sd{0.28} & 21.40 \sd{0.37} & 27.70 \sd{0.44} & 35.77 \sd{0.59} & 46.18 \sd{0.67} \\
        & \tmark{color_stesam} STE-SAM & 8.81 \sd{0.11} & 15.27 \sd{0.17} & 21.31 \sd{0.27} & 27.56 \sd{0.44} & 35.63 \sd{0.45} & 46.22 \sd{0.52} \\
        & \tmark{color_bold} BOLD & 8.25 \sd{0.07} & 15.29 \sd{0.15} & 21.57 \sd{0.29} & 27.83 \sd{0.33} & 35.82 \sd{0.25} & 46.28 \sd{0.19} \\
        & \tmark{color_hamsam} Hamming-SAM & \textbf{7.65} \sd{0.08} & \textbf{14.08} \sd{0.10} & \textbf{19.95} \sd{0.10} & \textbf{26.23} \sd{0.26} & \textbf{34.11} \sd{0.30} & \textbf{45.10} \sd{0.45} \\
        \midrule
        \multirow{5}{*}{ViT} & \tmark{color_fp} FP & 19.57 \sd{0.14} & 24.60 \sd{0.14} & 30.26 \sd{0.14} & 34.62 \sd{0.14} & 40.66 \sd{0.14} & 47.74 \sd{0.14} \\
        & \tmark{color_ste} STE & 22.91 \sd{0.59} & 27.51 \sd{0.42} & 32.87 \sd{0.21} & 36.95 \sd{0.10} & 42.43 \sd{0.32} & 48.34 \sd{0.61} \\
        & \tmark{color_stesam} STE-SAM & 23.58 \sd{0.56} & 28.12 \sd{0.11} & 33.19 \sd{0.43} & 37.18 \sd{0.70} & 42.73 \sd{0.94} & 48.72 \sd{1.14} \\
        & \tmark{color_bold} BOLD & 21.04 \sd{0.14} & 24.84 \sd{0.10} & 30.11 \sd{0.23} & 34.15 \sd{0.11} & 39.54 \sd{0.09} & 46.11 \sd{0.15} \\
        & \tmark{color_hamsam} Hamming-SAM & \textbf{20.29} \sd{0.11} & \textbf{24.04} \sd{0.49} & \textbf{29.24} \sd{0.58} & \textbf{33.03} \sd{0.12} & \textbf{38.37} \sd{0.12} & \textbf{44.71} \sd{0.12} \\
        \bottomrule
    \end{tabular}
}
\tikzexternalenable

\end{table}

\begin{table}[H]
    \centering
    \caption{CIFAR-100 clean and corruption error rates (\%) across methods and architectures.  Results averaged over 4 seeds ($\pm$ std shown).}
    \label{tab:benchmark_cifar100}
    \definecolor{color_fp}{HTML}{6A4C93}
\definecolor{color_ste}{HTML}{FFCA3A}
\definecolor{color_stesam}{HTML}{8AC926}
\definecolor{color_bold}{HTML}{1982C4}
\definecolor{color_hamsam}{HTML}{FF595E}

\newcommand{\sd}[1]{{\color{gray}\scriptsize $\pm$ #1}}

\newcommand{\tmark}[1]{{\protect\tikz[baseline=-.65ex] \protect\draw[fill=#1, draw=#1, line width=1.4pt] plot[mark size=2.0, only marks, mark=square*, mark options={xscale=1.5, yscale=1}] (0, 0);}}

\tikzexternaldisable

\scalebox{.80}{
    \setlength{\tabcolsep}{4.5pt}
    \renewcommand{\arraystretch}{1.2}
    \begin{tabular}{llcccccc}
        \toprule
        \multirow{2}{*}{Model} & \multirow{2}{*}{Method} & \multirow{2}{*}{Clean Error (\%) $\downarrow$} & \multicolumn{5}{c}{Corruption Error (\%) $\downarrow$} \\
        \cmidrule(lr){4-8}
        & & & Level 1 & Level 2 & Level 3 & Level 4 & Level 5 \\
        \midrule
        \midrule
        \multirow{5}{*}{ResNet18} & \tmark{color_fp} FP & 24.67 \sd{0.15} & 36.61 \sd{0.34} & 45.37 \sd{0.42} & 51.10 \sd{0.41} & 57.91 \sd{0.41} & 67.86 \sd{0.27} \\
        & \tmark{color_ste} STE & 30.27 \sd{0.45} & 41.00 \sd{0.38} & 49.82 \sd{0.68} & 55.39 \sd{0.69} & 62.13 \sd{0.64} & 70.92 \sd{0.60} \\
        & \tmark{color_stesam} STE-SAM & 30.36 \sd{0.08} & 40.61 \sd{0.04} & 49.35 \sd{0.24} & 55.09 \sd{0.30} & 61.79 \sd{0.22} & 70.47 \sd{0.12} \\
        & \tmark{color_bold} BOLD & 28.29 \sd{0.04} & 39.80 \sd{0.07} & 48.62 \sd{0.06} & 54.14 \sd{0.07} & 60.46 \sd{0.15} & 69.75 \sd{0.02} \\
        & \tmark{color_hamsam} Hamming-SAM & \textbf{27.63} \sd{0.03} & \textbf{37.51} \sd{0.05} & \textbf{45.21} \sd{0.03} & \textbf{50.62} \sd{0.02} & \textbf{57.17} \sd{0.07} & \textbf{66.26} \sd{0.39} \\
        \midrule
        \multirow{5}{*}{VGG16} & \tmark{color_fp} FP & 29.10 \sd{0.27} & 41.17 \sd{0.42} & 49.60 \sd{0.54} & 54.71 \sd{0.51} & 60.80 \sd{0.55} & 70.00 \sd{0.45} \\
        & \tmark{color_ste} STE & 35.35 \sd{0.57} & 44.98 \sd{0.37} & 53.16 \sd{0.28} & 58.11 \sd{0.20} & 64.04 \sd{0.06} & 71.68 \sd{0.13} \\
        & \tmark{color_stesam} STE-SAM & 35.77 \sd{0.50} & 45.22 \sd{0.15} & 53.13 \sd{0.11} & 58.04 \sd{0.21} & 63.73 \sd{0.33} & 71.16 \sd{0.43} \\
        & \tmark{color_bold} BOLD & 32.35 \sd{0.09} & 43.70 \sd{0.07} & 52.37 \sd{0.03} & 57.35 \sd{0.10} & 63.30 \sd{0.10} & 71.38 \sd{0.10} \\
        & \tmark{color_hamsam} Hamming-SAM & \textbf{31.26} \sd{0.11} & \textbf{41.60} \sd{0.18} & \textbf{49.78} \sd{0.18} & \textbf{54.95} \sd{0.20} & \textbf{61.03} \sd{0.26} & \textbf{69.35} \sd{0.24} \\
        \midrule
        \multirow{5}{*}{ViT} & \tmark{color_fp} FP & 43.92 \sd{0.01} & 50.66 \sd{0.03} & 57.64 \sd{0.14} & 61.46 \sd{0.16} & 66.90 \sd{0.12} & 72.57 \sd{0.27} \\
        & \tmark{color_ste} STE & 47.36 \sd{0.23} & 53.77 \sd{0.50} & 59.88 \sd{0.57} & 63.40 \sd{0.46} & 68.36 \sd{0.39} & 73.19 \sd{0.34} \\
        & \tmark{color_stesam} STE-SAM & 47.22 \sd{0.18} & 53.25 \sd{0.64} & 59.22 \sd{0.81} & 62.81 \sd{0.72} & 67.86 \sd{0.62} & 72.71 \sd{0.58} \\
        & \tmark{color_bold} BOLD & 47.51 \sd{0.11} & 53.10 \sd{0.14} & 59.63 \sd{0.05} & 63.18 \sd{0.15} & 68.08 \sd{0.19} & 72.92 \sd{0.11} \\
        & \tmark{color_hamsam} Hamming-SAM & \textbf{45.37} \sd{0.15} & \textbf{50.49} \sd{0.15} & \textbf{56.66} \sd{0.15} & \textbf{60.39} \sd{0.15} & \textbf{65.59} \sd{0.15} & \textbf{71.02} \sd{0.15} \\
        \bottomrule
    \end{tabular}
}
\tikzexternalenable

\end{table}

\subsection{Hamming-SAM on an STE Backend}
\label{app:ste_backend}

To test whether the gains depend on BOLD, we run Hamming-SAM on the STE backend used for the STE and STE-SAM baselines.
The ascent flips the top-$k$ bits ranked by $-w_i\,\partial\hat{\mathcal{L}}/\partial w_i$, by negating the corresponding latent weights.
The descent is the same SGD step on the latent weights as in STE-SAM, with the gradient taken at the adversarial point.
The two methods therefore differ only in the perturbation, and both use two backward passes per step.
All other settings follow \cref{app:exp_details}, and we use $k=500$, the default of the BOLD experiments.

\begin{table}[H]
    \centering
    \caption{CIFAR-10 results on the STE backend.  Clean accuracy (\%, mean $\pm$ std) and mean corruption accuracy on CIFAR-10-C (mCA, \%).  STE and STE-SAM are from \cref{tab:benchmark_cifar10} (4 seeds).  Hamming-SAM uses $k=500$ (4 seeds).}
    \label{tab:ste_backend}
    \small
    \begin{tabular}{llcc}
        \toprule
        \textbf{Model} & \textbf{Method} & \textbf{Clean Acc (\%)} & \textbf{mCA (\%)} \\
        \midrule
        \multirow{3}{*}{ResNet-18} & STE         & 92.06 $\pm$ 0.06 & 72.16 \\
                                   & STE-SAM     & 92.24 $\pm$ 0.10 & 71.22 \\
                                   & Hamming-SAM & \textbf{94.48} $\pm$ 0.18 & \textbf{76.60} \\
        \midrule
        \multirow{3}{*}{VGG-16}    & STE         & 91.28 $\pm$ 0.18 & 70.74 \\
                                   & STE-SAM     & 91.19 $\pm$ 0.11 & 70.80 \\
                                   & Hamming-SAM & \textbf{92.80} $\pm$ 0.02 & \textbf{76.26} \\
        \bottomrule
    \end{tabular}
\end{table}

\begin{figure}[H]
    \centering
    \tikzexternaldisable
    \setlength{\figurewidth}{6.2cm}
    \setlength{\figureheight}{3.5cm}
\begin{tikzpicture}

\definecolor{darkslategray38}{RGB}{38,38,38}
\definecolor{darkslategray51}{RGB}{51,51,51}
\definecolor{lightgray204}{RGB}{204,204,204}
\definecolor{dimgray10784138}{RGB}{234,109,113}
\definecolor{sandybrown23019082}{RGB}{230,190,82}
\definecolor{yellowgreen13318058}{RGB}{133,180,58}

\pgfplotstableread{
lv  m      s
0   7.94   0.06
1   14.40  0.08
2   20.38  0.36
3   26.03  0.75
4   33.57  1.49
5   44.80  1.58
}\resnetste
\pgfplotstableread{
lv  m      s
0   7.76   0.10
1   13.96  0.13
2   20.35  0.33
3   27.03  0.66
4   35.68  0.95
5   46.88  1.03
}\resnetstesam
\pgfplotstableread{
lv  m      s
0   5.52   0.18
1   11.08  0.16
2   15.88  0.10
3   21.33  0.10
4   28.43  0.09
5   40.26  0.29
}\resnethsam
\pgfplotstableread{
lv  m      s
0   8.72   0.18
1   15.27  0.28
2   21.40  0.37
3   27.70  0.44
4   35.77  0.59
5   46.18  0.67
}\vggste
\pgfplotstableread{
lv  m      s
0   8.81   0.11
1   15.27  0.17
2   21.31  0.27
3   27.56  0.44
4   35.63  0.45
5   46.22  0.52
}\vggstesam
\pgfplotstableread{
lv  m      s
0   7.20   0.02
1   12.22  0.19
2   16.62  0.35
3   21.52  0.63
4   28.70  0.91
5   39.63  0.95
}\vgghsam

\pgfplotsset{
  corrplot/.style={
    axis line style={darkslategray38},
    height=\figureheight, width=\figurewidth,
    xbar=0pt, bar width=2.6pt,
    y dir=reverse, ymin=-0.5, ymax=5.5, ytick={0,1,2,3,4,5},
    xmin=0, xmax=52,
    xlabel={Error (\%)},
    tick align=outside, tick pos=left, major tick length=0.6ex,
    xtick style={color=darkslategray38}, ytick style={color=darkslategray38},
    xmajorgrids, x grid style={lightgray204},
    tick label style={font=\scriptsize}, label style={font=\scriptsize},
    title style={font=\scriptsize, align=center, yshift=0.5ex},
  },
  bars/.style={draw=darkslategray51, very thin, fill=#1,
    error bars/.cd, x dir=both, x explicit, error bar style={line width=0.4pt, darkslategray51}},
}

\begin{groupplot}[group style={group size=2 by 1, horizontal sep=0.35cm, y descriptions at=edge left}]

\nextgroupplot[corrplot, title={ResNet18\\CIFAR-10}, ylabel={Corruption intensity}]
\addplot[bars=dimgray10784138]     table[y=lv, x=m, x error=s] {\resnethsam};
\addplot[bars=yellowgreen13318058] table[y=lv, x=m, x error=s] {\resnetstesam};
\addplot[bars=sandybrown23019082]  table[y=lv, x=m, x error=s] {\resnetste};

\nextgroupplot[corrplot, title={VGG16\\CIFAR-10}]
\addplot[bars=dimgray10784138]     table[y=lv, x=m, x error=s] {\vgghsam};
\addplot[bars=yellowgreen13318058] table[y=lv, x=m, x error=s] {\vggstesam};
\addplot[bars=sandybrown23019082]  table[y=lv, x=m, x error=s] {\vggste};

\end{groupplot}
\end{tikzpicture}

    \input{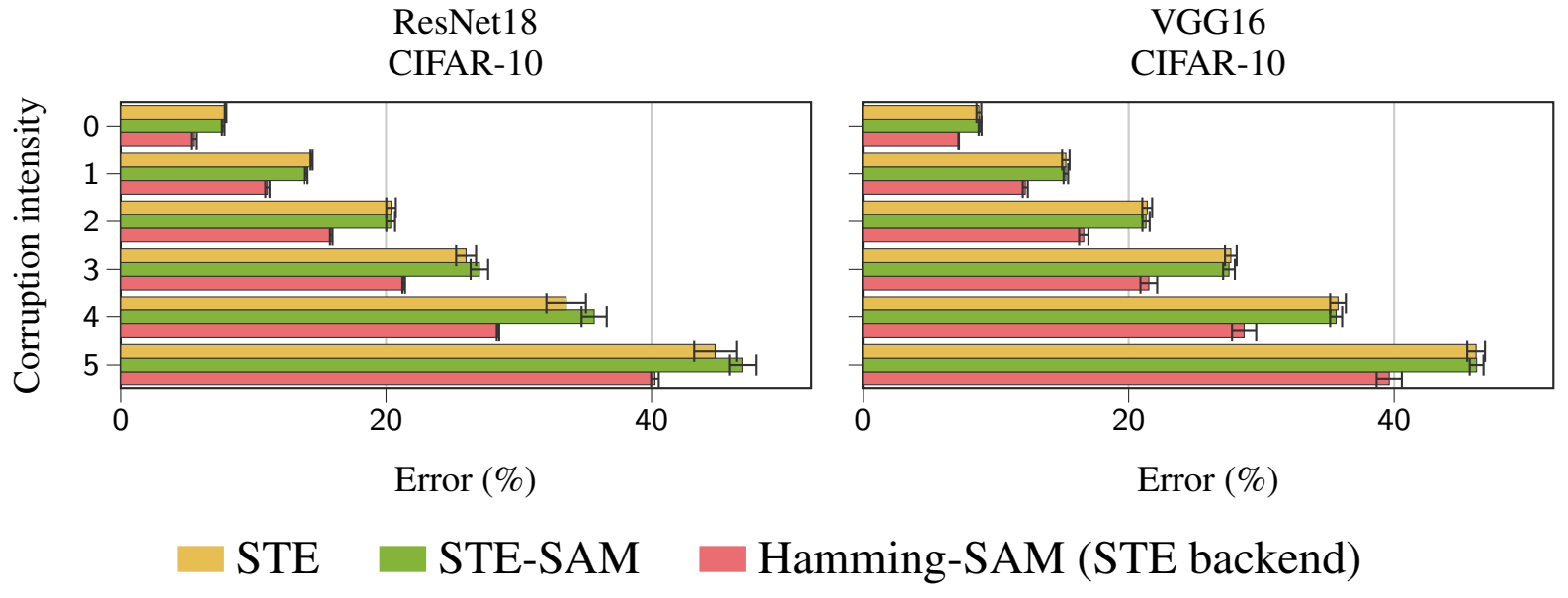}
    \tikzexternalenable
    \caption{\small Clean (level 0) and CIFAR-10-C error at each corruption intensity on the STE backend.  Hamming-SAM uses $k=500$.  Error bars show the standard deviation over seeds.}
    \label{fig:ste_corruption}
\end{figure}

\begin{figure}[H]
    \centering
    \tikzexternaldisable
    \setlength{\figurewidth}{3.6cm}
    \setlength{\figureheight}{2.9cm}
\begin{tikzpicture}

\definecolor{darkslategray38}{RGB}{38,38,38}
\definecolor{lightgray204}{RGB}{204,204,204}
\definecolor{dimgray10784138}{RGB}{234,109,113}
\definecolor{sandybrown23019082}{RGB}{230,190,82}
\definecolor{yellowgreen13318058}{RGB}{133,180,58}

\pgfplotstableread{
k     p  m      s
50    1  94.72  0.02
100   2  94.77  0.04
500   3  94.48  0.18
1000  4  94.21  0.07
}\resnetclean
\pgfplotstableread{
k     p  m      s
50    1  75.31  0.42
100   2  75.82  0.13
500   3  76.60  0.05
1000  4  76.82  0.82
}\resnetmca
\pgfplotstableread{
k     p  m      s
50    1  93.22  0.09
100   2  93.31  0.10
500   3  92.80  0.02
1000  4  92.19  0.08
}\vggclean
\pgfplotstableread{
k     p  m      s
50    1  75.75  0.37
100   2  75.88  0.42
500   3  76.26  0.60
1000  4  75.68  0.14
}\vggmca

\pgfplotsset{
  hsam/.style={dimgray10784138, mark=*, mark size=1.6pt,
    error bars/.cd, y dir=both, y explicit, error bar style={line width=0.6pt}},
  steline/.style={sandybrown23019082, dashed, domain=0.5:4.5, samples=2},
  stesamline/.style={yellowgreen13318058, densely dotted, domain=0.5:4.5, samples=2},
}

\begin{groupplot}[group style={group size=4 by 1, horizontal sep=1.15cm},
    axis line style={darkslategray38},
    height=\figureheight, width=\figurewidth,
    xmin=0.5, xmax=4.5,
    xtick={1,2,3,4}, xticklabels={50,100,500,1000},
    xlabel={$k$},
    tick align=outside, tick pos=left, major tick length=0.6ex,
    xtick style={color=darkslategray38}, ytick style={color=darkslategray38},
    ymajorgrids, y grid style={lightgray204},
    tick label style={font=\scriptsize}, label style={font=\scriptsize},
    title style={font=\scriptsize, align=center, yshift=0.5ex},
    every axis plot/.append style={line width=1pt},
]

\nextgroupplot[title={ResNet18\\Clean Acc (\%)}, ymin=91.6, ymax=95.2]
\addplot[steline] {92.06};
\addplot[stesamline] {92.24};
\addplot[hsam] table[x=p, y=m, y error=s] {\resnetclean};

\nextgroupplot[title={ResNet18\\mCA (\%)}, ymin=70.5, ymax=78]
\addplot[steline] {72.16};
\addplot[stesamline] {71.22};
\addplot[hsam] table[x=p, y=m, y error=s] {\resnetmca};

\nextgroupplot[title={VGG16\\Clean Acc (\%)}, ymin=90.8, ymax=93.7]
\addplot[steline] {91.28};
\addplot[stesamline] {91.19};
\addplot[hsam] table[x=p, y=m, y error=s] {\vggclean};

\nextgroupplot[title={VGG16\\mCA (\%)}, ymin=70.0, ymax=77.2]
\addplot[steline] {70.74};
\addplot[stesamline] {70.80};
\addplot[hsam] table[x=p, y=m, y error=s] {\vggmca};

\end{groupplot}
\end{tikzpicture}
    \tikzexternalenable
    \caption{\small Effect of the Hamming radius $k$ on the STE backend.  Red: Hamming-SAM (mean $\pm$ std over 4 seeds).  Dashed: STE.  Dotted: STE-SAM.}
    \label{fig:ste_k}
\end{figure}
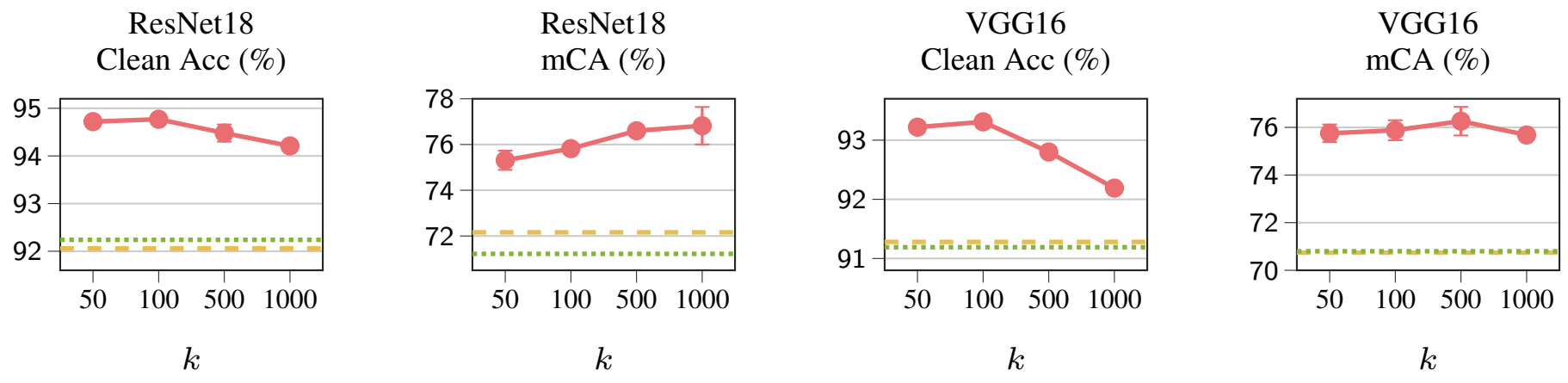

On the STE backend, Hamming-SAM improves clean accuracy over STE by $2.4$ points on ResNet-18 and $1.5$ points on VGG-16, while STE-SAM changes it by less than $0.2$ points.
Mean corruption accuracy improves by $4.4$ and $5.5$ points.
The gains hold for every $k \in \{50, 100, 500, 1000\}$.
Across this range, clean accuracy stays between $94.21$ and $94.77$ on ResNet-18 and between $92.19$ and $93.31$ on VGG-16, and mCA stays between $75.31$ and $76.82$ and between $75.68$ and $76.26$.
The discrete perturbation therefore improves both accuracy and robustness without BOLD, while latent-space SAM barely changes the STE baseline.
\cref{fig:ste_corruption} shows that the gap to STE and STE-SAM is larger at strong corruption levels than on clean data.
\cref{fig:ste_k} shows the effect of $k$.  Clean accuracy is highest at $k=100$ and decreases slightly for larger $k$, while mCA is highest at $k=500$ or $1000$.
This matches the bias--sharpness tradeoff observed on the BOLD backend (\cref{sec:exp_landscape}).

\newpage

\subsection{Visualizing the Loss Landscape}
\label{app:loss_landscape}

\begin{figure}[H]
    \centering
    \begin{subfigure}{0.4\textwidth}
        \includegraphics[width=\textwidth]{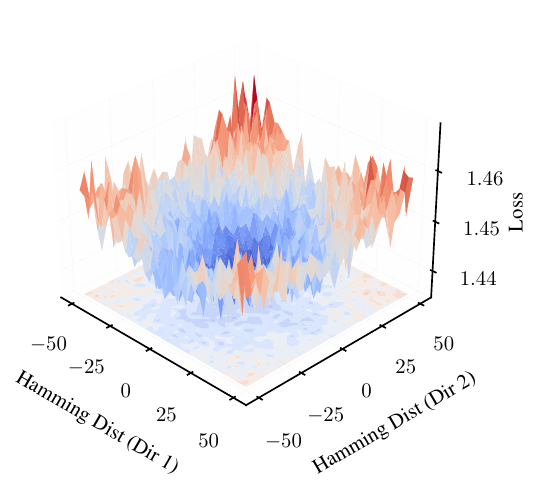}
        \caption{STE}
    \end{subfigure}
    \begin{subfigure}{0.4\textwidth}
        \includegraphics[width=\textwidth]{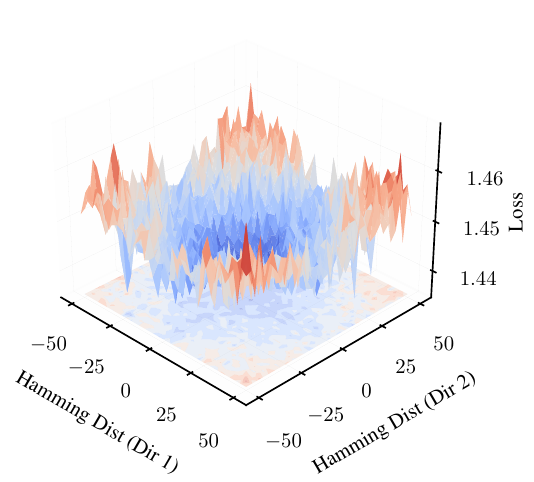}
        \caption{STE-SAM}
    \end{subfigure}
    \begin{subfigure}{0.4\textwidth}
        \includegraphics[width=\textwidth]{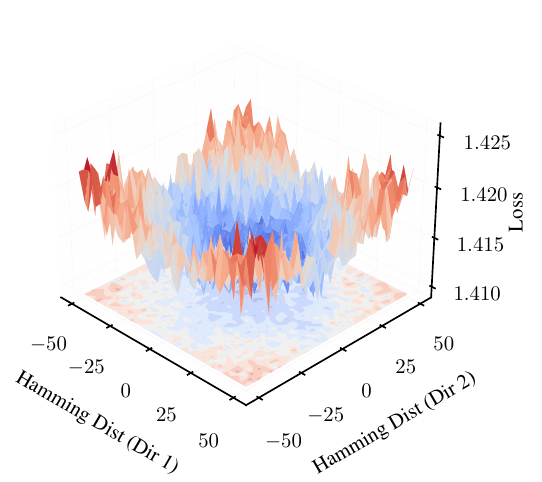}
        \caption{BOLD}
    \end{subfigure}
    \begin{subfigure}{0.4\textwidth}
        \includegraphics[width=\textwidth]{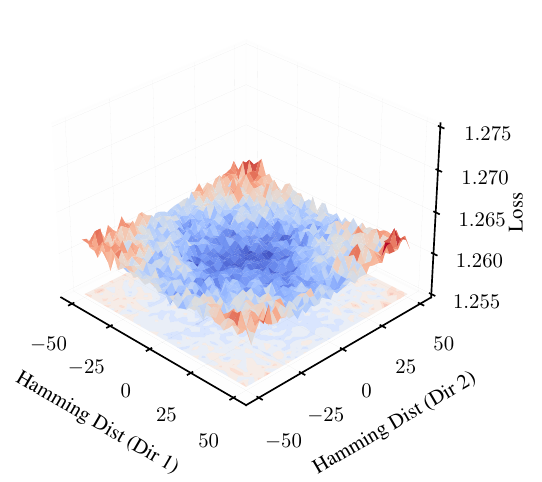}
        \caption{Hamming-SAM}
    \end{subfigure}

    \caption{\small \textbf{Empirical 3D loss landscapes of binary and Boolean MLPs perturbed along two random discrete Hamming directions, evaluated on a test set.}
    \textbf{(a, b, c)} Standard methods, including STE, STE-SAM, and standard BOLD, stay in high-loss local minima with rugged surfaces.
    \textbf{(d)} Hamming-SAM changes the optimization trajectory. By reducing discrete curvature, it leaves the rugged plateau and finds a deeper and wider basin that generalizes better.}
    \label{fig:3d_landscapes}
\end{figure}

\paragraph{Methodology for 2D Hamming subspace projection.}
To visualize the discrete optimization landscape of the MLP (3 hidden layers, 128 units),
we project the high-dimensional loss surface onto a 2D Hamming subspace.
Moving in a discrete parameter space requires flipping discrete bits,
that is, changing the sign of the binary or Boolean weights.

We construct two random, orthogonal perturbation directions.
For the discrete weights of the network, we generate two master masks
with 50\% density, which are subsequently orthogonalized to ensure zero overlap.
Each mask is split into positive and negative sets of indices to allow for
bidirectional perturbation. Let $h_1, h_2 \in [-k_{\max}, k_{\max}]$ represent
the discrete Hamming distance (number of bit flips) taken along these two
orthogonal directions. For our high-resolution visualizations,
we use $k_{\max} = 50$ and scan a $41 \times 41$ grid of evenly spaced
flip counts.

To average out the noise of random bit-flipping,
we apply Monte Carlo smoothing. At each grid point $(h_1, h_2)$,
we draw $N=10$ independent random permutations of the target bits,
flip the corresponding weights $w \leftarrow -w$, evaluate the test loss,
and record the expected (mean) loss $\mathbb{E}[L(h_1, h_2)]$.

\paragraph{Quantitative flatness metric.}
To quantitatively assess the topology of the visualized basins,
we define a scalar flatness score as the Mean Absolute Error (MAE) between
the expected perturbed losses across the grid and the base unperturbed loss $L(0,0)$:
\begin{equation}
    \text{Flatness} = \frac{1}{|G|} \sum_{(h_1, h_2) \in G} \left| \mathbb{E}[L(h_1, h_2)] - L(0,0) \right|
\end{equation}
where $G$ is the set of all $1{,}681$ evaluated discrete grid points.
A lower score indicates a flatter, more stable basin.

\paragraph{Landscape analysis.}
As illustrated in \cref{fig:3d_landscapes},
standard methods, including STE, STE-SAM ($\rho=0.05$), and standard BOLD, stay on non-convex,
rugged plateaus at a high test loss level.
The flatness score confirms this.
STE achieves a flatness score of 0.0044,
STE-SAM scores 0.0045, and BOLD slightly improves to 0.0039.
Note that latent-space SAM does not flatten the discrete landscape at all
here, scoring marginally worse than plain STE, which is the behavior the
vanishing perturbation effect of \cref{sec:pathology} predicts.

In contrast, Hamming-SAM ($k=500$) reduces discrete curvature,
which lets it leave this rugged plateau and reach a deeper,
flatter minimum. The flatness score for Hamming-SAM drops to 0.0011, roughly a
$3.5\times$ reduction relative to standard BOLD and a $4\times$ reduction
relative to STE.
This visualization shows that Hamming-SAM changes the discrete geometry
and finds a basin that supports both optimization and robust generalization.

\section{Empirical Verification of Theoretical Predictions}
\label{app:theory_verification}

We verify the key assumptions and predictions of the theoretical
framework (\cref{thm:ascent,thm:convergence,thm:convergence_sam} and
\cref{prop:multi_bit}) by measuring discrete landscape quantities at
training checkpoints across three architectures: a 3-layer MLP
($n = 442{,}368$), ResNet-18 ($n = 11{,}157{,}504$), and VGG-16
($n = 14{,}708{,}736$), all trained on CIFAR-10 with standard BOLD
and Hamming-SAM.  All quantities are computed on the
training data for consistency, and use exact finite differences rather
than the first-order estimate of \cref{sec:bv_main}.

\subsection{Submodularity Ratio $\gamma_k$ (\cref{thm:ascent})}
\label{app:gamma_k}

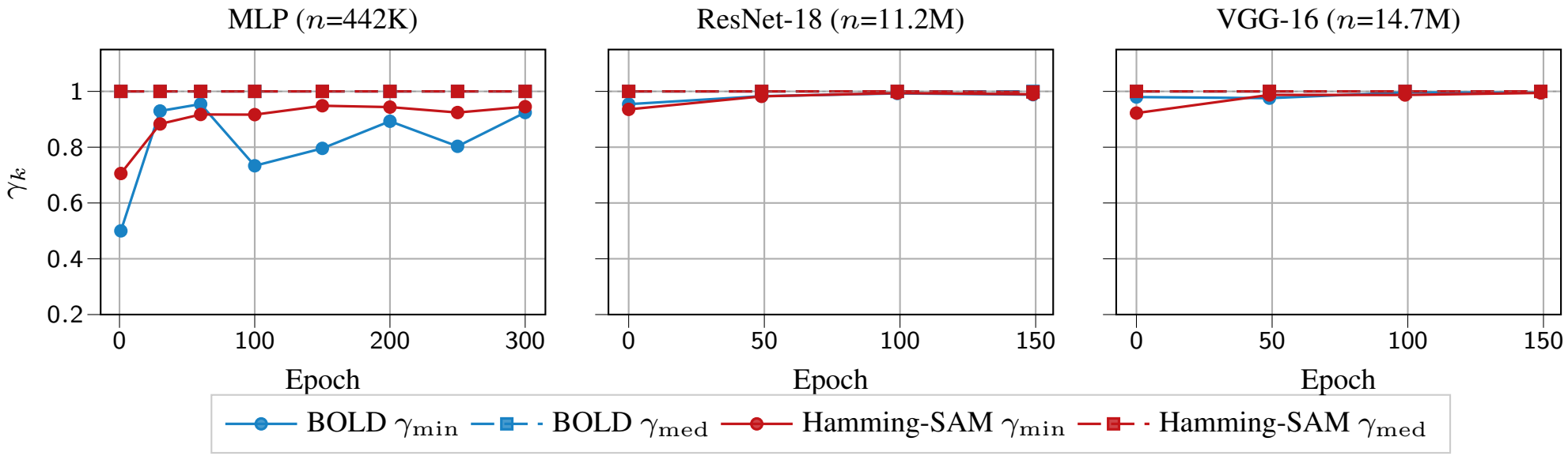
\begin{figure}[H]
    \centering
    \tikzexternaldisable
    \centering
    \scriptsize
    \setlength{\figurewidth}{5.3cm}
    \setlength{\figureheight}{3.8cm}
    \begin{tikzpicture}

\definecolor{darkgray176}{RGB}{176,176,176}
\definecolor{firebrick1952125}{RGB}{195,21,25}
\definecolor{gray}{RGB}{128,128,128}
\definecolor{lightgray204}{RGB}{204,204,204}
\definecolor{steelblue25130196}{RGB}{25,130,196}

\begin{groupplot}[group style={group size=3 by 1, horizontal sep=15pt}]
\nextgroupplot[
height=\figureheight,
major tick length=1ex,
tick align=outside,
tick pos=left,
title={MLP (\(\displaystyle n\)=442K)},
width=\figurewidth,
x grid style={darkgray176},
xlabel={Epoch},
xmajorgrids,
xmin=-13.95, xmax=314.95,
xtick style={color=black},
y grid style={darkgray176},
ylabel={\(\displaystyle \gamma_k\)},
ymajorgrids,
ymin=0.2, ymax=1.15,
ytick style={color=black}
]
\addplot [semithick, steelblue25130196, mark=*, mark size=1.3, mark options={solid}]
table {%
1 0.5
30 0.92972972972973
60 0.954460966542751
100 0.733333333333333
150 0.795698924731183
200 0.892957746478873
250 0.803097345132743
300 0.924240133071223
};
\addplot [semithick, steelblue25130196, dash pattern=on 5.55pt off 2.4pt, mark=square*, mark size=1.3, mark options={solid}]
table {%
1 1
30 1
60 1
100 1
150 1
200 1
250 1
300 1
};
\addplot [semithick, firebrick1952125, mark=*, mark size=1.3, mark options={solid}]
table {%
1 0.705660377358491
30 0.883168316831683
60 0.917197452229299
100 0.916666666666667
150 0.948502994011976
200 0.943624868282403
250 0.924193548387097
300 0.945132743362832
};
\addplot [semithick, firebrick1952125, dash pattern=on 5.55pt off 2.4pt, mark=square*, mark size=1.3, mark options={solid}]
table {%
1 1
30 1
60 1
100 1
150 1
200 1
250 1
300 1
};
\addplot [line width=0.32pt, gray, opacity=0.5, dash pattern=on 0.8pt off 1.32pt]
table {%
-13.95 1
314.95 1
};

\nextgroupplot[
height=\figureheight,
legend cell align={left},
legend columns=4,
legend style={
  fill opacity=0.8,
  draw opacity=1,
  text opacity=1,
  at={(0.5,-0.3)},
  anchor=north,
  draw=lightgray204
},
major tick length=1ex,
scaled y ticks=manual:{}{\pgfmathparse{#1}},
tick align=outside,
tick pos=left,
title={ResNet-18 (\(\displaystyle n\)=11.2M)},
width=\figurewidth,
x grid style={darkgray176},
xlabel={Epoch},
xmajorgrids,
xmin=-7.45, xmax=156.45,
xtick style={color=black},
y grid style={darkgray176},
ymajorgrids,
ymin=0.2, ymax=1.15,
ytick style={color=black},
yticklabels={}
]
\addplot [semithick, steelblue25130196, mark=*, mark size=1.3, mark options={solid}]
table {%
0 0.954198473282443
49 0.982882041705571
99 0.992665672455666
149 0.988151640308416
};
\addlegendentry{BOLD $\gamma_{\min}$}
\addplot [semithick, steelblue25130196, dash pattern=on 5.55pt off 2.4pt, mark=square*, mark size=1.3, mark options={solid}]
table {%
0 1
49 1
99 1
149 1
};
\addlegendentry{BOLD $\gamma_{\mathrm{med}}$}
\addplot [semithick, firebrick1952125, mark=*, mark size=1.3, mark options={solid}]
table {%
0 0.935483870967742
49 0.982162162162162
99 0.993962968233638
149 0.989626967757483
};
\addlegendentry{Hamming-SAM $\gamma_{\min}$}
\addplot [semithick, firebrick1952125, dash pattern=on 5.55pt off 2.4pt, mark=square*, mark size=1.3, mark options={solid}]
table {%
0 1
49 1
99 1.00000555421329
149 0.997829362049441
};
\addlegendentry{Hamming-SAM $\gamma_{\mathrm{med}}$}
\addplot [line width=0.32pt, gray, opacity=0.5, dash pattern=on 0.8pt off 1.32pt, forget plot]
table {%
-7.44999999999993 1
156.45 1
};

\nextgroupplot[
height=\figureheight,
major tick length=1ex,
scaled y ticks=manual:{}{\pgfmathparse{#1}},
tick align=outside,
tick pos=left,
title={VGG-16 (\(\displaystyle n\)=14.7M)},
width=\figurewidth,
x grid style={darkgray176},
xlabel={Epoch},
xmajorgrids,
xmin=-7.45, xmax=156.45,
xtick style={color=black},
y grid style={darkgray176},
ymajorgrids,
ymin=0.2, ymax=1.15,
ytick style={color=black},
yticklabels={}
]
\addplot [semithick, steelblue25130196, mark=*, mark size=1.3, mark options={solid}]
table {%
0 0.979695431472081
49 0.975712077721351
99 0.996422683875807
149 0.995145009099664
};
\addplot [semithick, steelblue25130196, dash pattern=on 5.55pt off 2.4pt, mark=square*, mark size=1.3, mark options={solid}]
table {%
0 1
49 1
99 0.999927556854933
149 1
};
\addplot [semithick, firebrick1952125, mark=*, mark size=1.3, mark options={solid}]
table {%
0 0.922222222222222
49 0.987339187257505
99 0.987117046347211
149 0.994511373119637
};
\addplot [semithick, firebrick1952125, dash pattern=on 5.55pt off 2.4pt, mark=square*, mark size=1.3, mark options={solid}]
table {%
0 1
49 1
99 1
149 1
};
\addplot [line width=0.32pt, gray, opacity=0.5, dash pattern=on 0.8pt off 1.32pt]
table {%
-7.45000000000005 1
156.45 1
};
\end{groupplot}

\end{tikzpicture}
    \tikzexternalenable

    \definecolor{brown1926161}{RGB}{193,21,27}
    \definecolor{darkcyan22112159}{HTML}{1982C4}

    \caption{\small{Submodularity ratio $\gamma_k$ over training.
    Solid lines: $\gamma_{\min}$. Dashed lines: $\gamma_{\mathrm{med}}$.}}
  \label{fig:gamma_k}

\end{figure}

\cref{tab:gamma_k} reports the submodularity ratio $\gamma_k$,
estimated by sampling 100 random $(S, L)$ pairs per checkpoint
(with $|S| \leq k$ and $|L| \leq 3$).
A visual illustration of $\gamma_k$ evolution is shown in \cref{fig:gamma_k}.

The median $\gamma_k$ is $1.000$ at every checkpoint but one
(ResNet-18 with Hamming-SAM at epoch~150, where it is $0.998$),
meaning the loss landscape is
exactly submodular in the majority of sampled configurations and
the greedy ascent achieves the full $(1{-}1/e)$ approximation ratio
there.  The minimum $\gamma_k$ never falls below $0.50$,
and at convergence it is at least $0.84$ on the MLP and at least
$0.98$ on the convolutional architectures.
The lowest value ($\gamma_{\min} = 0.50$) occurs for standard
BOLD on the MLP at epoch~1 (random initialization).  By epoch~300,
$\gamma_{\min}$ rises to $0.92$, indicating the landscape
becomes more submodular during training.
The same upward trend holds on the convolutional architectures (e.g.,
VGG with Hamming-SAM: $0.92 \to 0.99$), consistent with the landscape becoming more
submodular as the optimizer converges to a flat basin.
What the guarantee of \cref{thm:ascent} requires is only that
$\gamma_k$ be bounded away from zero, which holds with a wide margin
throughout.

\begin{table}[h]
  \caption{Submodularity ratio $\gamma_k$ at initialization and
    convergence.  Estimated from 100 random $(S,L)$ samples per
    checkpoint.}
  \label{tab:gamma_k}
  \centering\small
  \begin{tabular}{llrrrr}
    \toprule
    \textbf{Architecture} & \textbf{Method} & \textbf{Epoch}
      & $\gamma_{\min}$ & $\gamma_{\mathrm{med}}$
      & $\gamma_{\mathrm{mean}}$ \\
    \midrule
    \multirow{4}{*}{MLP ($n{=}442$K)}
      & BOLD        &   1 & 0.500 & 1.000 & 1.001 \\
      & BOLD        & 300 & 0.924 & 1.000 & 1.000 \\
      & Hamming-SAM &   1 & 0.906 & 1.000 & 1.002 \\
      & Hamming-SAM & 300 & 0.845 & 1.000 & 0.998 \\
    \midrule
    \multirow{4}{*}{ResNet-18 ($n{=}11.2$M)}
      & BOLD        &   0 & 0.954 & 1.000 & 1.004 \\
      & BOLD        & 150 & 0.988 & 1.000 & 1.002 \\
      & Hamming-SAM &   0 & 0.935 & 1.000 & 1.002 \\
      & Hamming-SAM & 150 & 0.990 & 0.998 & 0.997 \\
    \midrule
    \multirow{4}{*}{VGG-16 ($n{=}14.7$M)}
      & BOLD        &   0 & 0.980 & 1.000 & 1.002 \\
      & BOLD        & 150 & 0.995 & 1.000 & 1.000 \\
      & Hamming-SAM &   0 & 0.922 & 1.000 & 0.997 \\
      & Hamming-SAM & 150 & 0.995 & 1.000 & 1.001 \\
    \bottomrule
  \end{tabular}
\end{table}

\subsection{Discrete Hessian Spectrum (\cref{thm:convergence_sam})}
\label{app:hessian_verification}

\cref{tab:hessian_evolution} reports the evolution of the discrete
Hessian $H_{ij}$ during training, estimated by sampling 300 random
$(i,j)$ pairs per checkpoint.  Each entry is an exact finite difference of the
cross-entropy on a fixed subset of $1{,}024$ training images, computed in
single precision, so entries below about $10^{-8}$ are indistinguishable from
zero.

\paragraph{Interaction scale.}
At convergence, $kh$ ranges from $1.3 \times 10^{-4}$ (ResNet-18 with
Hamming-SAM) to $2.3 \times 10^{-2}$ (MLP with BOLD), which we use as a
measure of the sampled interaction scale.  The condition $\eta > kh$ of
\cref{thm:convergence_sam} refers to a fixed threshold on the exact Boolean
Variation, whereas the accumulator thresholds the accumulated backward signal
with an adaptive threshold (\cref{app:accumulator_threshold}), so the two are
not directly comparable.  \cref{thm:convergence_sam} involves only the pairs
formed by the descent bit and the ascent set.  \cref{app:sequential_greedy} shows
that interactions among the selected coordinates are of the same order as among
random pairs, so the random-pair statistic is a reasonable proxy for the
one entering the theorem.

\paragraph{Hamming-SAM converges to low-$h$ regions.}
On ResNet-18, $h_{\max}$ drops by about two orders of magnitude, from
$1.2 \times 10^{-6}$ (epoch~0) to $1.2 \times 10^{-8}$ (epoch~150)
under Hamming-SAM, close to the resolution of the loss, with the fraction of
entries below $10^{-8}$ rising from 11\% to 99\%.  VGG-16 shows a similar
pattern: $h_{\max}$ drops by $12\times$ from $1.7 \times 10^{-6}$ to
$1.4 \times 10^{-7}$, with this fraction rising from 19\% to 93\%.  On both
architectures, the fraction for same-layer pairs reaches 88--100\% at
convergence, confirming the structural attenuation discussed in
\cref{rem:eta_kh}.

Standard BOLD also shows decreasing $h_{\max}$, but the reduction
is smaller and $h_{\max}$ remains an order of magnitude larger
than that of Hamming-SAM at convergence on ResNet-18.

\begin{table}[h]
  \caption{Discrete Hessian evolution during training.
    $h_{\max} = \max|H_{ij}|$ over 300 sampled pairs,
    $\%_0$ = fraction with $|H_{ij}| \leq 10^{-8}$, and
    $kh = k \cdot h_{\max}$.}
  \label{tab:hessian_evolution}
  \centering\small
  \begin{tabular}{llrrrrr}
    \toprule
    \textbf{Arch.} & \textbf{Method} & \textbf{Epoch}
      & $h_{\max}$ & $\bar{h}$ & $\%_0$ & $kh$ \\
    \midrule
    \multirow{4}{*}{MLP}
      & BOLD        &   1 & $1.8 \times 10^{-6}$ & $2.5 \times 10^{-7}$ &  8\% & $7.9 \times 10^{-4}$ \\
      & BOLD        & 300 & $5.1 \times 10^{-5}$ & $2.3 \times 10^{-6}$ &  3\% & $2.3 \times 10^{-2}$ \\
      & Hamming-SAM &   1 & $7.2 \times 10^{-7}$ & $1.6 \times 10^{-7}$ & 15\% & $3.2 \times 10^{-4}$ \\
      & Hamming-SAM & 300 & $3.1 \times 10^{-6}$ & $6.2 \times 10^{-7}$ &  7\% & $1.4 \times 10^{-3}$ \\
    \midrule
    \multirow{4}{*}{ResNet-18}
      & BOLD        &   0 & $5.1 \times 10^{-7}$ & $1.0 \times 10^{-7}$ & 11\% & $5.6 \times 10^{-3}$ \\
      & BOLD        & 150 & $2.5 \times 10^{-7}$ & $1.1 \times 10^{-8}$ & 76\% & $2.8 \times 10^{-3}$ \\
      & Hamming-SAM &   0 & $1.2 \times 10^{-6}$ & $9.8 \times 10^{-8}$ & 11\% & $1.4 \times 10^{-2}$ \\
      & Hamming-SAM & 150 & $1.2 \times 10^{-8}$ & $1.5 \times 10^{-9}$ & 99\% & $1.3 \times 10^{-4}$ \\
    \midrule
    \multirow{4}{*}{VGG-16}
      & BOLD        &   0 & $1.2 \times 10^{-6}$ & $7.6 \times 10^{-8}$ & 16\% & $1.8 \times 10^{-2}$ \\
      & BOLD        & 150 & $3.4 \times 10^{-7}$ & $5.5 \times 10^{-9}$ & 93\% & $5.0 \times 10^{-3}$ \\
      & Hamming-SAM &   0 & $1.7 \times 10^{-6}$ & $7.5 \times 10^{-8}$ & 19\% & $2.4 \times 10^{-2}$ \\
      & Hamming-SAM & 150 & $1.4 \times 10^{-7}$ & $3.3 \times 10^{-9}$ & 93\% & $2.1 \times 10^{-3}$ \\
    \bottomrule
  \end{tabular}
\end{table}

\subsection{Multi-Bit Descent (\cref{prop:multi_bit})}
\label{app:multibit_verification}

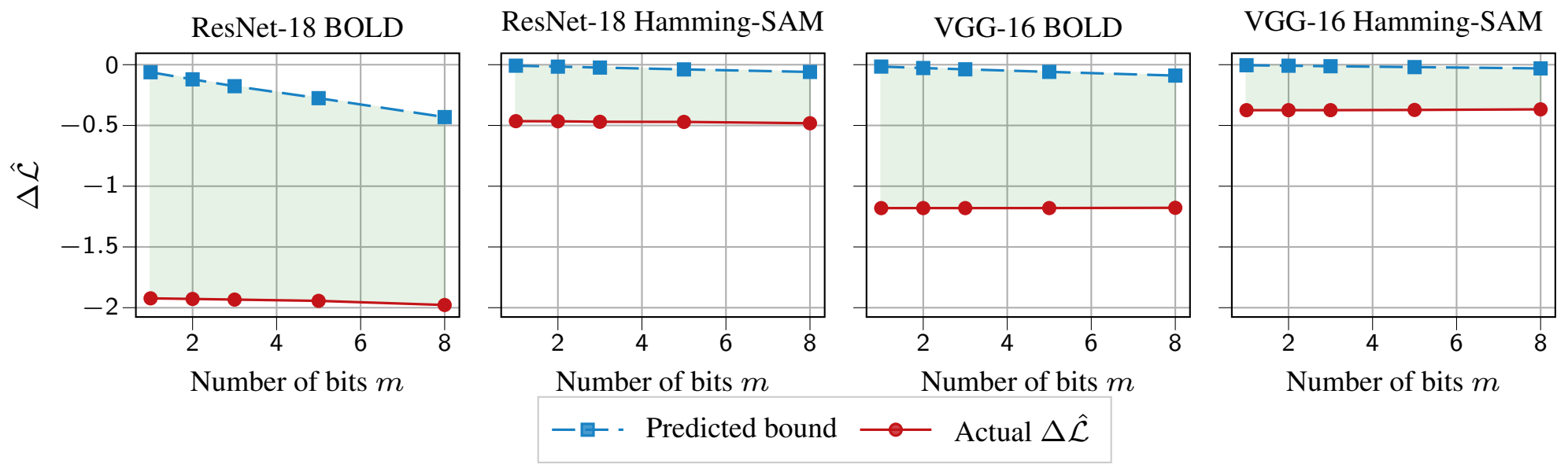
\begin{figure}[H]
    \centering
    \tikzexternaldisable
    \centering
    \scriptsize
    \setlength{\figurewidth}{4.3cm}
    \setlength{\figureheight}{3.8cm}
    \begin{tikzpicture}

\definecolor{darkgray176}{RGB}{176,176,176}
\definecolor{firebrick1952125}{RGB}{195,21,25}
\definecolor{green}{RGB}{0,128,0}
\definecolor{lightgray204}{RGB}{204,204,204}
\definecolor{steelblue25130196}{RGB}{25,130,196}

\begin{groupplot}[group style={group size=4 by 1, horizontal sep=10pt}]
\nextgroupplot[
height=\figureheight,
legend style={at={(1.0,-0.3)}, anchor=north, legend columns=2, column sep=1ex},
major tick length=1ex,
tick align=outside,
tick pos=left,
title={ResNet-18 BOLD},
width=\figurewidth,
x grid style={darkgray176},
xlabel={Number of bits \(\displaystyle m\)},
xmajorgrids,
xmin=0.65, xmax=8.35,
xtick style={color=black},
y grid style={darkgray176},
ylabel={\(\displaystyle \Delta\hat{\mathcal{L}}\)},
ymajorgrids,
ymin=-2.07649636759888, ymax=0.0942586201941595,
ytick style={color=black}
]
\path [draw=green, fill=green, opacity=0.1]
(axis cs:1,-1.9229823499918)
--(axis cs:1,-0.0621104761958122)
--(axis cs:2,-0.120582587134777)
--(axis cs:3,-0.176910747921283)
--(axis cs:5,-0.275491914271697)
--(axis cs:8,-0.43115139517613)
--(axis cs:8,-1.97782568633556)
--(axis cs:8,-1.97782568633556)
--(axis cs:5,-1.94374248385429)
--(axis cs:3,-1.93322815001011)
--(axis cs:2,-1.92798508703709)
--(axis cs:1,-1.9229823499918)
--cycle;

\addplot [semithick, steelblue25130196, dash pattern=on 5.55pt off 2.4pt, mark=square*, mark size=1.4, mark options={solid}]
table {%
1 -0.0621104761958122
2 -0.120582587134777
3 -0.176910747921283
5 -0.275491914271697
8 -0.43115139517613
};
\addplot [semithick, firebrick1952125, mark=*, mark size=1.4, mark options={solid}]
table {%
1 -1.9229823499918
2 -1.92798508703709
3 -1.93322815001011
5 -1.94374248385429
8 -1.97782568633556
};

\nextgroupplot[
height=\figureheight,
legend cell align={left},
legend columns=2,
legend style={
  fill opacity=0.8,
  draw opacity=1,
  text opacity=1,
  at={(1,-0.3)},
  anchor=north,
  draw=lightgray204
},
legend style={at={(1.0,-0.3)}, anchor=north, legend columns=2, column sep=1ex},
major tick length=1ex,
scaled y ticks=manual:{}{\pgfmathparse{#1}},
tick align=outside,
tick pos=left,
title={ResNet-18 Hamming-SAM},
width=\figurewidth,
x grid style={darkgray176},
xlabel={Number of bits \(\displaystyle m\)},
xmajorgrids,
xmin=0.65, xmax=8.35,
xtick style={color=black},
y grid style={darkgray176},
ymajorgrids,
ymin=-2.07649636759888, ymax=0.0942586201941595,
ytick style={color=black},
yticklabels={}
]
\path [draw=green, fill=green, opacity=0.1]
(axis cs:1,-0.464837715029716)
--(axis cs:1,-0.0082162618637085)
--(axis cs:2,-0.0159670874418225)
--(axis cs:3,-0.0237114809278864)
--(axis cs:5,-0.0387788921943866)
--(axis cs:8,-0.0596120996633545)
--(axis cs:8,-0.48190850764513)
--(axis cs:8,-0.48190850764513)
--(axis cs:5,-0.471046363934875)
--(axis cs:3,-0.469966189935803)
--(axis cs:2,-0.465662237256765)
--(axis cs:1,-0.464837715029716)
--cycle;

\addplot [semithick, steelblue25130196, dash pattern=on 5.55pt off 2.4pt, mark=square*, mark size=1.4, mark options={solid}]
table {%
1 -0.0082162618637085
2 -0.0159670874418225
3 -0.0237114809278864
5 -0.0387788921943866
8 -0.0596120996633545
};
\addlegendentry{Predicted bound}
\addplot [semithick, firebrick1952125, mark=*, mark size=1.4, mark options={solid}]
table {%
1 -0.464837715029716
2 -0.465662237256765
3 -0.469966189935803
5 -0.471046363934875
8 -0.48190850764513
};
\addlegendentry{Actual $\Delta\hat{\mathcal{L}}$}

\nextgroupplot[
height=\figureheight,
legend style={at={(1.0,-0.3)}, anchor=north, legend columns=2, column sep=1ex},
major tick length=1ex,
scaled y ticks=manual:{}{\pgfmathparse{#1}},
tick align=outside,
tick pos=left,
title={VGG-16 BOLD},
width=\figurewidth,
x grid style={darkgray176},
xlabel={Number of bits \(\displaystyle m\)},
xmajorgrids,
xmin=0.65, xmax=8.35,
xtick style={color=black},
y grid style={darkgray176},
ymajorgrids,
ymin=-2.07649636759888, ymax=0.0942586201941595,
ytick style={color=black},
yticklabels={}
]
\path [draw=green, fill=green, opacity=0.1]
(axis cs:1,-1.18038493394852)
--(axis cs:1,-0.0154726170003414)
--(axis cs:2,-0.027466216506582)
--(axis cs:3,-0.03806308678395)
--(axis cs:5,-0.0590092866696068)
--(axis cs:8,-0.0896390150155639)
--(axis cs:8,-1.17784810066223)
--(axis cs:8,-1.17784810066223)
--(axis cs:5,-1.17998361587524)
--(axis cs:3,-1.18020805716515)
--(axis cs:2,-1.18009395897388)
--(axis cs:1,-1.18038493394852)
--cycle;

\addplot [semithick, steelblue25130196, dash pattern=on 5.55pt off 2.4pt, mark=square*, mark size=1.4, mark options={solid}]
table {%
1 -0.0154726170003414
2 -0.027466216506582
3 -0.03806308678395
5 -0.0590092866696068
8 -0.0896390150155639
};
\addplot [semithick, firebrick1952125, mark=*, mark size=1.4, mark options={solid}]
table {%
1 -1.18038493394852
2 -1.18009395897388
3 -1.18020805716515
5 -1.17998361587524
8 -1.17784810066223
};

\nextgroupplot[
height=\figureheight,
legend style={at={(1.0,-0.3)}, anchor=north, legend columns=2, column sep=1ex},
major tick length=1ex,
scaled y ticks=manual:{}{\pgfmathparse{#1}},
tick align=outside,
tick pos=left,
title={VGG-16 Hamming-SAM},
width=\figurewidth,
x grid style={darkgray176},
xlabel={Number of bits \(\displaystyle m\)},
xmajorgrids,
xmin=0.65, xmax=8.35,
xtick style={color=black},
y grid style={darkgray176},
ymajorgrids,
ymin=-2.07649636759888, ymax=0.0942586201941595,
ytick style={color=black},
yticklabels={}
]
\path [draw=green, fill=green, opacity=0.1]
(axis cs:1,-0.374411012977362)
--(axis cs:1,-0.0044120610691607)
--(axis cs:2,-0.00846578444179613)
--(axis cs:3,-0.0126428431685781)
--(axis cs:5,-0.0197772805404384)
--(axis cs:8,-0.0305415011825971)
--(axis cs:8,-0.367522850632668)
--(axis cs:8,-0.367522850632668)
--(axis cs:5,-0.37248745188117)
--(axis cs:3,-0.374179400503635)
--(axis cs:2,-0.37424137070775)
--(axis cs:1,-0.374411012977362)
--cycle;

\addplot [semithick, steelblue25130196, dash pattern=on 5.55pt off 2.4pt, mark=square*, mark size=1.4, mark options={solid}]
table {%
1 -0.0044120610691607
2 -0.00846578444179613
3 -0.0126428431685781
5 -0.0197772805404384
8 -0.0305415011825971
};
\addplot [semithick, firebrick1952125, mark=*, mark size=1.4, mark options={solid}]
table {%
1 -0.374411012977362
2 -0.37424137070775
3 -0.374179400503635
5 -0.37248745188117
8 -0.367522850632668
};
\end{groupplot}

\end{tikzpicture}
    \tikzexternalenable

    \definecolor{brown1926161}{RGB}{193,21,27}
    \definecolor{darkcyan22112159}{HTML}{1982C4}

    \caption{\small{Multi-bit descent verification at convergence.
    Dashed blue: predicted upper bound $-m\delta + \binom{m}{2}h$.
    Solid red: actual loss change.  The bound holds whenever
    actual $\leq$ predicted (ResNet-18 and VGG-16: 40/40).}}
  \label{fig:multibit}

\end{figure}

\cref{prop:multi_bit} predicts that simultaneously flipping $m$ bits,
each with $\mathrm{BV}_{i_l}(w^t) < -\delta$, decreases the loss
provided $\delta > (m{-}1)h/2$.  \cref{fig:multibit} reports the
predicted bound $-m\delta + \binom{m}{2}h$ versus the actual loss
change for $m \in \{1, 2, 3, 5, 8\}$ at the final checkpoint.

On ResNet-18 and VGG-16, the bound holds for all
$m$ values at all checkpoints: 40/40 tests across both
architectures.  The actual loss decrease is consistently much larger
than the predicted bound (e.g., for VGG with Hamming-SAM at epoch~150, predicted
$\leq -0.031$ versus actual $= -0.368$ for $m = 8$).  The bound is therefore
conservative but valid.

\subsection{Sequential vs.\ Parallel Greedy Ascent Quality}
\label{app:sequential_greedy}

\begin{table}[h]
    \caption{Adversarial loss increase: top-$k$ versus sequential
    greedy (Hamming-SAM only, $k = \lfloor 10^{-3} n \rfloor$).  Ratio is
    $f_{\mathrm{seq}} / f_{\mathrm{top\text{-}k}}$,
    $h_{\mathrm{eff}} = (f_{\mathrm{seq}} - f_{\mathrm{top\text{-}k}})/(k(k{-}1))$ is the
    pairwise interaction that would account for the whole deficit (--: no
    deficit, as top-$k$ exceeds the sequential greedy), and
    $h_{\max}$ is the random-pair estimate of \cref{tab:hessian_evolution}.}
  \label{tab:seq_greedy}
  \centering\small
  \begin{tabular}{lrrrrrr}
    \toprule
    \textbf{Architecture} & \textbf{Epoch}
      & $f_{\mathrm{top\text{-}k}}$ & $f_{\mathrm{seq}}$
      & \textbf{Ratio} & $h_{\mathrm{eff}}$ & $h_{\max}$ \\
    \midrule
    \multirow{2}{*}{MLP ($k{=}442$)}
      &   1 & 0.12 & 0.36 & 2.8$\times$ & $1.2 \times 10^{-6}$ & $7.2 \times 10^{-7}$ \\
      & 300 & 0.20 & 0.74 & 3.7$\times$ & $2.8 \times 10^{-6}$ & $3.1 \times 10^{-6}$ \\
    \midrule
    \multirow{2}{*}{ResNet-18 ($k{=}11{,}157$)}
      &   0 & 5.38 & 6.20 & 1.2$\times$ & $6.6 \times 10^{-9}$ & $1.2 \times 10^{-6}$ \\
      & 150 & 6.35 & 8.90 & 1.4$\times$ & $2.0 \times 10^{-8}$ & $1.2 \times 10^{-8}$ \\
    \midrule
    \multirow{2}{*}{VGG-16 ($k{=}14{,}708$)}
      &   0 & 12.78 & 8.31 & 0.7$\times$ & -- & $1.7 \times 10^{-6}$ \\
      & 150 & 2.37  & 5.21 & 2.2$\times$ & $1.3 \times 10^{-8}$ & $1.4 \times 10^{-7}$ \\
    \bottomrule
  \end{tabular}
\end{table}

\cref{thm:ascent} analyzes the sequential greedy, while
\cref{alg:hamming_sam} uses a top-$k$ variant for
efficiency, and \cref{thm:ascent_topk} bounds the resulting deficit.
\cref{tab:seq_greedy} compares the adversarial loss
increase achieved by both.  The sequential greedy finds a stronger
adversarial point at every checkpoint except VGG-16 at initialization,
where the one-shot selection happens to pick a better set, which is
possible because neither rule is optimal.  This is consistent with
\cref{thm:ascent_topk}, which attributes the deficit of the one-shot rule to
interactions among the selected coordinates.
The top-$k$ variant still
identifies meaningful adversarial configurations in all
cases ($f_{\mathrm{top\text{-}k}} > 0$ everywhere), retaining
$71\%$ of the sequential gain on ResNet-18 and $45\%$ on VGG-16 at
convergence.  The deficit is smallest on ResNet-18
($1.2$--$1.4\times$) and largest on the MLP ($2.8$--$3.7\times$).

\paragraph{Relation to the top-$k$ bound.}
\cref{thm:ascent_topk} bounds the deficit by the interactions among the
selected coordinates.  These may differ from the random-pair statistics of
\cref{tab:hessian_evolution}, since the ascent selects the $k$ largest of $n$
Boolean Variations.  \cref{tab:seq_greedy} compares the two through the
effective interaction $h_{\mathrm{eff}}$.  This is the pairwise interaction that
would account for the whole deficit.  In every case, $h_{\mathrm{eff}}$ is below
the sampled $h_{\max}$ or within a factor of two of it.  Interactions among the
selected coordinates are therefore of the same order as among random pairs.
At convergence, the bound instantiated with the sampled $h_{\max}$ covers the
measured deficit on the MLP ($0.60$ against $0.54$) and on VGG-16 ($30$ against
$2.8$).  On ResNet-18, it accounts for most of the deficit ($1.4$ against
$2.55$).  \cref{thm:ascent_topk} attributes the remainder to interactions of
order three and higher.

\section{Limitations and Future Work}
\label{app:limitations}

Our evaluation covers standard benchmarks used in the SAM
literature (CIFAR-10/100, ImageNet) across diverse architectures
(MLP, VGG, ResNet, ViT), complemented by extensive theoretical
verification of the discrete landscape quantities that underpin
the framework.  We record below the assumptions that the analysis
leaves open, followed by two natural extensions.

\paragraph{Beyond binary weights.}
The theoretical framework, comprising Hamming-ball perturbations, the discrete
Hessian and the submodularity ratio, extends naturally to ternary
$\{-1, 0, +1\}$ or multi-bit quantized weights, where the
parameter space remains a discrete lattice.  Developing Hamming-SAM
variants for these settings could broaden the practical impact to
mixed-precision and low-bit quantization regimes.

\paragraph{Large-scale language models.}
Recent work has demonstrated the feasibility of Boolean and
low-bit large language models~\citep{wang2025bitnet, tran2026highly}.
Evaluating Hamming-SAM at this scale would test
whether the sharpness-robustness connection observed in our
experiments extends to billion-parameter architectures.  This is beyond our current
computational resources but is a promising direction.

\section{Experimental Details}
\label{app:exp_details}

\paragraph{Architectures.}
We follow standard practices in the BNN and BOLD
literature~\citep{hubara2016binarized, nguyen2024bold}.  For all
binary methods, full-precision linear and convolutional layers are
replaced with their Boolean (BOLD) or binary (STE) counterparts
without modifying the network architecture or introducing
additional learnable parameters.  We evaluate ResNet-18 and
ResNet-50~\citep{he2016resnet}, VGG-16~\citep{simonyan2015vgg},
ViT-Tiny~\citep{dosovitskiy2021vit}, and a 3-layer MLP with 128
hidden units per layer.

\paragraph{Training protocol.}
All experiments were implemented in Python using PyTorch \citep{paszke2019}.
For CIFAR-10/100~\citep{krizhevsky2009cifar}, models were trained
for 150 epochs with a minibatch size of 128.  For
ImageNet~\citep{deng2009imagenet}, models were trained for 90
epochs with a minibatch size of 512.  All runs use a cosine
learning rate scheduler.

\paragraph{Optimizer settings.}
For STE-based methods (STE, STE-SAM), we use SGD with a single
momentum term for fair comparison with the BOLD accumulator,
which also uses single momentum.  The learning rate was selected
from $\{0.01, 0.05, 0.1\}$.  We also tried Adam and found
similar results.  BOLD and Hamming-SAM use the BOLD
accumulator~\citep{nguyen2024bold} with the learning rate
selected from $\{10, 30, 50\}$.

\paragraph{Hyperparameter selection.}
For Hamming-SAM, the Hamming radius $k$ was selected from
$\{50, 100, 500, 1000\}$ on a held-out validation set.  For
STE-SAM, the perturbation radius $\rho$ was selected from
$\{0.005, 0.01, 0.05, 0.1, 0.5\}$ using the same protocol.
All other hyperparameters (learning rate, batch size, number of
epochs) were kept identical across methods for fair comparison.

\paragraph{Evaluation.}
Clean accuracy is reported on the standard test split.
Corruption robustness is evaluated on
CIFAR-10-C/100-C~\citep{hendrycks2019robustness} (19 corruption
types $\times$ 5 severity levels) and
ImageNet-C~\citep{hendrycks2019robustness}.  We report mean
corruption accuracy (mCA) averaged over all types and severities.
OOD detection on ImageNet uses AUROC.  All results are averaged
over 4 seeds with standard deviations reported.

\paragraph{Computational cost.}
\label{app:compute_cost}
Each Hamming-SAM step uses two backward passes, one at $w^t$ and one at
$\hat{w}$, as in continuous SAM.  On a single V100, a Hamming-SAM step
takes about $1.9\times$ the time of a standard BOLD step, both for VGG-16 on
CIFAR-10 and for ResNet-50 on ImageNet.  The top-$k$ selection uses PyTorch's
\texttt{topk} and adds less than $1\%$ to the step time.  The only extra memory
is the list of at most $k$ flipped indices, which takes a few kilobytes.

\newpage
\section{Illustrative Example}
\label{app:code_sample}

\cref{alg:code_hamming_sam} provides an illustraive example of the Hamming-SAM optimizer in Python using PyTorch \citep{paszke2019}.
It reads a per-bit signal from \texttt{p.grad}, so any backend can be used.  For STE (\cref{app:ste_backend}), pass the latent weights and a standard optimizer as \texttt{descent}.

\begin{algorithm}[H]
    \centering
    \caption{Illustrative Python code of Hamming-SAM optimizer}\label{alg:code_hamming_sam}
\begin{lstlisting}[language=Python]
import torch


class HammingSAM(torch.optim.Optimizer):
    """Hamming-SAM on {-1, +1}^n. p.grad holds any per-bit signal f'_i (BOLD
    backward pass or STE gradient), so that BV_i ~ -sign(w_i) * f'_i.

    Args:
        params: Boolean weights in {-1, +1} (BOLD) or latent weights (STE).
        lr: Learning rate of the default descent step (BOLD accumulator).
        k: Hamming radius.
        descent: Callable descent(group, i), or a torch optimizer on the latent
            weights (STE), e.g. descent=torch.optim.SGD(w, lr=0.05).
    """
    def __init__(self, params, lr, k=500, descent=None):
        super().__init__(params, dict(lr=lr))
        self.k = k
        self.descent = descent or self._accumulator_update
        for g in self.param_groups:
            g['accums'] = [torch.zeros_like(p) for p in g['params']]
            g['ratios'] = [1.0 for _ in g['params']]
            g['adv_flips'] = []

    @torch.no_grad()
    def first_step(self, zero_grad=False):
        """Phase 1: greedy ascent, flip the top-k bits with positive BV."""
        bv = torch.cat([(-p.data.sign() * p.grad.data).view(-1)
                        for g in self.param_groups
                        for p in g['params'] if p.grad is not None])

        vals, topk = torch.topk(bv, min(self.k, bv.numel()))
        topk = set(topk[vals > 0].cpu().tolist())  # only BV_i > 0

        offset = 0
        for g in self.param_groups:
            g['adv_flips'] = []
            for i, p in enumerate(g['params']):
                if p.grad is None: continue
                idx = [j - offset for j in topk
                       if offset <= j < offset + p.numel()]
                if idx:
                    idx = torch.tensor(idx, device=p.device)
                    g['adv_flips'].append((i, idx))
                    p.view(-1)[idx] *= -1
                offset += p.numel()

        if zero_grad: self.zero_grad()

    @torch.no_grad()
    def second_step(self, zero_grad=False):
        """Phase 2: restore, then descend with the signal at w_adv."""
        for g in self.param_groups:  # restore the flipped bits
            for i, idx in g.get('adv_flips', []):
                g['params'][i].view(-1)[idx] *= -1
        if isinstance(self.descent, torch.optim.Optimizer):
            self.descent.step()  # STE backend: latent update at w_adv
        else:
            for g in self.param_groups:
                for i, p in enumerate(g['params']):
                    if p.grad is not None:
                        self.descent(g, i)

        if zero_grad: self.zero_grad()

    @staticmethod
    def _accumulator_update(g, i):
        """Example Descent with BOLD accumulator step (Algorithm 8 in BOLD paper)."""
        p, a = g['params'][i], g['accums'][i]
        a.mul_(g['ratios'][i]).add_(p.grad, alpha=g['lr'])   
        flip = a * p >= 1                                    
        p[flip] *= -1
        a[flip] = 0                                         
        g['ratios'][i] = 1.0 - flip.float().mean().item()    
\end{lstlisting}
\end{algorithm}


\end{document}